\documentclass{article}

\usepackage[final]{neurips_2022}

\usepackage[utf8]{inputenc} % allow utf-8 input
\usepackage[T1]{fontenc}    % use 8-bit T1 fonts
\usepackage{hyperref}       % hyperlinks
\usepackage{url}            % simple URL typesetting
\usepackage{booktabs}       % professional-quality tables
\usepackage{amsfonts}       % blackboard math symbols
\usepackage{nicefrac}       % compact symbols for 1/2, etc.
\usepackage{microtype}      % microtypography
\usepackage{xcolor}         % colors
\usepackage{enumitem} 
\usepackage[utf8]{inputenc} % allow utf-8 input
\usepackage[T1]{fontenc}    % use 8-bit T1 fonts
\usepackage{hyperref}       % hyperlinks
\usepackage{url}            % simple URL typesetting
\usepackage{booktabs}       % professional-quality tables
\usepackage{amsfonts}       % blackboard math symbols
\usepackage{nicefrac}       % compact symbols for 1/2, etc.
\usepackage{microtype}      % microtypography
\usepackage{xcolor}         % colors
\usepackage{microtype}
\usepackage{graphicx}
\usepackage{booktabs} % for professional tables
\usepackage{amsfonts}
\usepackage{tikz}
\usepackage{tikz-qtree}
\usetikzlibrary{fit}
\usepackage{amssymb}
\usepackage{algorithmic}
\usepackage{algorithm}
\usepackage{mathtools}
\usepackage{amsmath}
\usepackage{amsthm}
\usepackage{subcaption}
\usepackage{xcolor}
\setcitestyle{numbers,close={]},open={[},citesep={,}}
\usepackage{hyperref}
\usepackage{cleveref}
\usepackage{lipsum}
\usepackage{wrapfig}

\DeclareMathOperator*{\argmin}{arg\,min}

\DeclareMathOperator*{\arginf}{arg\,inf}

\newcommand{\E}{\mathbb{E}}
\newcommand{\pr}{\mathbb{P}}

\newcommand{\si}{\sigma^{2}}
\newcommand{\sul}{\sum\limits}

\newcommand{\F}{\mathcal{F}}
\newcommand{\rv}{\mathbb{R}_{+}^{|V|}}
\newcommand{\X}{\mathcal{X}}
\newcommand{\T}{\mathcal{T}}
\newcommand{\cL}{\mathcal{L}}
\newcommand{\cE}{\mathcal{E}}
\newcommand{\D}{D}
\newcommand{\algnm}{\textsc{GP-MD}}
\newcommand{\septnm}{\textsc{MinC-Known}}
\newcommand{\sm}{\sum_{m=1}^{N_{ep}}\sum_{h=1}^{H}}

\tikzstyle{startstop} = [rectangle, rounded corners, minimum width=3cm, minimum height=2cm,text centered, text width=6cm, draw=black, fill=blue!20]
\tikzstyle{io} = [rectangle, rounded corners, minimum width=3cm, minimum height=2cm, text centered, text width=6.5cm, draw=black, fill=blue!20]
\tikzstyle{process} = [rectangle, rounded corners, minimum width=1cm, minimum height=2cm, text width=2.5cm, text centered, draw=black, fill=blue!20]
\tikzstyle{decision} = [rectangle, rounded corners, minimum width=3cm, minimum height=2cm, text width=3cm, text centered, draw=black, fill=blue!20]
\tikzstyle{kernel} = [rectangle, rounded corners, minimum width=3cm, minimum height=2cm, text centered, text width=4cm, draw=black, fill=blue!20]
\tikzstyle{carrow} = [Plain, minimum width=3cm, minimum height=2cm, text centered, text width=8cm, draw=black]
\tikzstyle{arrow} = [thick,->,>=stealth]

\newtheorem{theorem}{Theorem}

\newtheorem{lemma}{Lemma}

\newcommand{\lcb}{\text{lcb}}

\Crefname{equation}{Eq.}{Eqs.}
\Crefname{figure}{Fig.}{Figs.}
\Crefname{theorem}{Thm.}{Thms.}
\Crefname{center}{pic.}{pics.}
\title{Movement Penalized Bayesian Optimization\\ with Application to Wind Energy Systems}

\author{%
Shyam~Sundhar~Ramesh \\ %\thanks{Use footnote for providing further information
  ETH Zurich\\
  \texttt{shramesh@ethz.ch} \\
   \And
   Pier~Giuseppe~Sessa  \\
   ETH Zurich \\
   \texttt{sessap@ethz.ch} \\
   \AND
   Andreas~Krause \\
   ETH Zurich \\
   \texttt{krausea@ethz.ch} \\
   \And
   Ilija~Bogunovic \\
   University College London \\
   \texttt{i.bogunovic@ucl.ac.uk} \\
}

\begin{document}

\maketitle

\begin{abstract}
Contextual Bayesian optimization (CBO) is a powerful framework for sequential decision-making given side information, with important applications, e.g., in wind energy systems. In this setting, the learner receives context (e.g., weather conditions) at each round, and has to choose an action (e.g., turbine parameters). Standard algorithms assume no cost for switching their decisions at every round. However, in many practical applications, there is a cost associated with such changes, which should be minimized. We introduce the episodic CBO with movement costs problem and, based on the online learning approach for metrical task systems of \citet{coester2019pure}, propose a novel randomized mirror descent algorithm that makes use of Gaussian Process confidence bounds. We compare its performance with the offline optimal sequence for each episode and provide rigorous regret guarantees. We further demonstrate our approach on the important real-world application of altitude optimization for Airborne Wind Energy Systems. In the presence of substantial movement costs, our algorithm consistently outperforms standard CBO algorithms.\looseness=-1
\end{abstract}
%\vspace{-1.5ex}
\section{Introduction}
%\vspace{-1.5ex}
\label{sec: intro}

Bayesian optimization (BO) is a well-established framework for sequential black-box function optimization that %has found  numerous applications. % in numerous real-world applications. %, e.g., hyperparameter tuning \cite{snoek2012practical}, robotics \cite{lizotte2007automatic}, and chemical design \cite{griffiths2020constrained}, among others. 
relies on Gaussian Process (GP) models \cite{Rasmussen2006} to sequentially learn and optimize the  unknown objective. In many practical scenarios, however, one wants to additionally use available \emph{contextual} information when making decisions. In this setting, at each round, the learner receives a context from the environment and has to choose an action based upon it. %The key challenge of Contextual BO is thus to balance exploration by acquiring data to estimate the unknown function over the context-action space and to exploit by choosing an action that seems optimal according to the learned model. 
Previous works have developed contextual BO algorithms \cite{krause2011contextual, char2019offline, kirschner2020distributionally, park2020contextual}, and applied them to various important applications, e.g., vaccine design, nuclear fusion, database tuning, crop recommender systems, etc.
% automated energy savings \cite{ayala2021edgebol}, locomotion planning \cite{seyde2019locomotion} and e-commerce \cite{cai2018reinforcement}. 

A potential practical issue with these standard algorithms is that they  assume no explicit costs for \emph{switching} between their actions at every round. Frequent action changes can be extremely costly in many real-world applications. This work is motivated by the problem of real-time control of the altitude of an airborne wind energy (AWE) system.\footnote{AWE system is a wind turbine with a rotor supported in the air without a tower that can benefit from the persistence of wind at different high altitudes \cite{elliot2014flights}.} In AWE systems, the wind speed is often only measurable at the system's altitude, and determining the optimal operating altitude of an AWE system as the wind speed varies represents a challenging problem. Another fundamental challenge is that additional energy is required for adjusting the altitude, which makes the frequent altitude changes costly. Consequently, this work is motivated by the following question: 
%\begin{center}
%    \vspace{-3ex}
    \emph{How can we efficiently learn to optimize the AWE system's operating altitude despite varying wind conditions while minimizing the energy cost associated with turbine altitude changes?}%\vspace{-3ex}
%end{center}
\begin{wrapfigure}{R}{0.5\textwidth}
\centering
\begin{subfigure}[b]{0.5\textwidth}
\centering
\vspace{-2em}
\includegraphics[width=0.8\textwidth]{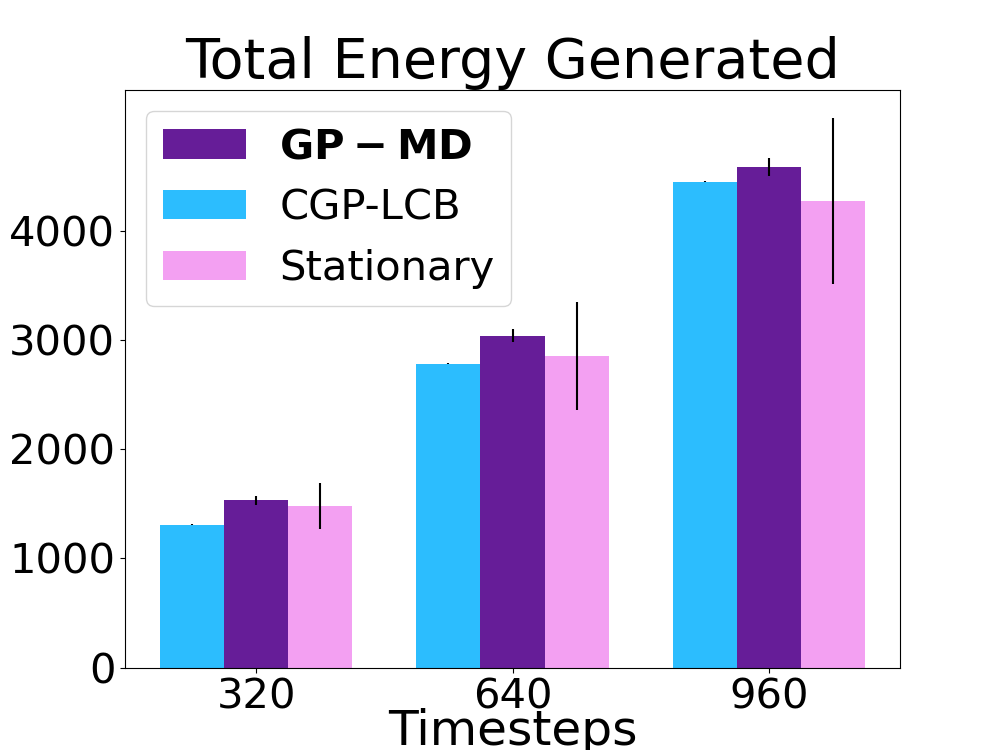}
%\vskip-5pt
\end{subfigure}
%\hspace{1.5em}
%\vspace{-1em}
 \caption{\small Total energy generated by the AWE system when operated with \algnm  ~(proposed in this work), \textsc{CGP-LCB}~\cite{krause2011contextual} and a stationary baseline. The stationary baseline employs no-learning and does not incur movement costs. GP-MD and CGP-LCB learn an operating strategy, but GP-MD outperforms CGP-LCB since it also considers movement costs.\looseness=-1
}\label{fig:teaser}
\vspace{-1em}
\end{wrapfigure}

In this work, we formalize the \emph{movement penalized} contextual BO problem. When the switching cost is a \emph{metric} (distance function), we propose a novel algorithm
%randomized algorithm running in the tree-based metric space. Our algorithm 
that effectively combines ideas from BO with the online learning strategies proposed in \cite{coester2019pure} for solving the so-called \emph{metrical task system} (MTS) problem \cite{borodin1992optimal}. %Because the actual objective is unknown apriori, the algorithm uses GP bandit techniques to construct confidence intervals around it that are %updated based on actively collected data and then used by the mirror descent subroutine.
Furthermore, our algorithm relies solely on noisy point evaluations (i.e., bandit feedback)%to learn and optimize the unknown objective
, allows for arbitrary context sequences, and besides the standard exploration-exploitation trade-off, it also balances the movement costs. As a result, it outperforms the standard movement-cost-agnostic contextual BO algorithms as well as movement-conservative baselines (see \Cref{fig:teaser}).\looseness=-1

% This should be revised to contain the following four paragraphs:
% {\color{purple}\begin{itemize}
%     \item BO and contextual BO
%     \item BO with costs and switching costs
%     \item The MTS problem
%     \item BO applications to wind energy systems
% \end{itemize}}
\textbf{Related Work.} Bayesian optimization (BO) refers to a sequential approach for optimizing an unknown objective (cost function) from noisy point evaluations. A great number of BO methods have been developed over the years (e.g., \cite{movckus1975, Srinivas2010,chowdhury2017kernelized}). While the focus of standard BO approaches is mainly on optimizing the unknown objective cost function, in this work, we additionally focus on penalizing frequent action changes. This problem makes the most sense in the \emph{contextual} BO setting \cite{krause2011contextual}, where the main objective changes with the observed contextual information and the learner also seeks to minimize the cost associated with frequent changes of its actions. Several works have tried to incorporate such and similar cost functions in the BO setting. \cite{lin2021bayesian} explicitly consider switching costs based on how deep into the pipeline the change in variable occurs. But they do not consider the contextual setting that is essential for our wind energy application and hence our work is not directly comparable with theirs. Moreover, it assumes that the system of interest has a modular structure (as detailed in \cite[Section-3]{lin2021bayesian}). In such a modular setting, the cost at each round is the number of modules that has to be changed rather than the amount of change of each variable. Other works include cost-aware and multi-objective sampling strategies in various settings such as batch \cite{kathuria2016batched, gonzalez2016batch}, multi-fidelity \cite{bogunovic2016truncated, kandasamy2017multi, song2019general}, multi-objective optimization \cite{abdolshah2019cost} and dynamic programming \cite{lam2016bayesian, lam2017lookahead}. Finally, two recent works \cite{folch2022snake,calandriello2022scaling} consider the problem of switching cost minimization in Bayesian optimization. They both consider the non-contextual setting, and while \cite{folch2022snake} lacks theoretical guarantees, \cite{calandriello2022scaling} does not explicitly consider the cost in the regret definition. Similarly, we consider movement/switching costs, but unlike these previous work, we specifically focus on minimizing the movement costs in the contextual setting.\looseness=-1

%\cite{koren2017}
%The usual BO setting does not consider the cost to switch our decisions which might be relevant in various practical settings (e.g. energy loss for changing the altitude of turbine).
% Several works have tried to tackle such a switching cost setting using various approaches including \cite{lin2021bayesian} which explicitly considers switching cost based on how deep into the pipeline the change in variable happens. Other related works in the past decade include cost-aware and multi-objective sampling strategies in various evaluation settings such as batch optimization \cite{kathuria2016batched},\cite{gonzalez2016batch}, multi-fidelity model \cite{kandasamy2017multi},\cite{kandasamy2016gaussian}, multi-objective optimization \cite{abdolshah2019cost} and dynamic programming \cite{lam2016bayesian},\cite{lam2017lookahead}. 

 % But this generalized result on any finite metric space $\X$ was constructed by first approximating $\X$ by a $\tau$-HST metric space (\cref{sec: MTS Setup}) with $\mathcal{O}(\log n)$ error \cite{bartal1996probabilistic}. Along with an  $\mathcal{O}(\log n)$-competitive randomized algorithm for MTS on $\tau$-HST metric spaces the $\mathcal{O}(\big(\log n)^{2}\big)$ bound was established.

The \emph{Metrical Task Systems} (MTS) problem \cite{borodin1992optimal} is a sequential decision-making problem widely studied in the \emph{online learning} literature. At each round, the learner observes a service cost function, chooses an action, and incurs the corresponding cost together with a movement cost penalizing the distance (according to some metric) between the current action and the one chosen at the previous round. The MTS problem is directly related to our problem setting (see \Cref{sec: MTS Setup}). After a long series of works, in \cite{bubeck2021metrical} and \cite{coester2019pure}, an  $\mathcal{O}\big((\log n)^{2}\big)$-competitive algorithm for MTS on any finite metric space was established. The approach of \citet{coester2019pure} relies on a tree representation of the decision space and an action randomization scheme via a mirror descent procedure. In contrast to our setting, these works assume that the service costs are {\em known} (i.e., they assume the {\em full-information} feedback). Moreover, in the online optimization literature, other related works study the online \emph{convex} optimization with switching costs \cite{goel2019beyond,shi2020online,goel2019online,lin2012dynamic} and convex body chasing problems \cite{bubeck2019competitively,argue2021chasing,sellke2020chasing, bubeck2020chasing}. We make no use of convexity and take a model-based (GP) approach to learn about the unknown service costs. \cite{lin2020online} consider non-convex objective functions, however, they impose certain conditions on the objective that are not easily modeled via GP as done in our work.% that may not be suitable for GP modeling as done in our work. 

%  \textcolor{red}{Even though, \cite{lin2020online} consider non-convex objective functions, they assume that the value of the objective function at any $x$ is higher by a factor than the cost/distance function between $x$ and a fixed point $v_t$. Such an assumption will not be suitable for GP modeling as done in our work
% }

The use of GP confidence bounds in online learning settings has been previously explored, e.g., in repeated multi-agent~\cite{sessa19noregret, sessa20contextual} and sequential games~\cite{sessa20sequential}, and to discover randomized max-min strategies~\cite{sessa20mixed}. However, none of these works has considered movement costs in the objective. This makes our problem  significantly different from the aforementioned ones, and requires a suitable action randomization scheme that can trade-off exploration, exploitation, and movement costs simultaneously.\looseness=-1
%{\color{red} TODO: Add 1-2 sentences on our previous works where we also use Online Learning Algorithms + Confidence bounds, e.g., GP-MW, Mixed strategies, or similar.}

% In the Online algorithms literature, there has been a long series of works involving movement costs. The MTS problem is directly related to our problem setting and after a long series of works, in \cite{bubeck2021metrical} and \cite{coester2019pure}, an  $\mathcal{O}\big((\log n)^{2}\big)$-competitive algorithm for MTS on any finite metric space was established. 

Various works have applied Bayesian Optimization in the context of \emph{wind energy systems} before. \cite{park2015bayesian,doekemeijer2019model,andersson2020real} use BO techniques to maximize the total energy yield in a wind farm in a  cooperative or a closed-loop framework. \cite{moustakis2019practical} use it to tune the parameters of the wind turbine to learn the effective wind speed. 
\cite{baheri2019combined} use BO for plant design (i.e., physical system design) of airborne wind energy systems. \cite{yazici4short} apply BO and other regression techniques to predict short-term wind energy production to make informed production offers. Finally, \cite{alkesaiberi2022efficient} survey several different methods for efficient wind-power prediction using machine-learning methods including BO. Our problem formulation differs from the above mentioned works, since we explicitly consider movement energy loss caused by the altitude change of the system.

%We consider the setting where there is a known and fixed metric determining the movement cost of our decisions. This setting is directly motivated by Airborne Wind Energy Systems(AWE) which tries to avail the high altitude winds. The operating altitude of such systems are flexible and can be shifted to an altitude wherein the wind speed is the highest. But such changes to operating altitude does cost some energy. Hence we want to balance between operating at the best altitude and minimizing the energy loss to change altitudes. The wind pattern(hence the power pattern) over the altitudes are weather/season dependent and naturally call for a Contextual BO approach to model the wind(power) pattern. And the energy loss due to altitude change can be modelled as movement cost.

\textbf{Contributions.}
%Our main contributions are as follows:\looseness=-1
%\begin{itemize}[noitemsep,topsep=0ex]
    % \item We formally introduce the contextual Bayesian optimization with movement costs problem and propose Mirror Descent based algorithm \ref{alg: mgpbo}. It first approximates the metric space by a $\tau-$HST space and runs mirror descent  over each node based on lower confidence bound on the objective function.
    %\item 
    We formally introduce Bayesian optimization with movement costs and propose a novel GP-MD algorithm (in \Cref{alg: mgpbo}). \algnm~%relies on noisy point evaluations to learn about the unknown objective (i.e.,~bandit feedback) and crucially 
    combines the online mirror descent (MD) algorithm with shrinking Gaussian Process (GP) confidence bound to decide on which point to evaluate next. %(see \Cref{sec: Alg and guarantee}). %{\color{purple} TODO We theoretically analyze the performance of the algorithm using the competitive ratio guarantees provided in \cite{coester2019pure}. We compare the performance with the episodic offline optimal sequences and provide $ (\alpha, \beta)-$approximate regret guarantees with convergence rate of $\mathcal{O}^{*}\big(N_{ep}^{\frac{1}{2}}\big)$ for $N_{ep}$ episodes and constant episodic length $H$.} 
    In our theoretical analysis, we establish rigorous sublinear regret guarantees for our algorithm by combining techniques from Bayesian optimization and metrical task systems approaches \cite{coester2019pure}.
    Finally, we demonstrate that \algnm~is able to successfully outperform previous contextual Bayesian optimization approaches on both synthetic and real-world data in the presence of movement costs. In particular, we consider the application to airborne wind energy systems and demonstrate that \algnm~can effectively operate such a system by considering movement costs and varying environmental conditions.\looseness=-1
%\end{itemize}

%\newpage
%\vspace{-1ex}
\section{Problem Statement}
\label{sec: Problem Formulation}
%\vspace{-1ex}
Let $f:\X \times \cE\to \mathbb{R_{+}}$ be an \emph{unknown} cost function defined over $\X\times\cE \subset \mathbb{R}^{p}$, where $\X$ is a finite set of actions, i.e., $|\X|=n$, and $\cE$ represents convex and compact space of contexts. We denote the \emph{known} metric (i.e., distance function) of $\X$ as $d(\cdot,\cdot)$, and similarly to other works in Bayesian optimization (e.g., \cite{Srinivas2010, chowdhury2017kernelized}) assume that the target cost function $f$ belongs to a reproducing kernel Hilbert space (RKHS) $\mathcal{H}_k$ of functions (defined on $\X \times \cE$), that corresponds to a known kernel $k : (\X \times \cE) \times (\X\times \cE) \to \mathbb{R}_{+}$ with $k((x,e),(x',e'))\leq 1$ for any action-context pair. 
In particular, we assume that for some known $B>0$, the target cost $f$ has a bounded RKHS norm, i.e., $f \in \F_{k} = \lbrace f \in \mathcal{H}_k: \|f\|_{k}\leq B \rbrace$. Also, we assume that the diameter of $\X$ ($\max_{x,x'\in \X}d(x,x')$) is bounded and denote it by $\psi$.
%Let the \emph{initial state} of the system correspond to action $x_{0} \in \X$, and d

We consider an episodic setting, wherein each episode runs over a finite time horizon $H$. Let the \emph{initial state} of the system in the first episode correspond to action $x_{0,1} \in \X$. At the end of every episode, the system resets to a new given initial action $x_{0,m} \in \X$ where $m\in$ $\lbrace 1,2,\dots,N_{ep} \rbrace$ denotes the episode index. In each episode $m$ and at every time step $h\in$ $\{1,2,\dots,H\}$, the environment reveals the context $e_{h,m} \in \cE$ to the learner. We make no assumptions on the context sequence provided by the environment (i.e., it can be arbitrary and different across episodes). The learner then chooses $x_{h,m} \in \X$ and observes the noisy function value:\looseness=-1
\begin{equation} \label{eq:noise_model}
    y_{h,m}=f(x_{h,m},e_{h,m})+\xi_{h,m},
\end{equation}
 where $\xi_{h,m}\sim \mathcal{N}(0,\si)$ with known $\sigma$, and independence over time steps. The goal of the learner is to minimize the cost incurred over the rounds in every episode, but at the same time to minimize the distance between its subsequent decisions as measured by $d(x_{h-1,m}, x_{h,m})$.  %Next, we formally state this optimization objective.\looseness=-1

Let $\D_{m}=\{x_{1,m},x_{2,m}, \dots, x_{H,m}\}$ denote the set of actions chosen by the learner over $H$ rounds in episode $m$. We recall that each action $x_{h,m} \in \D_{m}$ is chosen after observing the corresponding context $e_{h,m}$. The objective is to minimize the cumulative episodic cost for each episode $m$,
\begin{align}
\label{eq: formulate}
\begin{split}
    \mathrm{cost}_{m}(D_m)=\underbrace {\sum_{h=1}^{H}f(x_{h,m},e_{h,m})}_{S_{m}(D_m)}+\underbrace{\sum_{h=1}^{H}d(x_{h,m},x_{h-1,m})}_{M_{m}(D_m)},
\end{split}
\end{align}
where we refer to the two terms in \Cref{eq: formulate} as \emph{service cost} $S_{m}$ and \emph{movement cost} $M_{m}$. 

When $f$ is known, the problem can be seen as a MTS instance as detailed in \Cref{sec: MTS Setup}. Even in such a case, we cannot hope to solve this problem optimally, and nearly-optimal approximate algorithms were recently proposed (see \citet{coester2019pure}). 
%The best performance achieved in such a case is $\mathcal{O}\big((\log n)^{2}\big)$-competitive ratio and was recently proposed by \citet{coester2019pure}. Based on this performance in the known $f$ case we define the notion of $ (\alpha, \beta)$-approximate regret. 
%Consequently, we seek to match the performance to that of the best performance in the known $f$ case. % with an additional factor depending on $\alpha$. 
%As stated in \Cref{th: Theorem-1}, $\alpha=\mathcal{O}\big((\log n)^{3}\big)$ and additional factor being $\mathcal{O}\big(\log n\big)$. 
Hence, the learner's performance in episode $m$ is measured via $(\alpha, \beta)$-approximate regret: 
\begin{equation} \label{eq: Episodic Regret}
    r_{m}^{\alpha,\beta}=\mathrm{cost}_{m}(D_{m})-\alpha \cdot \mathrm{cost}_{m}(D_m^{*})-\beta,
\end{equation} 
where $D_m^*:=\argmin_{D \subset \X,\; |D| = H} \mathrm{cost}_{m}(\D)$ is the offline optimal action sequence obtained assuming the knowledge of the true sequence of contexts $\lbrace e_{h,m}\rbrace_{h=1}^H$ {\em in advance}, and $\alpha$ and $\beta$ are approximation constants (independent of $N_{ep}$). %We note that action sequence $\D_{m}^{*}$ is obtained assuming the knowledge of the true sequence of contexts $\lbrace e_{h,m}\rbrace_{h=1}^H$ in advance. 
In contrast, in our setting, the learner only gets to see the {\em current context} when making a decision and has no knowledge about the future ones.

%depending on $n$ and 
%Here the expectation is conditioned on $\mathcal{F}_{m-1}$ as $D_{m}$ is outputted by our algorithm based on $ \lcb_{m}$ which in turn is constructed from $\mathcal{F}_{m-1}$. Note that, $ R_{m}^{\alpha,\beta}$ will still be a random quantity as it is a conditional expectation w.r.t data collected which is still random. 

After $N_{ep}$ episodes, the total cumulative regret is defined as
\begin{equation} \label{eq: Cumulative Regret}
    R_{N_{ep}}^{\alpha,\beta} = \sum_{m=1}^{N_{ep}}r_{m}^{\alpha,\beta}.
\end{equation}
We seek an algorithm whose total cumulative regret grows sublinearly in $N_{ep}$, so that $\lim_{N_{ep} \to \infty}R_{H,N_{ep}}^{\alpha,\beta}/N_{ep} = 0$, for any set of initial states $\{x_{0,m}\}_{m=1}^{N_{ep}} \subset \X$.

% Finally, we note that the notion of $ (\alpha, \beta)$-approximate regret is heavily influenced from a performance metric known as the \emph{competitive ratio} and is often used in the online learning literature (e.g., \citet{Bubeck2018}, \citet{coester2019pure}). The $\alpha$ and $\beta$ in our regret definition are based on the competitive ratio guarantees provided in \cite{coester2019pure}.\looseness=-1

% \section{Proposed Approach and Preliminaries}
% We start this section by recalling the Gaussian Process model \cite{Rasmussen2006}, and results that will be useful in designing our main algorithm.
%\vspace{-1ex}
\subsection{Gaussian Process Model}\label{sec:gps}
%\vspace{-1ex}
In standard Bayesian optimization, a surrogate Gaussian Process model is typically used to model the target cost function. A Gaussian Process $GP(\mu(\cdot), k(\cdot, \cdot))$ over the input domain $\X \times \cE$, is a collection of random variables $(f(x, e))_{x \in \X, e \in \cE}$ where every finite number of them $(f(x_i, e_i))_{i=1}^n$, $n\in \mathbb{N}$, is jointly Gaussian with mean $\E[f(x_i)] = \mu(x_i, e_i)$ and covariance $\E[(f(x_i,e_i) - \mu(x_i,e_i))(f(x_j,e_j ) - \mu(x_j,e_j))] = k( (x_i, e_i), (x_j,e_j))$ for every $1 \leq i, j \leq n$. 

BO algorithms typically use zero-mean GP priors to model uncertainty in $f$, i.e., $f \sim GP(0, k(\cdot, \cdot))$, and Gaussian likelihood models for the observed data. As more data points are observed, GP (Bayesian) posterior updates are performed in which noise variables are assumed to be drawn independently across $t$ from $\mathcal{N}(0, \lambda)$. Here, $\lambda$ is a hyperparameter that might be different from the true noise variance $\sigma^2$. More precisely, given a sequence of previously queried points and their noisy observations the posterior is again Gaussian, with the posterior mean and variance given by: 
\begin{align}
  \mu_{t}(x, e) &= k_{t}(x,e)^T(K_{t} + \lambda I_{t})^{-1} Y_{t} \label{eq:posterior_mean} \\
  \sigma^2_{t}(x,e) &= k((x,e),(x,e)) - k_{t}(x,e)^T (K_{t} + \lambda I_{t})^{-1} k_{t}(x,e) \label{eq:posterior_stdev},
\end{align}
where $Y_{t} := [y_1, \dots, y_{t}]$ denotes a vector of observations, $K_{t} = [k((x_s, e_s), (x_{s'}, e_{s'})]_{s,s'\leq t}$ is the corresponding kernel matrix, and $k_{t}(x,e) = [k((x_1, e_1),(x,e)),\dots, k((x_{t},e_{t}),(x,e))]^T \in \mathbb{R}^{t \times 1}$.

\textbf{Maximum Information Gain.}
In the standard Bayesian optimization, the main quantity that characterizes the complexity of optimizing the target cost function is the maximum information gain \cite{Srinivas2010} defined at time $t$ as: 
\begin{equation} \label{eq:max_info_gain}
    \gamma_{t} = \max_{\{(x_{i},e_{i})\}_{i=1}^{t}} I (Y_t; f),
    %\frac{1}{2} \log \det (\textbf{1}_{T}+\lambda^{-1}K_{T}). 
\end{equation}
where $I(Y_t; f)$ denotes the mutual information between random observations $Y_t$ and GP model $f$. 
The mutual information for the GP model is given as:
\begin{equation} \label{eq:mutual_information}
    I(Y_t; f) = \frac{1}{2} \log \det (I_t +\lambda^{-1}K_{t}). 
\end{equation}
This quantity is kernel-specific and for compact and convex domains $\gamma_t$ is sublinear in $t$ for various classes of kernel functions \cite{Srinivas2010} as well as for kernel compositions (e.g., additive kernels in \cite{krause2011contextual}).

% This quantity is kernel-specific and for compact and convex domains $\gamma_t$ is sublinear in $t$ for various classes of kernel functions \cite{Srinivas2010} (e.g., $\mathcal{O}(d \log t)$ for the linear kernel, $\mathcal{O}((\log t)^{d+1})$ for the squared-exponential kernel, and $\mathcal{O}\big(t^{\tfrac{d}{2\nu + d}}(\log t)^{\tfrac{2\nu}{2\nu + d}}\big)$ for the Mat\'ern kernel~\cite{vakili2021information}), as well as for kernel compositions (e.g., additive kernels in \cite{krause2011contextual}).

\textbf{Confidence Bounds.}
We also use the following result (\cite{Srinivas2010, Yadkori2013, chowdhury2017kernelized}) that is frequently used in Bayesian optimization to provide confidence bounds around the unknown function.
\begin{lemma}\label{lemma:confidence_lemma}
Assume the $\sigma$-sub-Gaussian noise model as in \Cref{eq:noise_model}, and let $f$ belong to $\F_k$.  Then, the following holds with probability at least $1- \delta$ simultaneously over all $t \geq 1$ and $x \in \X$, $e \in \mathcal{E}$:
    \begin{equation}
        |\mu_{t}(x,e)-f(x,e)|\leq \beta_{t}\sigma_{t}(x,e),
    \end{equation}
    where $\beta_t = \tfrac{\sigma}{\lambda^{1/2}} \sqrt{2 \ln (1/\delta) + 2\gamma_{t} }+B$, and $\mu_t$ and $\sigma_t$ are defined in \Cref{eq:posterior_mean,eq:posterior_stdev} with $\lambda > 0$.
\end{lemma}
% We explicitly define the upper and lower confidence bound for every $x \in \X, e \in \cE$: 
Based on the previous, we also define the lower confidence bound for every $x \in \X, e \in \cE$ as:
\begin{equation}
    %\ucb_{t}(x,e)&:= \mu_t(x,e)+\beta_{t}\sigma_{t}(x,e), \label{eq:ucb}\\
    %\lcb_{t}(x,e)&:= \max{\lbrace \mu_t(x,e)-\beta_{t}\sigma_{t}(x,e),0\rbrace}, \label{eq:lcb}
    \lcb_{t}(x,e):= \mu_t(x,e)-\beta_{t}\sigma_{t}(x,e). \label{eq:lcb}
\end{equation}
%where in \Cref{eq:lcb} we also use the fact that the cost function is non-negative. 
%  Hence, \Cref{eq:lcb} lower bounds the target cost function with high probability, and we explicitly use it in the proposed algorithm. 
We use $\lcb_{m}(x,e)$ when it is computed based on data collected before episode $m$. 
%  Finally, we note that in the episodic (batch) setting, we update $\lcb_{t}(x,e)$ only at the end of every episode using the data collected up until that episode. Hence we denote the confidence bounds as  $\ucb_{m}(x,e)$ and $\lcb_{m}(x,e)$ for episode $m$ which is used for all rounds in episode $m$.
%\vspace{-1ex}
\subsection{Relation to Metrical Task Systems (MTS)}
\label{sec: MTS Setup}
%\vspace{-1ex}
When $f$ is known, our optimization objective in \Cref{eq: formulate} can be seen as a particular type of MTS problem, where $f(\cdot, e_{h,m})$
%$c_{h,m}(\cdot) = f(\cdot, e_{h,m})$ 
is the MTS service cost that changes for every $h$ and $m$. 
% for a particular episode $m$. 
Compared to a standard MTS (see \Cref{sec: MTS}), our problem formulation is more challenging since %cost function $f$ is a-priori unknown %(hence service cost $c_{h,m}(\cdot)$ is not available), 
%and 
the learner can only learn about $f$ from previously observed data.
The approach proposed in this paper builds on the algorithm by \citet{coester2019pure} for standard MTS problems. However, to cope with the aforementioned challenge, our approach exploits the regularity assumptions regarding $f$ and utilizes the constructed lower confidence bounds \Cref{eq:lcb} to  \emph{hallucinate} information about the unavailable service cost at each round. Before presenting our overall approach, we describe a preliminary step proposed by \cite{coester2019pure}, which consists of representing our metric space $(\X, d)$ by a \emph{Hierarchically Separated Tree} (HST) metric space.

%\begin{definition}[HST]
\textbf{HST metric space.} Consider a tree $\mathcal{T}=(V,E)$ with root $r$, leaves $\cL \subset V$ and non-negative weights $w_{v}$, for each $ v \in V$, which are non-increasing along root-leaf paths. Let $d_{\T}(l,l')$ denote a  distance metric between any two leaves $l,l'\in \cL$ given as the sum of the encountered weights on the path from $l$ to $l'$ (see \Cref{fig:distance example}).  $(\cL,d_{\T})$ is a HST metric space, and $\tau$-HST metric space if the weights are exponentially decreasing, i.e., $w_{u}\leq w_{v}/\tau$, with $v$ being the parent of $u$.
%\end{definition}

Similarly to \cite{coester2019pure}, we use the algorithm from  \cite{fakcharoenphol2004tight} (which we name via the authors’ surnames as FRT) to approximate the given metric space ($\X,d$) by a $\tau$-HST one. In particular, we use FRT in \Cref{alg: mgpbo} as a computationally efficient preprocessing step to create a tree $\T$ with leaves $\cL$ corresponding to actions in $\X$, distance metric $d_{\T}$, and root node $r$.  We explain the intrinsic MTS motivation for this preprocessing step in \Cref{sec:hst_metric_sec}, and defer additional details to  \Cref{sec: FRT}.
%\vspace{-1ex}
\section{The \algnm\ Algorithm}
%\vspace{-1ex}
\label{sec: Alg and guarantee}
%As mentioned in the previous section, we run a mirror descent algorithm over each internal vertex $u\in V$. But as $f(\cdot,e_{h,m})$ is defined only over the leaves, we need to define a cost corresponding to each internal vertex. 

\begin{algorithm}[t!]
    \caption{$\algnm$}
    \begin{algorithmic}[1]
        \STATE \textbf{Require:}  Action space $\X$, kernel function $k(\cdot,\cdot)$, metric $d(\cdot,\cdot)$
        % \STATE Build (using FRT algo.) a $\tau$-HST,  $\T=(V,E)$, with leaves $\cL=\X$ and topological ordering $\mathcal{OD}$($V\backslash\cL$).
        \STATE Run FRT$(\X,d(\cdot,\cdot))$ and obtain $\tau$-HST  $\T=(V,E)$ with leaves $\cL=\X$% and %$\mathcal{OD}$($V\backslash\cL$)%\STATE Set $\mu_1(x,e)=0$ and $\sigma_1(x,e)=k((x,e),(x,e))$, for all $x\in \X$ and $e\in \cE $
        \FOR{$m=1,\dots, N_{ep}$}
            \STATE Receive $x_{0,m}$ and initialize $z_{0,m}$ (\Cref{eq: z initialize-1}), conditional prob. $q_0=\Delta^{-1}(z_{0,m})$ as in \Cref{eq: Q-Z relation}
            \FOR{$h=1,\dots, H$}
                \STATE Observe context $e_{h,m}$ and initialize costs: 
               $\lcb_m(v,e_{h,m})=0$, $ \forall v \in V\backslash\cL$
                \FOR{$u \in \mathcal{OD}(V\backslash\cL)$}
                        \STATE Update vertex prob. $q_h^{(u)}$ from $q_{h-1}^{(u)}$ and $\lcb_{m}(\cdot,e_{h,m})$ via Mirror Descent (\Cref{eq:MD_update})
                        \vspace{0.5em}
                    \STATE Update cost for vertex $u$:
                    %\vspace{-0.8em}
                    \begin{equation*}
                      \label{eq:cost update} \lcb_{m}(u,e_{h,m})=\langle q_{h}^{(u)},\lcb_{m}(\cdot,e_{h,m})\rangle=\sul_{\nu\in\mathcal{C}(u)}q_{h,\nu} \cdot \lcb_{m}(\nu,e_{h,m})
                    \end{equation*}
                 %\vspace{-1.5em}
                \ENDFOR
                \STATE Compute prob. vector $z_{h,m}=\Delta(q_h)$ (\Cref{eq: Q-Z relation}) and leaves' prob. $l(z_{h,m})$ (\Cref{eq: leaf probability})
                \STATE  Estimate optimal coupling $\zeta_{h-1,h,m}$ between $l(z_{h-1,m})$ and $l(z_{h,m})$ as in \Cref{eq: opt couple}
                \vspace{-0.0em}
                \STATE Sample action $x_{h,m} \sim \zeta_{h-1,h,m}(\cdot|x_{h-1,m})$ and observe $y_{h,m} = f(x_{h,m},e_{h,m}) + \xi_{h,m}$
            \ENDFOR
            \STATE Update $\mu_{m+1}(\cdot,\cdot)$ and $\sigma_{m+1}(\cdot,\cdot)$ as per \Cref{eq:posterior_mean} and \Cref{eq:posterior_stdev}
        \ENDFOR
    \end{algorithmic}
    \label{alg: mgpbo}
\end{algorithm}
In this section, we introduce $\algnm$, a novel algorithm for the  contextual BO problem with movement costs defined in \Cref{sec: Problem Formulation}. %As mentioned in the previous section, $\algnm$ builds on the randomized algorithm by \cite{coester2019pure} and starts by representing the metric space ($\X,d$) with a tree $\T=(V,E)$ and a root note denoted as $r$. Tree leaves $\cL$ correspond to the available actions in $\X$.
At each episode $m$ and round $h$, the state of $\algnm$ can be summarized by a vector of probabilities $z_{h,m}\in K_{\T}$ over the vertices of $\T$, where %$K_{\T}$ is the convex polytope in $\rv$ such that
$K_{\T}:=\Big\{z\in\rv: z_{r}=1, z_{u}=\sul_{\nu\in\mathcal{C}(u)}z_{\nu} \quad \forall u\in V \backslash \cL\Big\},$
and $\mathcal{C}(u)$ denotes the children of $u$. Each entry $z_\nu$ represents the probability that the selected action $x_{h,m}$ belongs to the leaves of the subtree rooted at $\nu$, i.e., $z_{\nu}=\pr(x_{h,m}\in \cL(\nu))$. Below, we specify how $z_{h,m}$ is computed at each round. Moreover, given any $z\in K_{\T}$, the vector
\begin{equation}
\label{eq: leaf probability}
l(z):=[z_{l}, \:l\in \cL]  \in [0,1]^{n},  
\end{equation}
defines a probability distribution over the leaves $\cL$, and hence the actions $\X$. As in \cite{coester2019pure}, given probability vectors $z_{h,m}$ and  $z_{h-1,m}$, $\algnm$ computes the \emph{minimal distance} distribution 
\begin{equation}\label{eq: opt couple}
    \zeta_{h-1,h,m}=\arginf\limits_{\zeta\in \Pi(l(z_{h-1,m}),l(z_{h,m}))}\E_{\zeta}[d_{\T}(U_{h-1,m}, U_{h,m})],
\end{equation}
where $U_{h-1,m}$ and $U_{h}$ are random variables having marginals $l(z_{h-1,m})$ and $l(z_{h,m})$ respectively. Finally, action $x_{h,m}$ is sampled from the conditional minimal distance distribution $x_{h,m} \sim \zeta_{h-1,h,m}(\cdot|x_{h-1,m})$ (Line 13 in \Cref{alg: mgpbo}). %\textcolor{blue}{ Note that the competitive ratio guarantees from \cite{coester2019pure} hold only if the expected movement cost is in the form of a Wasserstein-1-distance between subsequent marginal distributions. Sampling from this conditional minimal distance distribution ensures this and allows us to invoke the guarantees from \cite{coester2019pure}. } 
At the end of each episode $m$, the newly observed data are then used to update posterior mean and standard deviation about the cost function. 

Finally, we describe how probability vectors $z_{h,m}$ are computed at each round, a key step of $\algnm$ (Lines 8--12 in \Cref{alg: mgpbo}). We follow the recursive Mirror Descent (MD) procedure proposed by \cite{coester2019pure}, with the important difference that we are dealing with an \emph{unknown} context-dependent cost function. Hence, we make use of the Gaussian process model and corresponding confidence estimates as defined in \Cref{sec:gps}.

To obtain probabilities $z_{h,m}$, we consider \emph{conditional} probability vectors $q \in Q_{\T}$, where $Q_{\T}$ is the set of valid conditional probabilities
$Q_{\T}:=\Big\{q\in \mathcal{R}_{+}^{|V\backslash r|}:  \sul_{\nu\in\mathcal{C}(u)}q_{\nu} = 1 \quad \forall u\in V \backslash \cL\Big\}.$
For each vertex $\nu$ with parent $u$, $q_\nu$ represents the conditional probability $\pr(x_{h,m}\in \cL (\nu)|x_{h,m}\in \cL(u))$. Moreover, given $q_{h}\in Q_{\T}$ we define the vector $q_{h}^{(u)}:=[ q_{h,\nu}, \: \nu \in \mathcal{C}(u)]$ as the conditional distribution over children of $u$, and let $Q_{\T}^{(u)}$ be the set of all valid distributions $q_{h}^{(u)}$.\looseness=-1

In each episode $m$, conditional probability vector $q_{h}$ for round $h$ is obtained recursively, from leaves to root, as a function of $q_{h-1}$, the observed context $e_{h,m}$, and the current estimate about the cost associated to each particular vertex. More precisely, let $\mathcal{OD}(V\backslash\cL)$ be a topological ordering of the internal vertices $V\backslash\cL$ so that every child in $\T$ occurs before its parent. Then, for each $u \in \mathcal{OD}(V\backslash\cL)$ conditional probabilities $q_h^{(u)}$ are obtained via the Mirror Descent update: 
\begin{equation}\label{eq:MD_update}
    q_h^{(u)} =\argmin_{p\in Q_{\T}^{(u)}}\Big\{D^{(u)}(p\|q_{h-1}^{(u)}) +\langle p,\lcb_m^{(u)}(\cdot,e_{h,m})\rangle\Big\}.
\end{equation}
Function $D^{(u)}$ is the Bregman divergence with respect to a suitable potential function (see Appendix~\ref{app:divergence}), while 
$\lcb_m^{(u)}(\cdot,e_{h,m}) :=[\lcb_m(\nu,e_{h,m}), \: \forall\nu\in \mathcal{C}(u)]$
is a lower confidence bound estimate of the costs corresponding to children of vertex $u$. For $v\in \cL$, $\lcb_m(\nu,e_{h,m})$ are obtained by the GP-regression techniques outlined in \Cref{sec:gps}, while for internal vertices these are computed recursively from their children nodes as: \begin{equation}\label{eq: parental cost}
    \lcb_m(u,e_{h,m}):=\sul_{\nu\in\mathcal{C}(u)}q_{h,\nu} \lcb_m(\nu,e_{h,m})\,.
\end{equation}
% This update process continues recursively up to the tree root. 
The movement cost is primarily controlled by this usage of Bregman divergence based mirror descent. Also, sampling from the conditional minimal distance distribution further restricts movement between alternate actions. 
Once the vector of conditional probabilities $q_h\in Q_{\T}$ has been updated, we can obtain the corresponding probability vector $z_{h,m}$ via the mapping $\Delta: Q_{\T}\to K_{\T}$ such that: 
\begin{equation}
z=\Delta(q)  \hspace{0.5em}\Rightarrow  \hspace{0.5em}z_{\nu}=z_{u}q_{\nu} \quad \forall u\in V\backslash \cL,\; \nu \in \mathcal{C}(u).\label{eq: Q-Z relation}
\end{equation}
%\vspace{-1ex}
\subsection{Theoretical Guarantees}
%\vspace{-1ex}
Our main theorem bounds the cumulative regret of $\algnm$.
\begin{theorem}\label{th: Theorem-1}
Let $\X$ be represented by a $\tau$-HST space with $\tau > 4$ (Line~2 of \Cref{alg: mgpbo}), and set $\delta\in (0,1)$. Then, with probability at least $1-\delta$, the regret of %sequences of actions $(D_{m})_{m=1}^{N_{ep}}$ selected by 
$\algnm$ over $N_{ep}$ episodes %have a sublinear cumulative regret bound: 
is bounded by\looseness=-1
% \begin{align*}R_{H,N_{ep}}^{\alpha,\beta}\leq \mathcal{O}\Big(\beta_{N_{ep}}  \big(N_{ep}H \exp{(\gamma_{H})}\gamma_{HN_{ep}} 
%   &+4H\log(\delta^ {-1}) +8H\log(4H)+1 \big)^\frac{1}{2}+ \\ H(B+\mathrm{diam}(\mathcal{X}))\log\big(\tfrac{N_{ep}\log(N_{ep})}{\delta}\big)\Big),
% \end{align*}
\begin{align*}
R_{N_{ep}}^{\alpha,\beta}=\mathcal{O}\Big(\beta_{N_{ep}}  \big(N_{ep}H^{2}\gamma_{HN_{ep}} 
  + H\log(\tfrac{H}{\delta}) \big)^\frac{1}{2}+ H(B+\psi)\log\big(\tfrac{N_{ep}\log(N_{ep})}{\delta}\big)\Big),
\end{align*}
with approximation factors $\alpha=\mathcal{O}\big((\log n)^{2}\big)$ and $\beta=\mathcal{O}(1)$. Here, $H$ is the episodes' length, $\beta_{N_{ep}}$ is the confidence level from \Cref{lemma:confidence_lemma}, and   $\gamma_{HN_{ep}}$ is the maximum information gain defined in \Cref{eq:max_info_gain}.
\end{theorem}

For most of the popularly used kernels, \Cref{th: Theorem-1} can be made more explicit by substituting bounds on $\gamma_{HN_{ep}}$ (e.g., in the case of linear kernel and compact domain, we have $\gamma_t  = O(p \log t)$, while for squared-exponential kernel it holds $\gamma_t = O((\log t)^{p+1})$ \cite{Srinivas2010}; see also \Cref{sec:gps}). In such cases, we make the following two important observations regarding our result: i) The obtained regret bound is sublinear in the number of episodes $N_{ep}$ and hence  $\lim_{N_{ep} \to \infty}R_{N_{ep}}^{\alpha,\beta}/N_{ep} = 0$; ii) The bound is independent of the input space size, i.e., the number of actions $n$ (although the approximation factor $\alpha$ depends logarithmically on $n$, similarly to \cite{coester2019pure}). These imply that \algnm~approaches $\alpha$-competitive ratio performance of the MTS algorithm by \cite{coester2019pure}, while learning about the service cost from noisy point evaluations (i.e., bandit feedback) only. Finally, in our analysis, we treat $H$ as constant.
%and note that a similar linear dependence on $H$ is also present in the results and analysis of \cite{chowdhury2019online} that we also build upon (see~\cite[Theorem-1,2]{chowdhury2019online}). 
%and improving the dependency on $H$ in our bound is an interesting direction for future work. 

Proof of \Cref{th: Theorem-1} is detailed in \Cref{sec: regret proof}. Next, we outline some main steps. We make use of the competitive ratio guarantees for the used Mirror Descent algorithm from \cite[Corollary 4]{coester2019pure} to bound the expected hallucinated service cost $ \sum_{h=1}^{H}\langle \lcb_{m}(\cdot,e_{h,m}),l(z_{h,m})\rangle$. Here, we use the fact that $\lcb_{m}(\cdot,\cdot)$ does not change within an episode. Since this is not the actual service cost, we bound $\sum_{h=1}^{H}\langle f(\cdot,e_{h,m}),l(z_{h,m})\rangle$ by $\sum_{h=1}^{H}\langle \lcb_{m}(\cdot,e_{h,m}),l(z_{h,m})\rangle$ with an additional learning error. The sum of the learning errors over all episodes can be rewritten as $\sum_{m=1}^{N_{ep}}\sum_{h=1}^{H}\langle \sigma_{m}^{2}(\cdot,e_{h,m}),l(z_{h,m})\rangle$. We use the concentration of the conditional mean result from \cite[Lemma 3]{kirschner2018information} to upper bound it by the actual realizations $\sum_{m=1}^{N_{ep}}\sum_{h=1}^{H}\sigma_{m}^{2}(x_{h,m},e_{h,m})$, and use the result of \cite[Lemma-2]{chowdhury2019online}, to further upper bound it with the maximum information gain quantity $\gamma_{N_{ep}H}$. 

Finally, the movement cost 
%is calculated w.r.t. $d_{\T}(\cdot,\cdot)$ and the Wasserstein cost for this metric can be rewritten as $\sum_{m=1}^{N_{ep}}\sum_{h=1}^{H} \|z_{h,m}-z_{h-1,m}\|_{l_{1}(w)}$ (see \Cref{eq: MTS Wasserstein}) which 
can also be bounded similar to that of the previously mentioned expected hallucinated service cost $\sum_{h=1}^{H}\langle \lcb_{m}(\cdot,e_{h,m}),l(z_{h,m})\rangle$ with an extra $\alpha=\mathcal{O}\big((\log n)^{2}\big)$ factor using \cite[Corollary 4]{coester2019pure}.\looseness=-1 %using results from \cite{coester2019pure}.

\begin{figure*}[t!]
    \centering
    \begin{subfigure}[b]{.3\textwidth}
        \centering
        \includegraphics[width=\linewidth]{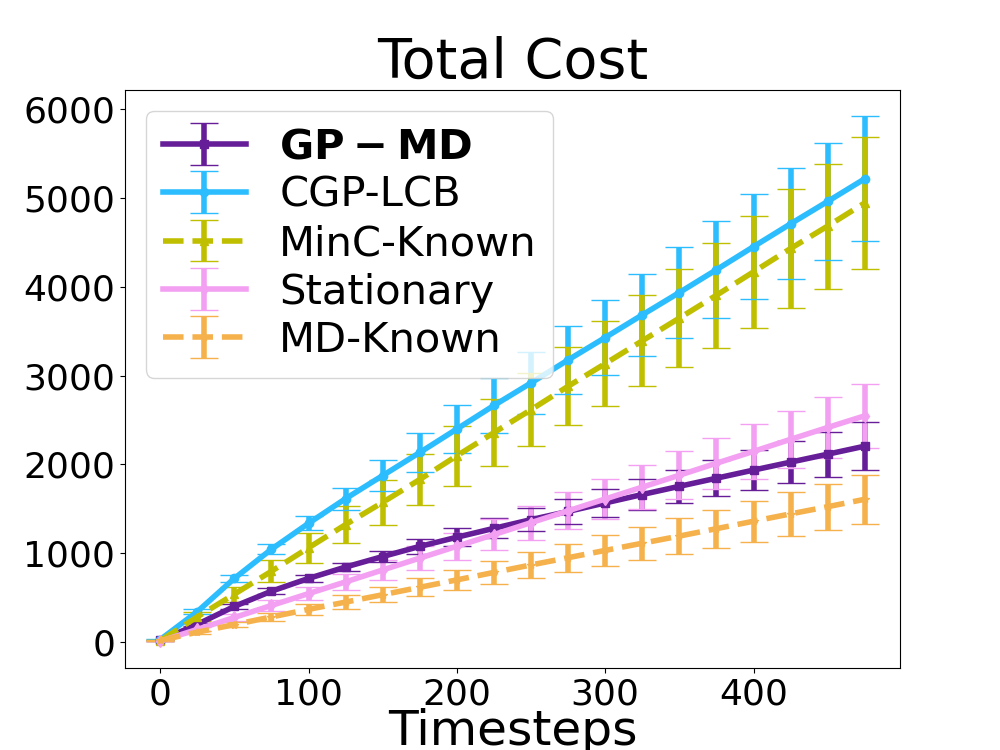}
        \caption{$\rho=0.5$}
        \label{fig:fig11}
    \end{subfigure}
    \hspace{1.5em}
    \begin{subfigure}[b]{.3\textwidth}
       \centering
        \includegraphics[width=\linewidth]{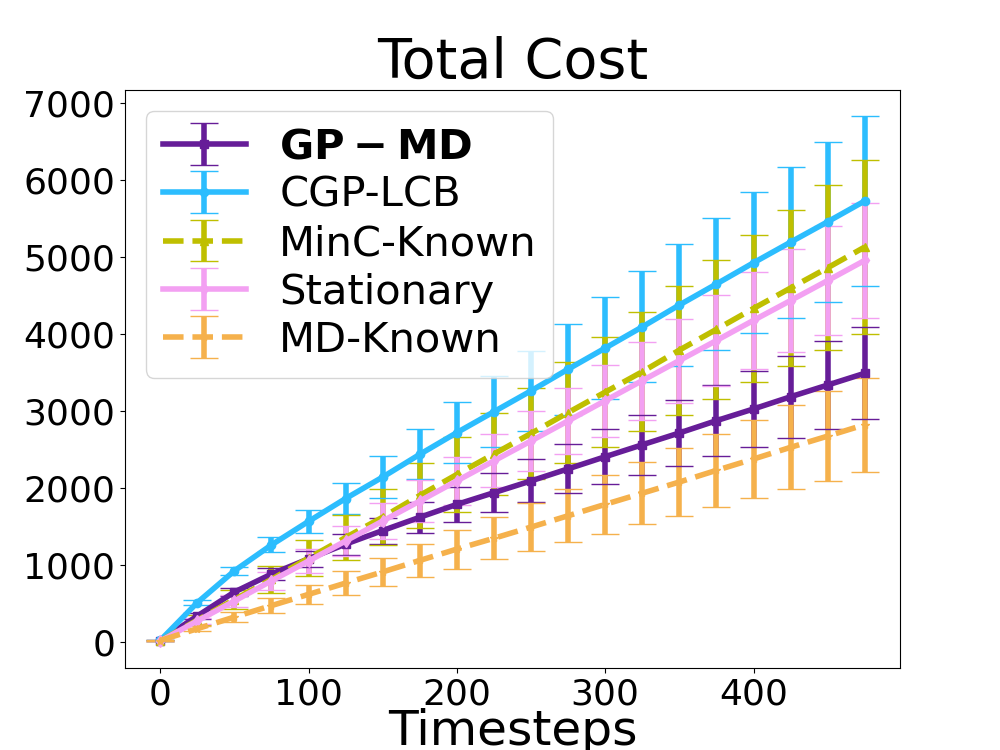}
        \caption{$\rho=1$}
        \label{fig:fig13}
    \end{subfigure}
    \hspace{1.5em}
    \begin{subfigure}[b]{.3\textwidth}
         \centering
        \includegraphics[width=\linewidth]{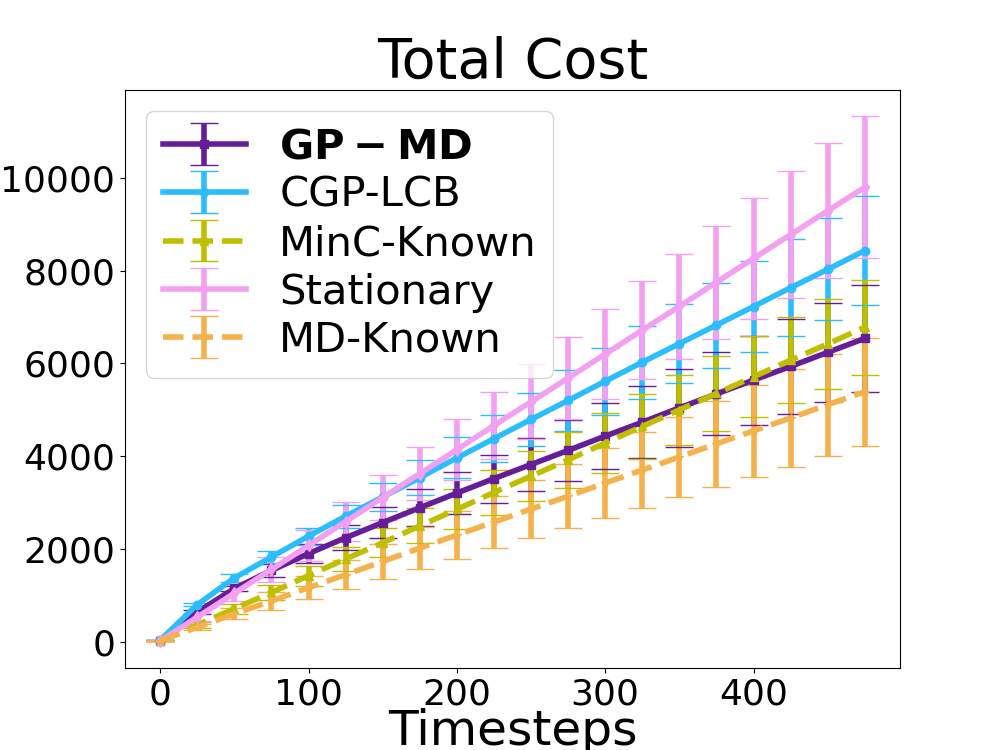}
        \caption{$\rho=2$}
        \label{fig:fig12}
    \end{subfigure}
    \vspace{1.5em}
    \begin{subfigure}[b]{.3\textwidth}
        \centering
        \includegraphics[width=\linewidth]{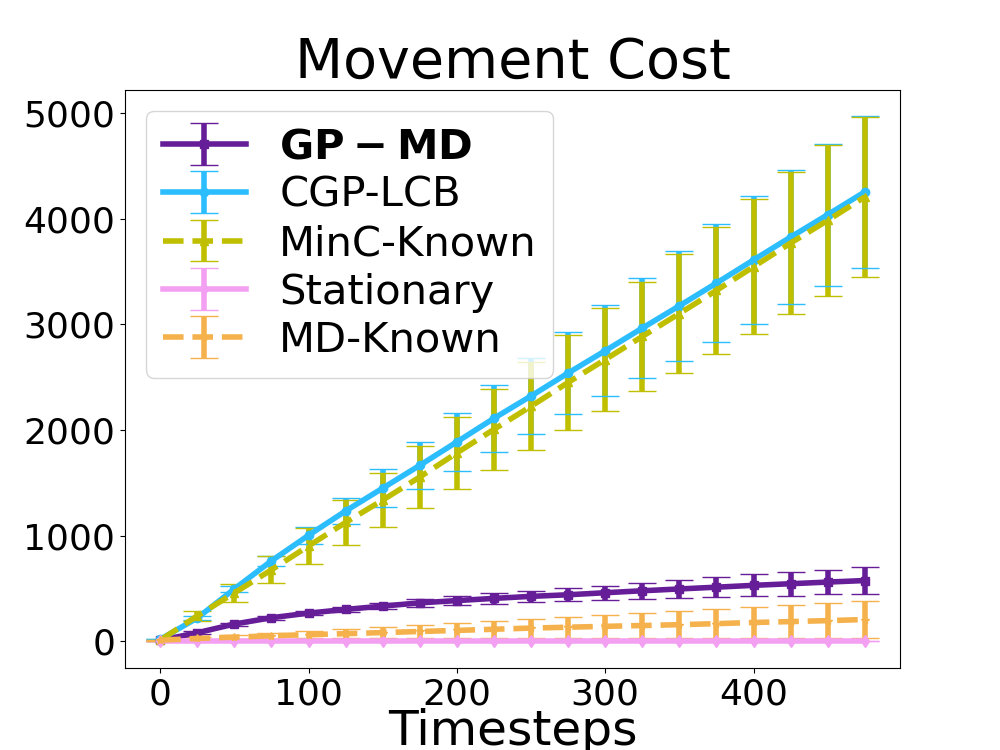}
        \caption{$\rho=0.5$}
        \label{fig:fig14}
    \end{subfigure}
    \hspace{1.5em}
    \begin{subfigure}[b]{.3\textwidth}
        \centering
        \includegraphics[width=\linewidth]{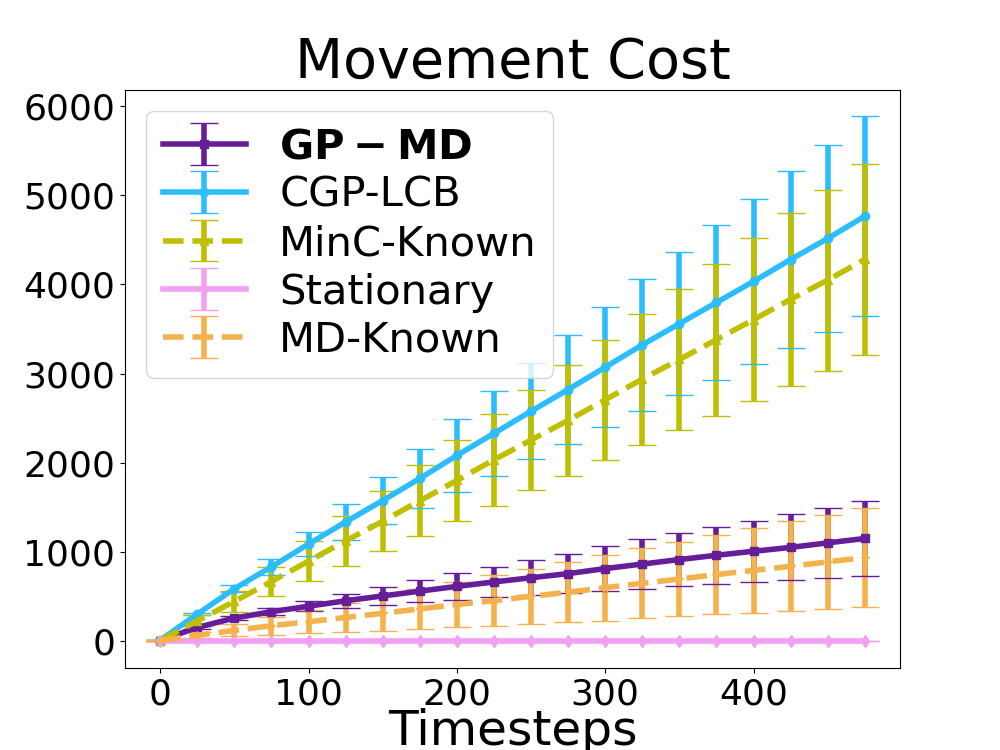}
        \caption{$\rho=2$}
        \label{fig:fig15}
    \end{subfigure}
    \hspace{1.5em}
    \begin{subfigure}[b]{.3\textwidth}
        \centering
        \includegraphics[width=\linewidth]{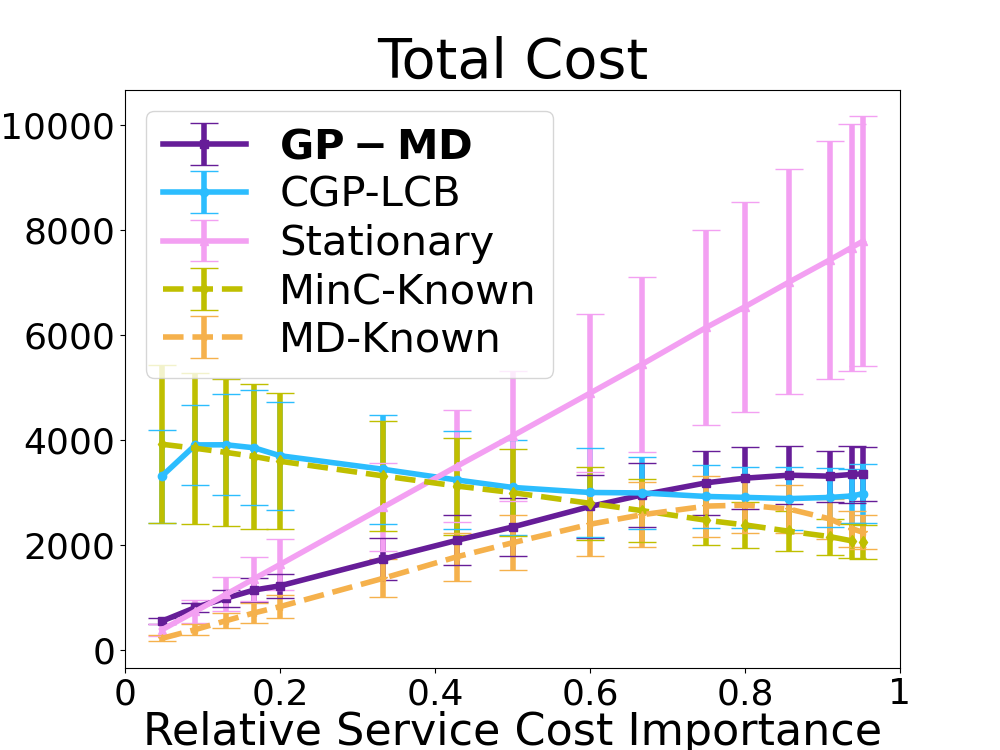}
        \caption{Varying $\rho$}
        \label{fig:fig16}
    \end{subfigure}
    \vspace{-1cm}
    \caption{\small Total and movement cost performance of algorithms on synthetic functions for varying importance of movement/service cost (i.e., different $\rho$ values). \algnm~outperforms \textsc{CGP-LCB} in terms of total incurred cost, and its performance closely follows one of the idealized benchmark \textsc{MD-Known}. \algnm~also minimizes the movement cost while \textsc{CGP-LCB} suffers from significant movements (\Cref{fig:fig14,fig:fig15}). The performance of \algnm~remains robust when the movement cost importance in the total cost objective diminishes (\Cref{fig:fig16}).\looseness=-1
    }
    \label{fig:figure}
    \vspace{-0.5cm}
\end{figure*}

\vspace{-1ex}
\section{Experiments}
\vspace{-1ex}
This section provides numerical results on synthetic and real-world data. We compare the performance of our \algnm~algorithm with the following baselines:
\begin{itemize}[noitemsep,topsep=0ex]
\item \textsc{Stationary} selects the stationary strategy $x_{h} = x_{0}$ for all $h$,
    \item \textsc{CGP-LCB}~\cite{krause2011contextual} neglects the movement cost and sets $x_{h}= \arg\min_x\, \lcb_{h}(x,e_{h})$ for all $h$,
    \item \septnm~assumes $f(\cdot)$ is known and chooses $x_{h}= \arg\min_x\, f(x,e_{h})$, and
    \item \textsc{MD-Known} assumes $f(\cdot)$ is known and runs mirror descent from \cite{coester2019pure} on %the actual cost 
    $f(\cdot,e_{h})$.
\end{itemize}
 \textsc{MD-Known} and $\septnm$ unrealistically assume that $f(\cdot)$ is known and can be seen as upper-bound for the achievable performance of $\algnm$ and \textsc{CGP-LCB}, respectively. We use the same constant value $\beta=2.0$ for the exploration parameter in both $\algnm$ and \textsc{CGP-LCB} (since the theoretical worst-case bounds are found to be overly pessimistic and can impede performance~\citep{Srinivas2010}).
We run the algorithms over a single episode (as done in \textsc{CGP-LCB}). We also discover that updating the confidence bounds after every timestep in \algnm~leads to better performance in practice.\looseness=-1 
%we have analyzed theoretically earlier where we considered a episodic setting and updated the confidence bounds only at the end of every episode.}  \looseness=-1

\textbf{Synthetic experiments.} We consider synthetic experiments, where the objective function is a random GP sample. The considered action space $\X$ is a subset of $[0,1]^{2}$ consisting of $400$ points that form the uniform grid, while the context space $\cE$ consists of $40$ contexts that are uniformly sampled from $(0,1)$. We sample objective function (i.e., actual cost) $f:\X \times\cE \to \mathbb{R}$ from a $GP(0,k)$, where $k$ is a squared exponential kernel with lengthscale parameter set to $l=0.2$. We use the Euclidean distance between the domain points as the movement cost, calculate the distance matrix (between every pair of points) and the average movement cost. We subtract the minimum value from $f$ and scale it such that the average function value is equal to the average movement cost. We also set the noise parameter to $1\%$ of the function range. We introduce the trade-off parameter $\rho \in \{0.25,0.5,1,2,4\}$ that only multiplies the service cost, i.e, $\rho f(x,e)$, but not the movement cost. This is to test the performance of the algorithms for varying importance of the service/movement costs.  For each $\rho$ we sample $25$ different functions and run the algorithms for $500$ timesteps wherein at each step the contexts are randomly sampled.\looseness=-1 

In \Cref{fig:fig11}-\Cref{fig:fig15}, we show the total cumulative cost as a function of timesteps for different $\rho$ values. Then, in~\Cref{fig:fig16}, we show the performance of the algorithms (for known kernel parameters) when run for $800$ timesteps for varying importance of the service and movement costs. In particular, we consider a convex combination of the service and movement costs, where we set the respective coefficients multiplying these two objectives as $\rho/(1 + \rho)$ and $1 /(1 + \rho)$.\looseness=-1

As shown in \Cref{fig:fig11}-\Cref{fig:fig12}, the performance of \algnm~is generally close to the one of the idealized, unrealistic benchmark \textsc{MD-Known}, which, as expected, performs the best. The stationary baseline performs comparably when $\rho$ is small, while its performance deteriorates for larger values. As expected, both \textsc{MinC-Known} and \textsc{CGP-LCB} incur higher total costs than \algnm~when the movement cost is of the higher or same relative importance as the service cost (i.e., $\rho \in \lbrace 0.5,1.0 \rbrace$), while the performance gap slowly decreases when the service cost becomes dominant ($\rho = 2.0$). In~\Cref{fig:fig14,fig:fig15}, we also show the corresponding movement costs, and observe that movement cost ignorant \textsc{CGP-LCB} incurs significant movement costs, while our \algnm~ successfully minimizes the movement costs. Finally, in ~\Cref{fig:fig16}, we observe that the performance of \algnm~ is robust, i.e., it clearly outperforms \textsc{CGP-LCB} whenever the movement cost dominates  the total cost objective, while its performance remains comparable to the one of \textsc{CGP-LCB} (that is built to minimize service cost) when the movement cost becomes dominated by the service cost. 

%\vspace{-1ex}
\subsection{Altitude Optimization in AWE Systems}
%\vspace{-1ex}
\label{sec:AO_in_AWE}
In airborne wind energy (AWE) systems, the turbine's operating altitude can be changed depending on the wind pattern. We follow the setup of \citet{baheri2017} that applied \textsc{CGP-LCB} \cite{krause2011contextual} which ignores movement-costs, to learn this control task. In this section, we use a dataset from \cite{bechtle2019airborne} which contains wind-speed information over various locations in Europe for a period ranging from 2011 to 2017, and also includes measurements at different altitudes per location. We consider the wind speed data from the second half of 2016. 
Our goal is to maximize the generated energy, while taking into account the energy loss due to moving the turbine from one altitude to another. \looseness=-1

We consider $25$ different altitudes (ranging from $10$m to $1600$m) as the action space and the context space to be hours in the day (i.e., $24$ values). We define our unknown service objective function to be $f(x,t)=\max_{x'}(E_{S}(x',t))- E_{S}(x,t)$ where $E_{S}(x,t)$ denotes the energy generated based on the windspeed at altitude $x$ and time $t$. Based on the discrete-time power generation formula from \citet{baheri2017} (Eq.~(10)), we have
\begin{equation}\label{eq: power objective function}
    E_{S}(x,t)=\big(c_{1}(\min\lbrace V_{w}(x,t),V_{r}\rbrace)^{3}-c_{2}V_{w}^{2}(x,t)\big)\Delta t.
\end{equation}
Here $V_{w}(x,t)$ denotes the windspeed at altitude $x$ and time $t$, and $V_{r}$ denotes the rated windspeed of the turbine. The constants $c_{1}=0.0579$ and $c_{2}=0.09$ are system dependent. %Here $x$ denotes the altitude and corresponds to our action space and $t$ denotes to the time of day and corresponds to our contexts $e$.
This corresponds to the energy generated at a particular altitude $x$ for $\Delta t$ time. Similarly to \citet{baheri2017}, we use $\Delta t = 60$ since we consider intervals of one hour length. Next, we define the movement cost to be the energy lost in changing altitude (from $x$ to $x'$):\looseness=-1
\begin{equation}
    E_{M}(x,x')=c_{3}V_{r}^{2}|x-x'|,
\end{equation}
where $c_{3}=0.15$ (see \Cref{sec:exptsupp} for more details).\footnote{According to the power equation from \cite{baheri2017}, $E_{M}(x,x')$ would depend on $V_{w}(x,t)$, whereas, we assume our movement cost is based on a fixed metric and is independent of contexts. Hence, we simply approximate $V_{w}(x,t)$ by $V_{r}$ and consider time-independent movement costs.}

We assume that a wind speed gauge is attached to the turbine, and the operator knows the wind speed at the current altitude. Hence, instead of directly learning $f(x,t)$, we learn $V_{w}(x,t)$ and use its confidence bounds to calculate the confidence bounds of $f(x,t)$. 
To learn $V_{w}(x,t)$, we normalize the inputs, and fit a GP with RBF kernel (lengthscale=3.67, outputscale=6.85 and noise parameter=2.73). 

\begin{figure*}[t!]
    \centering
    \begin{subfigure}[b]{.3\textwidth}
        \centering
        \includegraphics[width=\linewidth]{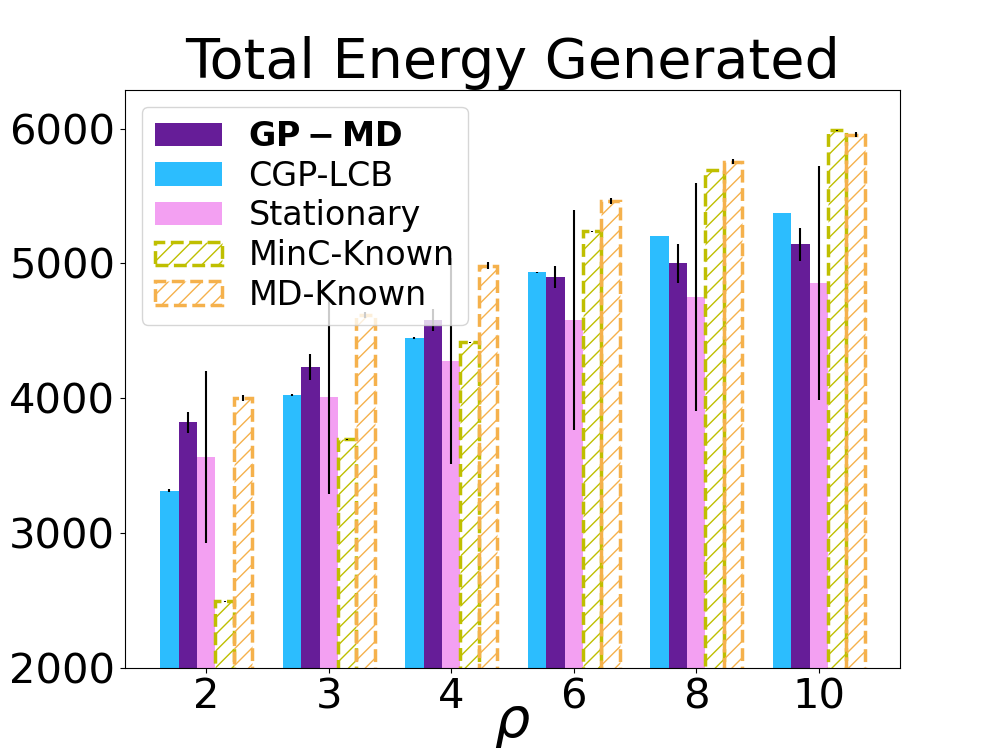}
        \caption{$\text{Lat.}=53, \text{Long.}=-10$}
        \label{fig:fig21}
    \end{subfigure}
    \hspace{1.5em}
    \begin{subfigure}[b]{.3\textwidth}
        \centering
        \includegraphics[width=\linewidth]{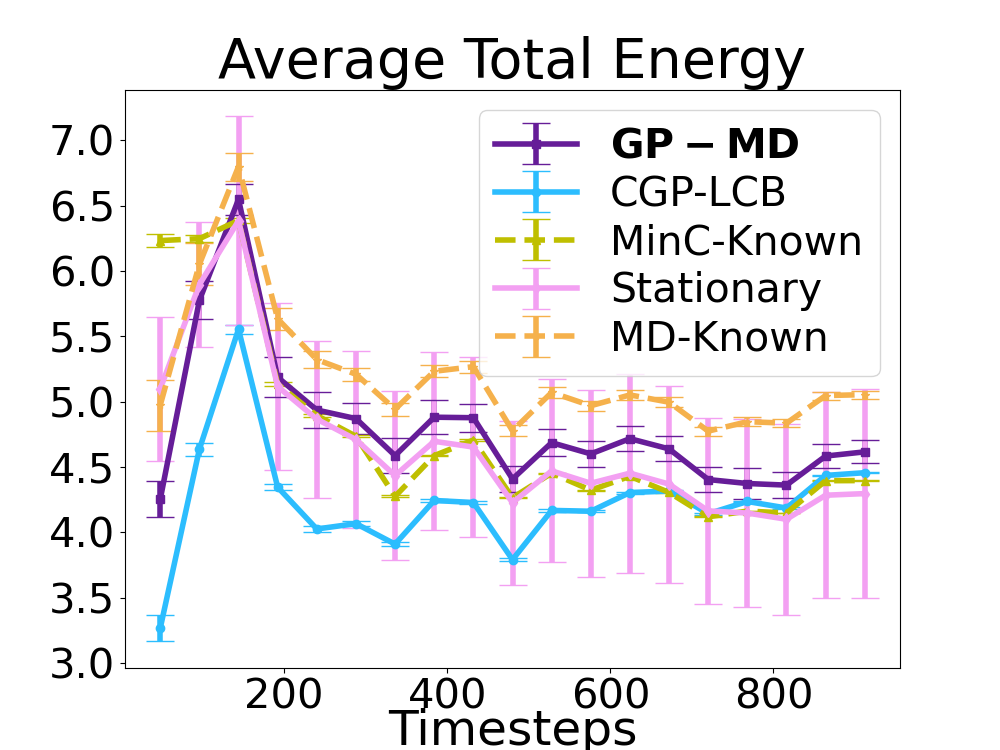}
        %\caption{Figure2}
        \caption{Total Energy ($\rho=4$)}
        \label{fig:fig22}
    \end{subfigure}
    \hspace{1.5em}
    \begin{subfigure}[b]{.3\textwidth}
        \centering
        \includegraphics[width=\linewidth]{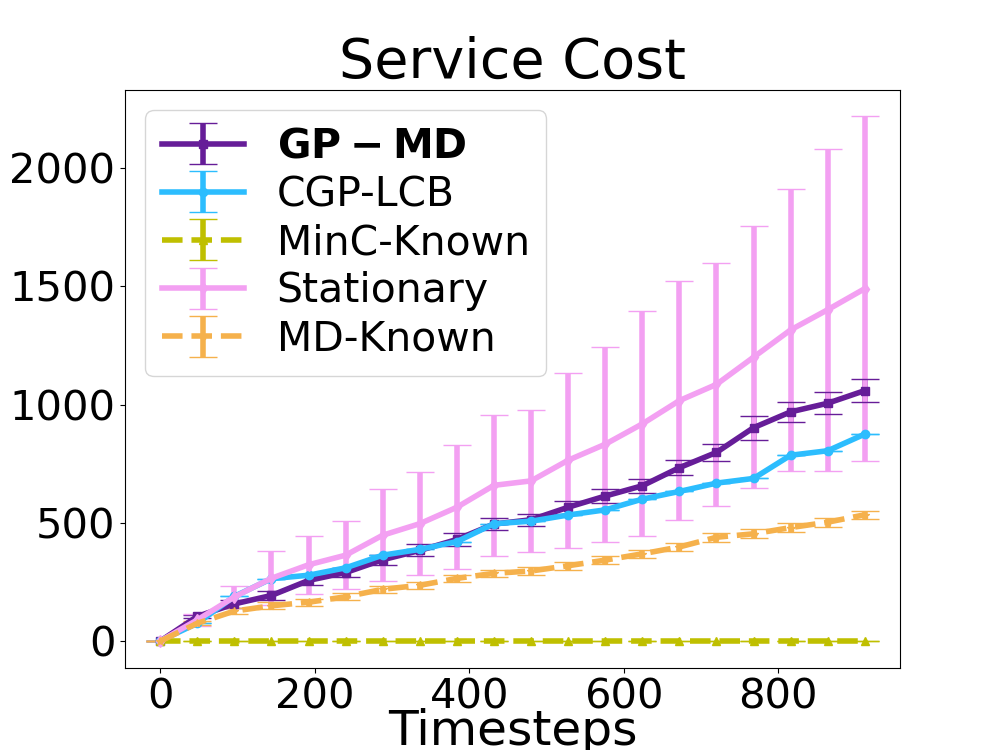}
        %\caption{Figure3}
        \caption{Service Cost ($\rho=4$)}
        \label{fig:fig23}
    \end{subfigure}
    \vspace{2em}
    \begin{subfigure}[b]{.3\textwidth}
        \centering
        \includegraphics[width=\linewidth]{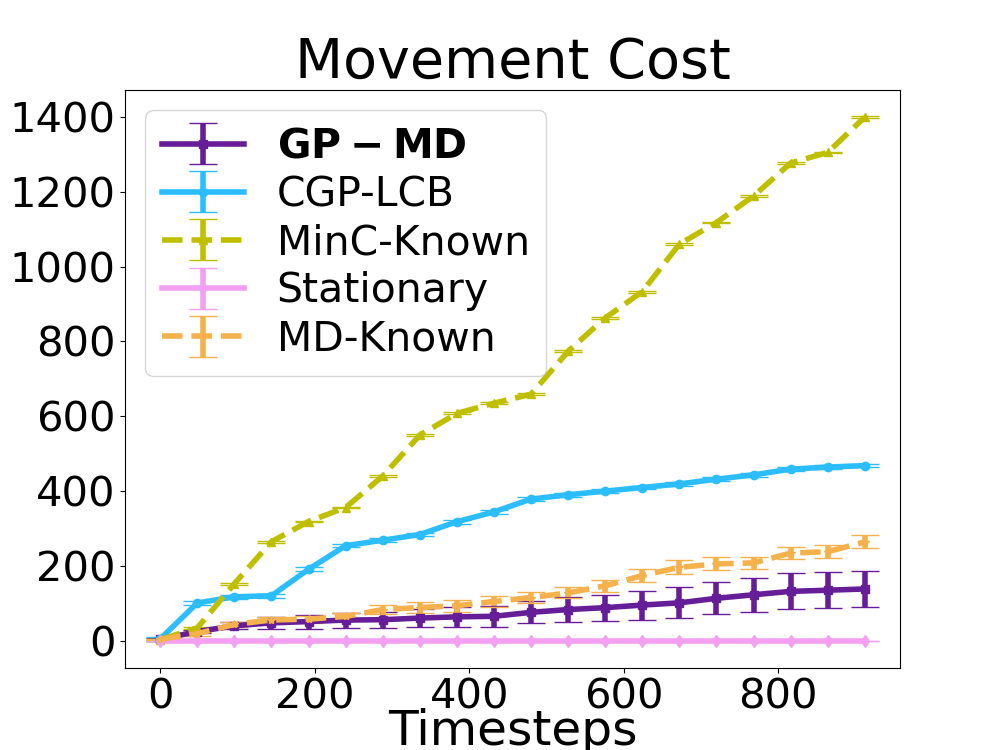}
        %\caption{Figure4}
        \caption{Movement Cost ($\rho=4$)}
        \label{fig:fig24}
    \end{subfigure}
    \hspace{1.5em}
    \begin{subfigure}[b]{.3\textwidth}
        \centering
        \includegraphics[width=\linewidth]{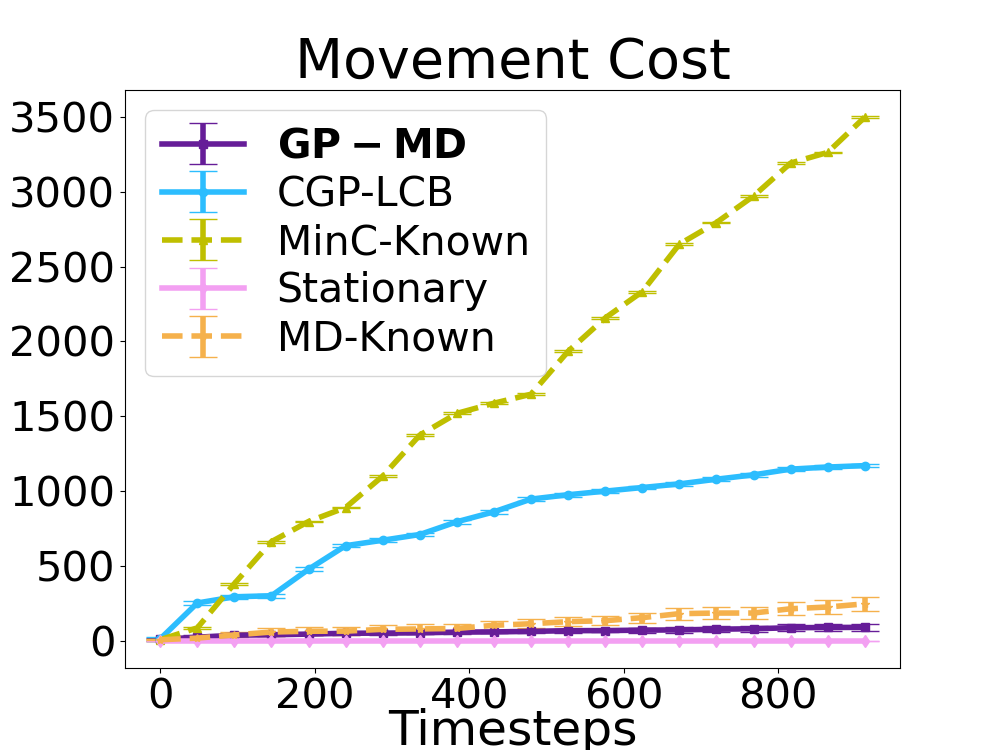}
        %\caption{Figure5}
        \caption{Movement Cost ($\rho=1$)}
        \label{fig:fig25}
    \end{subfigure}
    \hspace{1.5em}
    \begin{subfigure}[b]{.3\textwidth}
        \centering
        \includegraphics[width=\linewidth]{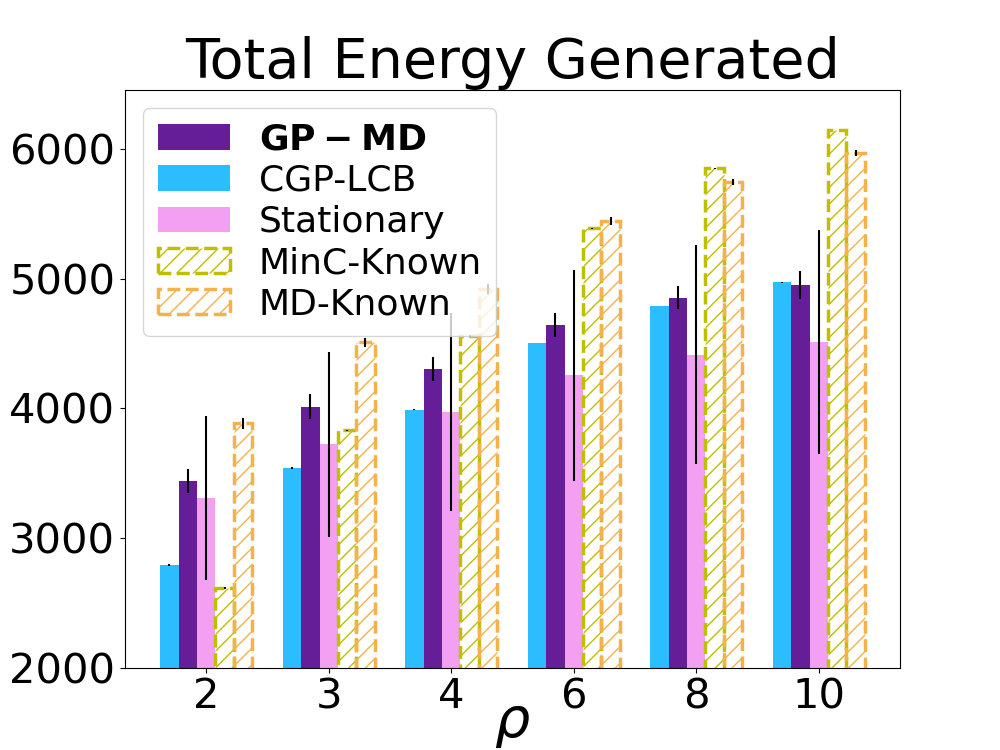}
        %\caption{Figure6}
         \caption{$\text{Lat.}=53, \text{Long.}=-4$}
        \label{fig:fig26}
    \end{subfigure}
%        \caption{Here $\alpha$ is defined as $\frac{\rho}{1+\rho}$ and cost values are divided by $1+\rho$ so as to make the costs for different $\rho$ comparable.}
    \vspace{-1cm}
    \caption{\footnotesize AWE altitude optimization task; \Cref{fig:fig21}: Total energy generated for $960$ hours based on the wind data at a single location (Latitude$=53$ and Longitude$=-10$). %for various values of tradeoff parameter $\rho$. %All values are divided by a factor of $(1+\rho)$*(minimum movement cost) for easy visualization of performance comparisons. 
    \algnm~outperforms previously used \textsc{CGP-LCB} (that optimizes for service costs only) for a range of $\rho$ values that favor the service against the movement cost. \Cref{fig:fig22}: The average total generated energy over $960$ hours. \Cref{fig:fig23,fig:fig24}: The service and movement costs for $\rho=4$.  \Cref{fig:fig25}: The movement costs for $\rho=1$. The movement energy loss is slightly lower for \algnm~as compared to $\rho=4$ due to higher importance towards movement cost reduction. \Cref{fig:fig26}: Same as \Cref{fig:fig21}, albeit by using wind data from a different location (Latitude$=53$, Longitude$=-4$).}
    \label{fig:figure2}
    \vspace{-0.5cm}
\end{figure*}

We run the algorithms for different $\rho$ for $960$ timesteps, where again $\rho$ is used to multiply $E_{S}$. For each $\rho$, the algorithms were initiated with every possible starting point (25 different altitudes), and ran for 3 iterations. Based on this we plot the total energy generated w.r.t.~varying $\rho$ in \Cref{fig:figure2}.
In \Cref{fig:fig21,fig:fig26}, we show the performance of the algorithms at two different locations (we also consider additional locations and time periods in \Cref{sec:exptsupp}). We use different values of $\rho > 1$ to show the robustness of our algorithm (as $\rho$ increases, the importance of the service cost w.r.t. the movement cost in the overall objective increases). Our algorithm outperforms \textsc{CGP-LCB} for a range of $\rho$ values.  As $\rho$ keeps increasing, we observe that \textsc{MinC-Known} closes the performance gap to \textsc{MD-Known}, and the same is happening with \textsc{CGP-LCB} w.r.t.~\textsc{GP-MD}. In \Cref{fig:fig22}, we focus on a particular $\rho=4$, and notice that $\algnm$ performs better than \textsc{CGP-LCB} and \textsc{Stationary} algorithm at this location. In \Cref{fig:fig23}, we plot the service cost and observe that both learning algorithms $\algnm$ and \textsc{CGP-LCB} have lower service cost than the \textsc{Stationary} baseline. We also note that due to the implicit service cost definition, the \textsc{MinC-Known} baseline achieves zero service cost. In \Cref{fig:fig24,fig:fig25}, we compare the movement energy loss for $\rho=4$ and $\rho=1$. As expected, $\rho=1$ results in slightly lower $\algnm$~movement energy loss due to the tradeoff shifting towards the movement cost.

\vspace{-1ex}
\section{Conclusions}
\label{sec: conclusions}
\vspace{-1ex}
We have considered the problem of optimizing an unknown cost function subject to time-varying contextual information, as well as \emph{movement costs} of changing the selected action from round to round. Our problem formulation is motivated by Airborne Wind Energy systems, where one seeks to optimize the operating altitude of the wind turbine to maximize the amount of generated energy. We propose a novel algorithm, $\algnm$, which makes use of GP confidence bounds and employs the mirror descent techniques from \cite{coester2019pure} for solving MTS problems. We analyze the theoretical performance of our algorithm by providing a rigorous regret bound. Moreover, we demonstrate its performance in synthetic experiments and on an AWE application by using real-world data. $\algnm$  carefully trades off service and movement costs while at the same time learning about the unknown objective function and yielding improved performance (i.e., generating more energy) compared to the considered baselines. Our setup and analysis open up multiple interesting directions for further exploration. For instance, an extension to continuous action spaces via discretization arguments is an immediate direction for future work. Another interesting direction is to analyze the single-episode setting and obtain general sublinear regret guarantees.% We analyze this setting in \Cref{sec: single-episode} considering batch updates of the GP confidence bounds and obtain regret guarantees that are sublinear for the squared exponential kernel. %Finally, due to the well-known limitations of the GPs, it would be interesting to explore if it can be replaced with neural networks or other non-linear function spaces.}
\looseness=-1

\section{Acknowledgements}
The authors would like to thank~James R.~Lee and~Christian Coester for the various discussions regarding their paper \cite{coester2019pure} during the course of this work. This project has received funding from the European Research Council (ERC) under the European Unions Horizon 2020 research and innovation programme grant agreement No 815943.

\bibliography{cgpref}

\onecolumn
\newpage
{\centering
    {\huge \bf Supplementary Material}
    
    {\Large \bf Movement Penalized Bayesian Optimization \\ with Application to Wind Energy Systems \\ [2mm] {\normalsize \bf {} \par }  
}}
\appendix
\section{Metrical Task Systems (MTS)}
\label{sec: MTS}
Let $(\mathcal{X},d)$ be a finite metric space  with $|\mathcal{X}|=n>1$ as defined in \Cref{sec: Problem Formulation}. Henceforth, we denote the points in $\X$ as $\lbrace x^{1},\dots,x^{n}\rbrace $. 
The MTS problem runs over a single episode of horizon $T$. At every time instant $1\leq t\leq T$, the learner receives a non-negative cost function over $\mathcal{X}$, $c_{t}:\X\rightarrow \mathbb{R}_{+}$
corresponding to each point in $\mathcal{X}$. The goal of any online optimization algorithm in this setting is to choose $x_{t}\in \X$ such that both the cost incurred over the rounds and sum over the distances between its subsequent decisions as measured by $d(x_{t},x_{t-1})$ is minimized. We do not include episode $m$ in the variables' subscripts as it runs over a single episode and instead include the time horizon $T$. Hence, $D_{T}=\{x_{1},x_{2},\dots, x_{T}\}$ denotes the action sequence of length $T$ outputted by an algorithm and $S_{T}(D_{T})$ and $M_{T}(D_{T})$ (\Cref{eq: formulate}) denote the corresponding service and movement costs respectively.  Then the total cost incurred by such an algorithm for initial state $x_{0}$ up to a time horizon $T$ is $$\mathrm{cost}_{T}(D_{T})=\sul_{t=1 }^{T}c_{t}(x_{t})+d(x_{t},x_{t-1}).$$ 
Next, we recall some notions about competitive ratio for MTS from \cite{coester2019pure} which are useful to prove our regret guarantees in \Cref{sec: regret proof}. Here, $D_{T}^{*}$ denotes the offline optimal sequence which minimizes $\mathrm{cost}_{T}(D_{T})$. Here we note that this offline optimal sequence $D_{T}^{*}$ depends on the initial state $x_{0}$.\\
\textbf{Competitive Ratio}:
If there exist constants $\alpha,\beta$ such that for every cost sequence $(c_{t})_{t=1}^{T}$, arbitrary initial state $x_{0}\in \X$ and distance metric $d(\cdot,\cdot)$,
\begin{eqnarray*}
\mathrm{cost}_{T}(D_{T})&\leq&\alpha \mathrm{cost}_{T}(\D_{T}^{*})+\beta,
\end{eqnarray*}
then the algorithm is $\alpha$-competitive.\\
\textbf{Refined Competitive Ratio Guarantees}:
If there exist constants $\alpha,\alpha',\beta,\beta'$ such that for every cost sequence $(c_{t})_{t=1}^{T}$, arbitrary initial state $x_{0}\in \X$ and distance metric $d(\cdot,\cdot)$,
\begin{align}\label{eq: service guarantee}
S_{T}(D_{T})&\leq\alpha \mathrm{cost}_{T}(\D_{T}^{*})+\beta,\\
\label{eq: movement guarantee}
M_{T}(D_{T})&\leq\alpha' \mathrm{cost}_{T}(\D_{T}^{*})+\beta',
\end{align}
then the algorithm is $\alpha$-competitive for service costs and $\alpha'$-competitive for movement costs.

\section{Hierarchically Separated Tree (HST) Metric}
\label{sec: HST}
\begin{figure}
\begin{center}
\begin{tikzpicture}
[
    level 1/.style = {black, sibling distance = 8cm},
    level 2/.style = {black, sibling distance = 4 cm},
    level 3/.style = {black, sibling distance = 2 cm},scale=0.9, every node/.style={scale=0.9}
]
 
\node {$r$}
    child {[fill] circle (2pt)
    child {[fill] circle (2pt)
    child {[fill] circle (2pt)}
    child { node[draw] {$l_{1}$}edge from parent [red]node[right]{$w_{l_{1}}$}}edge from parent [red]node[right]{$w_{par(l_{1})}$}}
    child {[fill] circle (2pt)
    child {[fill] circle (2pt)}
    child {  node[draw] {$l_{2}$}edge from parent [red]node[right]{$w_{l_{2}}$}}edge from parent [red]node[right]{$w_{par(l_{2})}$}}}
    child {[fill] circle (2pt)
    child {[fill] circle (2pt)
    child {[fill] circle (2pt)}
    child {[fill] circle (2pt)}}
    child {[fill] circle (2pt)
    child {[fill] circle (2pt)}
    child {[fill] circle (2pt)}}
    };
    %child {[fill] circle (2pt)
    %child {[fill] circle (2pt)
    %child {node {great-grandchild}}}
    %child {[fill] circle (2pt)}
    %edge from parent node [right] {x}};
 
\end{tikzpicture}
\end{center}
\caption{$
    d_{\T}(l_{1},l_{2})=w_{l_{1}}+w_{par(l_{1})}+w_{l_{2}}+w_{par(l_{2})}.
$}\label{fig:distance example}
\end{figure}
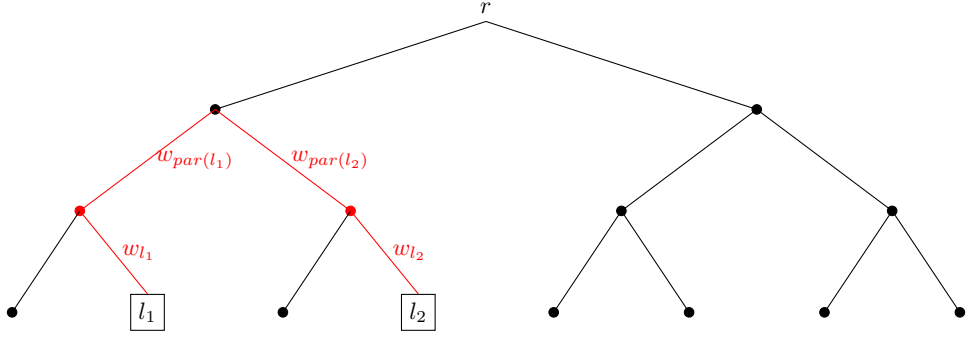

We define the tree $\mathcal{T}=(V,E)$ with the root vertex being $r$ and weight corresponding to each vertex $ v \in V$ being $w_{v}$. Let $\cL$ denote the set of leaves in this tree $\T$. Consider the case when the weights are non-increasing while moving from root to any leaf. We assign the edge from any vertex $u$ to its parent $par(u)$ the weight $w_{u}$. \\
\textbf{Distance Function}:
%\label{sec: Tree metric}
We define the distance metric $d_{\T}(l,l')$ between any two leaves $l,l'$ in the tree as the weighted length of the path from $l$ to $l'$. For example, as shown in \Cref{fig:distance example},
Then the space with states $\cL$ and distance metric $d_{\T}$ is defined as the HST metric space denoted by $(\cL,d_{\T})$ (\cite{coester2019pure}).
\subsection{$\tau$-HST Metric}
\label{sec:hst_metric_sec}
In an HST Metric space if the weights are exponentially decreasing, i.e., $w_{u}\leq w_{par(u)}/\tau$ then we call it a $\tau$-HST Metric. Such metric spaces are of particular interest due to a result from \cite{bartal1996probabilistic} which states 
that any online algorithm which is $\mathcal{O}(g(n))$-competitive for $\tau$-HST metric space is $\mathcal{O}(g(n)\log{} n)$-competitive for arbitrary $n$-point metric spaces where $g(n)$ is some function on $n$.
\subsection{FRT Algorithm}
\label{sec: FRT}
 The FRT algorithm from \cite{fakcharoenphol2004tight} is a randomized algorithm and outputs a tree $\T$ whose leaves correspond to points in $\X$ but the tree distance between any two points (leaves) $x^{i},x^{j} \in \X$ (or $\cL$) is $d_{\T}(x^{i},x^{j})$ and satisfies for any $x^{i},x^{j}\in \X$ 
\begin{align}
\label{eq: approximation hst-1}
\mathbb{P}[d_{\T}(x^{i},x^{j})&\geq d(x^{i},x^{j})]=1,\\
\label{eq: approximation hst-2}
\E_{\T}\left[d_{\T}(x^{i},x^{j})\right]&\leq\mathcal{O}(\log n)d(x^{i},x^{j}),   
\end{align}
 where $d(\cdot,\cdot),$ is the original metric
and expectation is w.r.t. the random tree $\T$ generated by the FRT algorithm \cite{fakcharoenphol2004tight}.
\section{Action Representation}
This section is intended to explain in detail about the randomized algorithm setup for MTS as elucidated in \cite{coester2019pure} and \cite{bubeck2021metrical}. But instead of using the cost sequence $(c_{t})_{t=1}^{T}$, we explain in terms of $f(\cdot,e_{t})$ which is more relevant to our setting. In our proof, we only consider episodic expectation (see \Cref{eq: regret conversion}) and want to understand how the randomized algorithm evolves within a single episode. Hence similar to MTS setup from \citet{coester2019pure} we detail this randomization section for a single episode. 
\subsection{Action Randomization}
Let $\mathcal{P}(\mathcal{X})$ be the set of probability measures supported on $\mathcal{X}$. For $\mu, \nu \in \mathcal{P}(\mathcal{X})$ denote $\mathbb{W}^{1}(\mu, \nu)$ as the Wasserstein-1-distance between $\mu$ and $\nu$ defined based on $d(\cdot,\cdot)$. A randomized online algorithm at time $t$ outputs a random action $x_{t}\sim p_{t}$ where probability distribution $p_{t}\in \mathcal{P}(\X)$. As defined earlier, we denote this random output sequence as $D_{T}=\{x_{1},x_{2},\dots, x_{T}\}$. In this randomized setting it is more intuitive to consider the \emph{expected cost}. 

The \emph{expected movement cost} for distance metric $d(\cdot,\cdot)$ is  defined as follows:
$$\E\left[ M_{T}(D_{T})\right]=\E\left[\sul_{t=1}^{T}d(x_{t-1},x_{t})\right]=\sul_{t=1}^{T}\E[d(x_{t-1},x_{t})].$$
Here note that each term in the sum $\E[d(x_{t-1},x_{t})]$ is expectation w.r.t. a joint distribution of $(x_{t-1},x_{t})$. Under certain assumptions on the sampling process (described in \Cref{sec: joint}), the joint distribution becomes
\begin{equation}
\label{eq: Wasserstein formulation-1}
\E\left[ M_{T}(D_{T})\right]=\sul\limits_{t=1}^{T}\mathbb{W}^{1}(p_{t-1},p_{t}),    
\end{equation}
And the \emph{expected service cost} is then, 
\begin{equation}
\label{eq: service cost inner product}
\E\left[ S_{T}(D_{T})\right]=\E\left[\sul_{t=1}^{T}f(x_{t},e_{t})\right]=\sul_{t=1}^{T}\langle f(\cdot,e_{t}),p_{t}\rangle,
\end{equation}
\begin{equation}\label{eq: Wasserstein Total Cost}
    \E\left[\mathrm{cost}_{T}(D_{T})\right]=\sul_{t=1}^{T}\mathbb{W}^{1}(p_{t-1},p_{t})+\sul_{t=1}^{T}\langle f(\cdot,e_{t}),p_{t}\rangle.
\end{equation}
An randomized online algorithm is said to be \emph{$\alpha$-competitive} if for all $x_{0}\in \X$ and any $f$, it outputs a sequence $D_{T}$ whose expected cost for some $\beta > 0$ satisfies the following condition w.r.t. the offline optimal sequence $D_{T}^{*}$ for some metric $d(\cdot,\cdot)$:
$$\E\left[\mathrm{cost}_{T}(D_{T})\right]\leq \alpha\mathrm{cost}_{T}(D_{T}^{*})+\beta.$$
\subsection{Sampling from Joint Distribution}
\label{sec: joint}
Since our goal is to minimize the total cost we choose a joint distribution $\zeta_{t}\in \Pi(p_{t-1},p_{t})$ which minimizes the expected cost as follows:
$$\E_{\zeta_{t}}[d(x_{t-1},x_{t})]=\inf\limits_{\zeta\in \Pi(p_{t-1},p_{t})}\E_{\zeta}[d(U_{t-1},U_{t})], $$
where $\Pi(p_{t-1},p_{t})$ denotes the set of all random variables $(U_{t-1},U_{t})$ whose marginals are $p_{t-1}$ and $p_{t}$, respectively. By the definition of Wasserstein-1 distance where the Wasserstein cost function is assumed to be $d(\cdot,\cdot)$ we have,

\begin{equation}\label{eq: wasserstein definition}
    \inf\limits_{\zeta\in \Pi(p_{t-1},p_{t})}\E_{\zeta}[d(U_{t-1},U_{t})]=\mathbb{W}^{1}(p_{t-1},p_{t}).
\end{equation}
Hence, we want to ensure that the subsequent actions in \cref{alg: mgpbo} $(x_{t-1},x_{t})$ follows the distribution $\zeta_{t}$ making our total movement cost 
\begin{equation}
\label{eq: Wasserstein formulation-2}
\E\left[ M_{T}(D_{T})\right]=\sul\limits_{t=1}^{T}\mathbb{W}^{1}(p_{t-1},p_{t}).   
\end{equation}
In order to achieve this, we take the following steps. We first note that the initial action $x_{0}$ is given. Hence, after obtaining $p_{1}$ using Mirror Descent procedure as done in Line-11 of \Cref{alg: mgpbo}, we estimate $\zeta_{1}$, then calculate the conditional distribution $\zeta_{1}(U_{1}|U_{0}=x_{0})$, and sample $x_{1}$ from this conditional distribution. At any future time instant $t$, after obtaining $p_{t}$, we repeat this process and calculate the conditional distribution $\zeta_{t}(U_{t}|U_{t-1}=x_{t-1})$ and sample $x_{t}$ from it (Line 12 of \Cref{alg: mgpbo}). In this way we ensure that at each time instant $t$, $(x_{t-1},x_{t})$ is sampled from $\zeta_{t}$ and hence attaining the movement cost as sum of Wasserstein-1 distances.

\subsection{Randomized Action Representation in $\tau$-HST}
\label{sec: random state appendix}
In this section, we explain the randomized action representation in $\tau$-HST space introduced in \Cref{sec: Alg and guarantee} in further detail. Also, we show the Wasserstein-1 distance (\Cref{eq: wasserstein definition}) can be simplified in the $d_{\T}$-metric when the actions are leaves of $\T$ as done in \cite{bubeck2021metrical} and \cite{coester2019pure}. 

From the analysis in the previous section, in order to calculate the movement cost, we need to compute the Wasserstein-1 distances between 2 probability distributions over the leaves of the constructed tree. As we now consider $\tau$-HST metric space, the distance metric is $d_{\T}$ and Wasserstein-1 distance is denoted as $\mathbb{W}^{1}_{\T}(\cdot,\cdot)$. We first recall the following  representation of the randomized action described in \Cref{sec: Alg and guarantee} from \cite{bubeck2021metrical}. This will be useful in both the Wasserstein-1 distance calculation and in \Cref{alg: mgpbo}.

Let $\T=(V,E)$ be a tree with vertices $V$, edges $E$, root $r$ and leaves $\cL$. We define a convex polytope on the space $\mathbb{R}_{+}^{V}$    $$K_{\T}:=\Big\{z\in\rv: z_{r}=1,z_{u}=\sul_{\nu\in\mathcal{C}(u)}z_{\nu} \quad \forall u\in V \backslash \cL\Big\},$$
where  $\mathcal{C}(u)$ denotes the children of $u$. Note that by the above definition for any $z\in K_{\T}$, it holds that $$\sul_{l\in \cL}z_{l}=1 .$$ 
And the Wasserstein-1 distance between 2 random actions in $(\cL,d_{\T})$ specified by the probability distributions $l(z)$ and $l(z')$ is as follows:
\begin{equation}
\label{eq: MTS Wasserstein}
\mathbb{W}^{1}_{\T}(l(z),l(z')):=\sul_{u\in V}w_{u}|z_{u}-z'_{u}| =||z-z'||_{l_{1}(w)}.
\end{equation}

where $w_{u}$ are the weights of the edges from $u$ to $par(u)$ in $\T$. Hence the total cost w.r.t. $d_{\T}(\cdot,\cdot)$-metric denoted by $\mathrm{cost}_{T}^{\T}(D_{T})$ (\Cref{eq: Wasserstein Total Cost}) now becomes
\begin{equation}
\E\left[\mathrm{cost}_{T}^{\T}(D_{T})\right]=\sul_{t=1}^{T}||z_{t-1}-z_{t}||_{l_{1}(w)}+\\\sul_{t=1}^{T}\langle f(\cdot,e_{t}),l(z_{t})\rangle. 
\end{equation}

Hence, for the $\tau$-HST metric space $(\cL,d_{\T})$, where leaves correspond to actions, $z$ defines a probability distribution over all the states. Each entry $z_u$ represents the probability that the selected action $x$ belongs to the leaves of the subtree rooted at $u$, i.e., $z_{u}=\pr(x\in \cL(u))$. Also,  We note that $z$ is completely defined when all $z_{l} \in \cL$ is provided. And for a deterministic state  $x\in\X$, the corresponding  state in $K_{\T}$ is
\begin{equation}\label{eq: z initialize-1}
   z_{l}=\left\{\begin{array}{lr}
        1, & \text{for } l=x\\
        0, & \text{for } l\neq x\\
        \end{array}\right\} \quad \forall l\in\cL, \qquad 
     z_{u}=\left\{\begin{array}{lr}
        1, & \text{for } x\in\cL(u)\\
        0, & \text{for } x\notin \cL(u)\\
        \end{array}\right\} \quad \forall u\in V\backslash \cL
\end{equation}
We can visualize this polytope with the example figure:\\
\begin{tikzpicture}
[
    level 1/.style = {black, sibling distance = 8cm},
    level 2/.style = {black, sibling distance = 4 cm},
    level 3/.style = {black, sibling distance = 2 cm}, ,scale=0.9, every node/.style={scale=0.9}
]
\node {$z(r)=1$}
    child {node[draw] {$z_{1}+z_{2}+z_{3}+z_{4}$}
    child {node[draw] {$z_{1}+z_{2}$}
    child {node[draw] {$z_{1}$}}
    child { node[draw] {$z_{2}$}}}
    child {node[draw] {$z_{3}+z_{4}$}
    child {node[draw] {$z_{3}$}}
    child {  node[draw] {$z_{4}$}}}}
    child {node[draw] {$z_{5}+z_{6}+z_{7}+z_{8}$}
    child {node[draw] {$z_{5}+z_{6}$}
    child {node[draw] {$z_{5}$}}
    child { node[draw] {$z_{6}$}}}
    child {node[draw] {$z_{7}+z_{8}$}
    child {node[draw] {$z_{7}$}}
    child {  node[draw] {$z_{8}$}}}}
    ;
    %child {[fill] circle (2pt)
    %child {[fill] circle (2pt)
    %child {node {great-grandchild}}}
    %child {[fill] circle (2pt)}
    %edge from parent node [right] {x}};
 
\end{tikzpicture}\\
\subsection{Action Representation using Conditional Probabilities}
\label{sec: conditional random state appendix}
In \Cref{sec: Alg and guarantee}, we provided an intutive understanding of how the algorithm works based on the calculation of $q^{(u)}$ corresponding to each internal vertex $u$ (\Cref{eq:MD_update}). We also discussed how this can be viewed as a conditional probability $\pr(x\in \cL (\nu)|x\in \cL(u))$ for $\nu\in \mathcal{C}(u)$ (children of $u$) and how it can be used to calculate the actual probabilities over the actions-$l(z)$ (\Cref{eq: Q-Z relation}).

We first begin with a visualization of this $q$ based on the visualization in \Cref{sec: random state appendix} and \Cref{eq: Q-Z relation}:\\
\\
\begin{tikzpicture}
[
    level 1/.style = {black, sibling distance = 8cm},
    level 2/.style = {black, sibling distance = 4 cm},
    level 3/.style = {black, sibling distance = 2 cm},scale=0.9, every node/.style={scale=0.9}
]
\node {}
    child {node[draw] {$z_{1}+z_{2}+z_{3}+z_{4}$}
    child {node[draw] {$\frac{z_{1}+z_{2}}{z_{1}+z_{2}+z_{3}+z_{4}}$}
    child {node[draw] {$\frac{z_{1}}{z_{1}+z_{2}}$}}
    child { node[draw] {$\frac{z_{2}}{z_{1}+z_{2}}$}}}
    child {node[draw] {$\frac{z_{3}+z_{4}}{z_{1}+z_{2}+z_{3}+z_{4}}$}
    child {node[draw] {$\frac{z_{3}}{z_{3}+z_{4}}$}}
    child {  node[draw] {$\frac{z_{4}}{z_{3}+z_{4}}$}}}}
     child {node[draw] {$z_{5}+z_{6}+z_{7}+z_{8}$}
    child {node[draw] {$\frac{z_{5}+z_{6}}{z_{5}+z_{6}+z_{7}+z_{8}}$}
    child {node[draw] {$\frac{z_{5}}{z_{5}+z_{6}}$}}
    child { node[draw] {$\frac{z_{6}}{z_{5}+z_{6}}$}}}
    child {node[draw] {$\frac{z_{7}+z_{8}}{z_{5}+z_{6}+z_{7}+z_{8}}$}
    child {node[draw] {$\frac{z_{7}}{z_{7}+z_{8}}$}}
    child {  node[draw] {$\frac{z_{8}}{z_{7}+z_{8}}$}}}};
    %child {[fill] circle (2pt)
    %child {[fill] circle (2pt)
    %child {node {great-grandchild}}}
    %child {[fill] circle (2pt)}
    %edge from parent node [right] {x}};
 
\end{tikzpicture}\\
\\
We explain why working in terms of $q$ rather than $z$ is beneficial. The Bregman divergence $D^{(u)}$ in the mirror descent update (\Cref{eq:MD_update}) uses a potential function and the authors of \cite{coester2019pure} observed that the conditional probability based potential function imitates the weighted entropy of the probability distribution over the leaves of a $\tau$-HST tree. The intuition for the above statement from \cite{coester2019pure} is described next.

We first recall $Q_{\T}$ and  $Q_{\T}^{(u)}$ from \Cref{sec: Alg and guarantee}. $Q_{\T}$ is the set of valid conditional probabilities
\begin{equation}\label{eq: Q_T-definition}
    Q_{\T}:=\Big\{q\in \mathcal{R}_{+}^{|V\backslash r|}:  \sul_{\nu\in\mathcal{C}(u)}q_{\nu} = 1 \quad \forall u\in V \backslash \cL\Big\}.
\end{equation}
 Moreover, given $q\in Q_{\T}$ we define the vector $q^{(u)}:=[ q_{\nu}, \: \nu \in \mathcal{C}(u)]$ as the conditional distribution over children of $u$, and let $Q_{\T}^{(u)}$ be the set of all valid distributions $q^{(u)}$.
 
We define the potential function $\Phi^{(u)}$ used for Bregman Divergence $D^{(u)}$ in the mirror descent update (\Cref{eq:MD_update}) for $q\in Q_{\T}$ and the corresponding $q^{(u)}\in Q_{\T}^{(u)}$ for a particular vertex $u\in V$ as follows:
\begin{equation}\label{eq: Potential Function}
\Phi^{(u)}(q^{(u)}):=\frac{1}{\kappa}\sul_{\nu\in\mathcal{C}(u)}\frac{w_{\nu}}{\eta_{\nu}}(q_{\nu}^{(u)}+\delta_{\nu})\log(q_{\nu}^{(u)}+\delta_{\nu}),    
\end{equation}
where\begin{align*}
\theta_{u}&:=\frac{|\cL(u)|}{|\cL(par(u))|},\\
\eta_{u}&:=1+\log(1/\theta_{u}),\\
\delta_{u}&:=\frac{\theta_{u}}{\eta_{u}},\\
\kappa&\geq 1\quad (\text{fixed constant for all }u).
\end{align*}

\textbf{Potential Function Intuition.} In this section we elucidate how the potential function defined above \Cref{eq: Potential Function} can be viewed as an approximate weighted entropy of a probability distribution over the leaves of a $\tau-$HST tree. Recalling our definitions, we have  $\T=(V,E)$, a tree with vertices $V$, edges $E$ and leaves $\cL$. $Y$ is a random variable with support $\cL$. Let $\varepsilon_{u}$ denote the event that $\{ Y\in \cL(u)\}$ where $\cL(u)$ is the set of leaves under the tree rooted at $u$.  For example, from the picture below, $\cL(u)=\{l_{1},l_{2},l_{3},l_{4}\}$\\
\begin{tikzpicture}
[
    level 1/.style = {black, sibling distance = 8cm},
    level 2/.style = {black, sibling distance = 4 cm},
    level 3/.style = {black, sibling distance = 2 cm},scale=0.9, every node/.style={scale=0.9}
]
 
\node {$r$}
    child {node[draw]{$u$} 
    child {[fill] circle (2pt) edge from parent [red]
    child { node[draw] {$l_{1}$}edge from parent [red]}
    child { node[draw] {$l_{2}$}edge from parent [red]}}
    child {[fill] circle (2pt)
    child { node[draw] {$l_{3}$}edge from parent [red]}
    child {  node[draw] {$l_{4}$}edge from parent [red]}edge from parent [red]}}
    child {[fill] circle (2pt)
    child {[fill] circle (2pt)
    child {[fill] circle (2pt)}
    child {[fill] circle (2pt)}}
    child {[fill] circle (2pt)
    child {[fill] circle (2pt)}
    child {[fill] circle (2pt)}}
    };
    %child {[fill] circle (2pt)
    %child {[fill] circle (2pt)
    %child {node {great-grandchild}}}
    %child {[fill] circle (2pt)}
    %edge from parent node [right] {x}};
 
\end{tikzpicture}\\
We calculate the entropy of such a random variable $Y$ and connect it with the potential function \Cref{eq: Potential Function}. Using the definition of entropy,
\begin{align}
H(Y)&=\sul_{l\in \cL}P[\varepsilon_{l}]\log\frac{1}{P[\varepsilon_{l}]}\\
&=\sul_{l\in \cL}P[\varepsilon_{l}]\log\frac{P[\varepsilon_{par(l)}]}{P[\varepsilon_{l}]P[\varepsilon_{par(l)}]}\\
&=\sul_{l\in \cL}P[\varepsilon_{l}]\log\frac{P[\varepsilon_{par(l)}]}{P[\varepsilon_{l}]}+\sul_{l\in \cL}P[\varepsilon_{l}]\log\frac{1}{P[\varepsilon_{par(l)}]}\\
&=\sul_{l\in \cL}P[\varepsilon_{l}]\log\frac{P[\varepsilon_{par(l)}]}{P[\varepsilon_{l}]}+\sul_{u\in par(\cL)}\sul_{l\in \mathcal{C}(u)}P[\varepsilon_{l}]\log\frac{1}{P[\varepsilon_{u}]}\\
\label{eq: entropy 1}
&=\sul_{l\in \cL}P[\varepsilon_{l}]\log\frac{P[\varepsilon_{par(l)}]}{P[\varepsilon_{l}]}+\sul_{u\in par(\cL)}P[\varepsilon_{u}]\log\frac{1}{P[\varepsilon_{u}]}\\
\label{eq: entropy 2}
&=\sul_{l\in \cL}P[\varepsilon_{l}]\log\frac{P[\varepsilon_{par(l)}]}{P[\varepsilon_{l}]}+\sul_{u\in par(\cL)}P[\varepsilon_{u}]\log\frac{P[\varepsilon_{par(u)}]}{P[\varepsilon_{u}]}+ \sul_{u\in par(par(\cL))}P[\varepsilon_{u}]\log\frac{1}{P[\varepsilon_{u}]}\\
\label{eq: entropy 3}
&=\sul_{u\in V\backslash\{r\}}P[\varepsilon_{u}]\log\frac{P[\varepsilon_{par(u)}]}{P[\varepsilon_{u}]}.
\end{align}
Here the transition from \Cref{eq: entropy 1}
to \Cref{eq: entropy 2} is obtained from simplifying $\sul_{u\in par(\cL)}P[\varepsilon_{u}]\log\frac{1}{P[\varepsilon_{u}]}$ as we simplified $\sul_{l\in \cL}P[\varepsilon_{l}]\log\frac{1}{P[\varepsilon_{l}]}$. Applying the previous argument recursively we obtain the final equation (\Cref{eq: entropy 3}). From \Cref{eq: entropy 3} the weighted version of the entropy can be defined as 
\begin{equation}\label{eq: weighted entropy}
H(Y,w)=\sul_{u\in V\backslash\{r\}}w_{u}P[\varepsilon_{u}]\log\frac{P[\varepsilon_{par(u)}]}{P[\varepsilon_{u}]}.\\
\end{equation}

Note that from definition of $z\in K_{\T}$ in \Cref{sec: random state appendix}, $z_{u}$ and $z_{par(u)}$ are analogous to $P[\varepsilon_{u}]$ and $P[\varepsilon_{par(u)}]$ respectively. Using this and a negative weighted version of \Cref{eq: weighted entropy} (inversion inside $\log$) the authors of \cite{coester2019pure} define the following regularizer for $z\in K_{\T}$.
\begin{equation}\label{eq: regularizer}
\Phi(z):=\sul_{u\in V\backslash \{r\}}\frac{w_{u}}{\eta_{u}}(z_{u}+\delta_{u}z_{par(u)})\log(\frac{z_{u}}{z_{par(u)}}+\delta_{u}).    
\end{equation}

Here $\delta_{u}$ signifies noise added to entropy calculation and ensures fast updates of $\frac{z_{u}}{z_{par(u)}}$ when it is close to zero. $\eta_{u}$ signifies the relative importance given to present costs at time $t$ vs past costs observed upto time $t-1$ (For further details refer to \citet[Section 1.3]{coester2019pure}).

From \Cref{eq: Q-Z relation}, we know that $q_{u}$ is analogous to $\frac{z_{u}}{z_{par(u)}}$. Based on \Cref{eq: regularizer} and replacing $q_{u}$, the authors of \cite{coester2019pure} define potential function for $q^{(u)}\in Q_{\T}^{(u)}$ corresponding to $q\in Q_{\T}$ for a particular $u$ as follows:
$$\Phi^{(u)}(q^{(u)}):=\frac{1}{\kappa}\sul_{\nu\in\mathcal{C}(u)}\frac{w_{\nu}}{\eta_{\nu}}(q_{\nu}+\delta_{\nu})\log(q_{\nu}+\delta_{\nu}).$$
\section{Algorithm}
\label{sec: Algorithm Appendix}
In this section we first provide some insights from \cite[Sectiom 2.1]{coester2019pure2} to solve the optimization problem in \Cref{eq:MD_update}. Then we proceed to practically illustrate the flow of the algorithm from leaves to root w.r.t. $q\in Q_{\T}$ calculations using \Cref{eq:MD_update}.
\subsection{Divergence and Optimization Calculations}\label{app:divergence}
In order to solve the optimization problem \Cref{eq:MD_update}, we first need to calculate the Bregman Divergence $D^{(u)}$. The Bregman Divergence for some potential function $\Phi(\cdot)$ is defined as follows:
\begin{equation}\label{eq: bregman divergence}
     D_{\Phi}(y || x):= \Phi(y)-\Phi(x)-\langle\nabla\Phi(x), y-x\rangle.
\end{equation}
In our case, Bregman Divergence $D^{(u)}(\cdot||\cdot)$ calculates the divergence between two conditional probability vectors $q^{(u)},q'^{(u)}\in Q_{\T}^{(u)}$ defined over the children of vertex $u$ w.r.t. the potential function $\Phi^{(u)}$. Here $q,q'\in Q_{\T}$ (\Cref{eq: Q_T-definition}) and potential function $\Phi^{(u)}$ is as defined in \Cref{eq: Potential Function}. Hence we have,
$$D^{(u)}(q^{(u)} || q'^{(u)}):=\frac{1}{\kappa}\sul_{\nu\in\mathcal{C}(u)}\frac{w_{\nu}}{\eta_{\nu}}\left[(q_{\nu}+\delta_{\nu})\log(\frac{q_{\nu}+\delta_{\nu}}{q_{\nu}'+\delta_{\nu}})+q'_{\nu}-q_{\nu}\right].$$
Now the authors of \cite{coester2019pure} use KKT conditions and Lagrange multipliers to solve \Cref{eq:MD_update} by substituting the definition of Bregman Divergence in \Cref{eq: bregman divergence}.  It yields that the solution to \Cref{eq:MD_update} for $q'^{(u)}=q_{h}^{(u)}$, $q^{(u)}=q_{h-1}^{(u)}$ and $e=e_{h,m}$ satisfies 
\begin{equation}\label{eq: zero-derivative}
\nabla\Phi^{(u)}(q'^{(u)})=\nabla\Phi^{(u)}(q^{(u)})-\lcb_{m}^{(u)}(\cdot,e)-\beta^{(u)}-\alpha^{(u)}.    
\end{equation}

Here $\beta^{(u)}$ and $\alpha^{(u)}$ are the Lagrange multipliers for the constraints in $Q_{\T}^{(u)}$ to ensure that for any $q^{(u)}\in Q_{\T}^{(u)}$, $q^{(u)}$ is actually a probability vector and comprises of the following 2 constraints, $$ \sul_{\nu \in \mathcal{C}(u)}q_{\nu}^{(u)}=1\quad \quad \text{and} \quad q_{\nu}^{(u)}\geq 0 \quad \text{for} \quad  \nu \in \mathcal{C}(u).$$
%Then $$\lm^{(u)}=\beta^{(u)}\textbf{1}-\alpha^{(u)},$$
%where $\alpha^{(u)}=\{\alpha^{(u)}_{\nu}: \nu \in \mathcal{X}(u)\}$ corresponds to non-negativity constraint of $ q_{\nu}^{(u)}$ and $\beta^{(u)}$ corresponds to $\sul_{\nu \in \mathcal{C}(u)}q_{\nu}^{(u)}\geq1$. (the other side will trivially hold)\\

Now calculating the gradient of the potential function in \Cref{eq: Potential Function} we have, $$\left(\nabla\Phi^{(u)}(q^{(u)})\right)_{\nu}=\frac{1}{\kappa}\frac{w_{\nu}}{\eta_{\nu}}(1+\log(q_{\nu}+\delta_{\nu})).$$
Substituting this in \Cref{eq: zero-derivative}, the solution to \Cref{eq:MD_update} for $q'^{(u)}=q_{h}^{(u)}$, $q^{(u)}=q_{h-1}^{(u)}$ and $e=e_{h,m}$ will be 
\begin{equation} \label{eq: Mirror Descent}
q'^{(u)}_{\nu}=(q_{\nu}^{(u)}+\delta_{\nu})\exp\{\kappa \frac{\eta_{\nu}}{w_{\nu}}(\beta^{(u)}-(\lcb_{m}^{(u)}(\nu,e)-\alpha_{\nu}))\}-\delta_{\nu}.
\end{equation}
\Cref{eq: Mirror Descent} can be solved in polynomial time using interior point methods. But for practical purposes, we use projected gradient descent w.r.t. $\alpha$ to solve this problem.

\subsection{Illustration}
In this section we illustrate the subroutine in lines $6-9$ of \Cref{alg: mgpbo}. For the tree $\T$ with vertices $V$ and leaves $\cL$ shown in \Cref{fig:topotree}, let the topological ordering $\mathcal{OD}(V\backslash\cL)$ (Line 7 in \Cref{alg: mgpbo}) of internal vertices $V\backslash\cL$ be $\{u_{1},u_{2},u_{3},u_{4},u_{5},u_{6},u_{7}\}$. Note that the ordering ensures that every child in $\mathcal{T}$ occurs before its parent.
\begin{figure}
\begin{center}
\begin{tikzpicture}
[
    level 1/.style = {black, sibling distance = 6cm},
    level 2/.style = {black, sibling distance = 3 cm},
    level 3/.style = {black, sibling distance = 1.5 cm},scale=1, every node/.style={scale=1}
]
\node [draw]{ $u_{7}$}
    child {node[draw] {$u_{5}$}
    child {node[draw] {$u_{1}$}
    child {node[draw] {$l_{1}$}}
    child { node[draw] {$l_{2}$}}}
    child {node[draw] {$u_{2}$}
    child {node[draw] {$l_{3}$}}
    child {  node[draw] {$l_{4}$}}}}
     child {node[draw] {$u_{6}$}
    child {node[draw] {$u_{3}$}
    child {node[draw] {$l_{5}$}}
    child { node[draw] {$l_{6}$}}}
    child {node[draw] {$u_{4}$}
    child {node[draw] {$l_{7}$}}
    child {  node[draw] {$l_{8}$}}}};
    %child {[fill] circle (2pt)
    %child {[fill] circle (2pt)
    %child {node {great-grandchild}}}
    %child {[fill] circle (2pt)}
    %edge from parent node [right] {x}};
 
\end{tikzpicture}
\end{center}
\caption{Example Tree $\T$}\label{fig:topotree}
\end{figure}
\Cref{alg: mgpbo} runs the mirror descent update \Cref{eq:MD_update} according to the order mentioned in $\mathcal{OD}(V\backslash\cL)$.\\\\ 
We first illustrate the state of the costs corresponding to each vertex $v\in V$ after the initialization in Line-6 of \Cref{alg: mgpbo} through \Cref{fig:Cost initialize} for $e=e_{h,m}$.

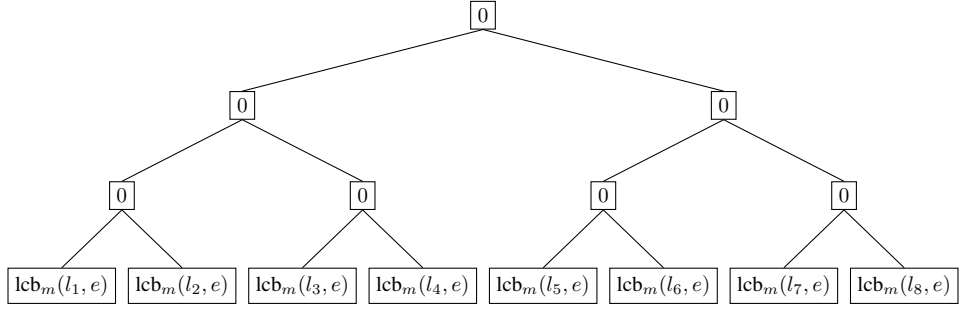
\begin{figure}
\begin{center}
\begin{tikzpicture}
[
    level 1/.style = {black, sibling distance = 8cm},
    level 2/.style = {black, sibling distance = 4 cm},
    level 3/.style = {black, sibling distance = 2 cm},scale=0.8, every node/.style={scale=0.8}
]
\node [draw]{ $0$}
    child {node[draw] {$0$}
    child {node[draw] {$0$}
    child {node[draw] {$\lcb_{m}(l_{1},e)$}}
    child { node[draw] {$\lcb_{m}(l_{2},e)$}}}
    child {node[draw] {$0$}
    child {node[draw] {$\lcb_{m}(l_{3},e)$}}
    child {  node[draw] {$\lcb_{m}(l_{4},e)$}}}}
     child {node[draw] {$0$}
    child {node[draw] {$0$}
    child {node[draw] {$\lcb_{m}(l_{5},e)$}}
    child { node[draw] {$\lcb_{m}(l_{6},e)$}}}
    child {node[draw] {$0$}
    child {node[draw] {$\lcb_{m}(l_{7},e)$}}
    child {  node[draw] {$\lcb_{m}(l_{8},e)$}}}};
    %child {[fill] circle (2pt)
    %child {[fill] circle (2pt)
    %child {node {great-grandchild}}}
    %child {[fill] circle (2pt)}
    %edge from parent node [right] {x}};
 
\end{tikzpicture}
\end{center}
\caption{Cost of each vertex $u\in V$ after initialization in Line-6 of \Cref{alg: mgpbo} ($e=e_{h,m}$).}\label{fig:Cost initialize}
\end{figure}

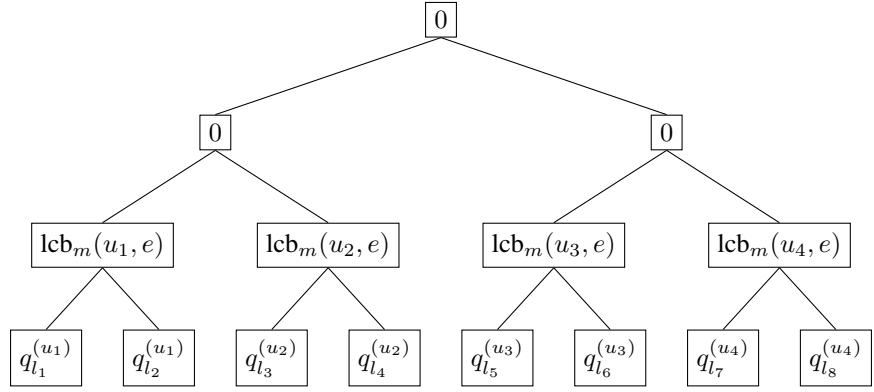
\begin{figure}
\begin{center}
\begin{tikzpicture}
[
    level 1/.style = {black, sibling distance = 6cm},
    level 2/.style = {black, sibling distance = 3 cm},
    level 3/.style = {black, sibling distance = 1.5 cm},scale=1, every node/.style={scale=1}
]
\node [draw]{ $0$}
    child {node[draw] {$0$}
    child {node[draw] {$\lcb_{m}(u_{1},e)$}
    child {node[draw] {$q^{(u_{1})}_{l_{1}}$}}
    child { node[draw] {$q^{(u_{1})}_{l_{2}}$}}}
    child {node[draw] {$\lcb_{m}(u_{2},e)$}
    child {node[draw] {$q^{(u_{2})}_{l_{3}}$}}
    child {  node[draw] {$q^{(u_{2})}_{l_{4}}$}}}}
     child {node[draw] {$0$}
    child {node[draw] {$\lcb_{m}(u_{3},e)$}
    child {node[draw] {$q^{(u_{3})}_{l_{5}}$}}
    child { node[draw] {$q^{(u_{3})}_{l_{6}}$}}}
    child {node[draw] {$\lcb_{m}(u_{4},e)$}
    child {node[draw] {$q^{(u_{4})}_{l_{7}}$}}
    child {  node[draw] {$q^{(u_{4})}_{l_{8}}$}}}};
    %child {[fill] circle (2pt)
    %child {[fill] circle (2pt)
    %child {node {great-grandchild}}}
    %child {[fill] circle (2pt)}
    %edge from parent node [right] {x}};
\end{tikzpicture}
\end{center}
\caption{Conditional Probabilities for $\{l_{1},\dots,l_{8}\}$ and costs for $\{u_{1},\dots,u_{4}\}$ ($e=e_{h,m})$ }\label{fig:first-layer-run}
\end{figure}

Once the sub-routine in Lines $8-9$ is run for $u_{j=1,2,3,4}$ in $\mathcal{OD}(V\backslash\cL)$ one obtains the conditional probabilities $q_{l}^{(par(l))}$ for all leaves $l\in\cL$ using \Cref{eq:MD_update} and costs for internal vertices $\{u_{1},u_{2},u_{3},u_{4}\}$ denoted by $\lcb_{m}(u_{j},e_{h,m})$ using \Cref{eq: parental cost}. In \Cref{fig:first-layer-run} we depict this calculated costs and probabilities.

%$(\lcb_{t}(l_{i},e_{t})$, $p_{l_{i}}^{(par(l_{i}))})$ referring to the cost of the leaf and the conditional probability outputted by the algorithm, the transition of the algorithm layer by layer for the cost of the internal vertices $\lcb_{t}(u,e_{t})$ and conditional probability $p_{u}^{(par(u))}$ will look as follows:

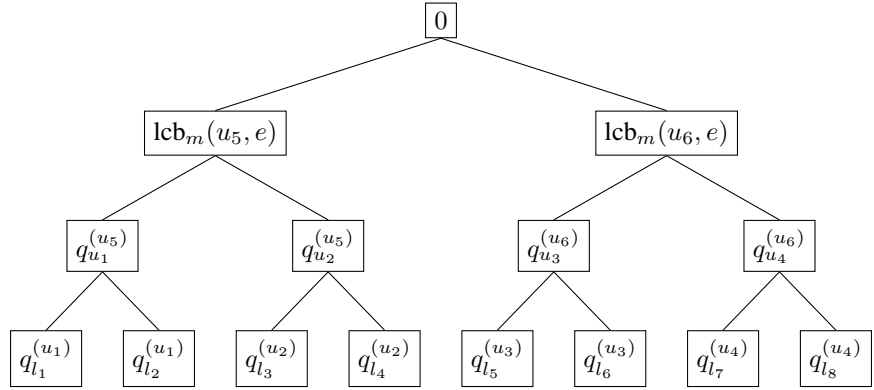
\begin{figure}
\begin{center}
\begin{tikzpicture}
[
    level 1/.style = {black, sibling distance = 6cm},
    level 2/.style = {black, sibling distance = 3 cm},
    level 3/.style = {black, sibling distance = 1.5 cm},scale=1, every node/.style={scale=1}
]
\node [draw]{ $0$}
    child {node[draw] {$\lcb_{m}(u_{5},e)$}
    child {node[draw] {$q^{(u_{5})}_{u_{1}}$}
    child {node[draw] {$q^{(u_{1})}_{l_{1}}$}}
    child { node[draw] {$q^{(u_{1})}_{l_{2}}$}}}
    child {node[draw] {$q_{u_{2}}^{(u_{5})} $}
    child {node[draw] {$q^{(u_{2})}_{l_{3}}$}}
    child {  node[draw] {$q^{(u_{2})}_{l_{4}}$}}}}
     child {node[draw] {$\lcb_{m}(u_{6},e)$}
    child {node[draw] {$q^{(u_{6})}_{u_{3}}$}
    child {node[draw] {$q^{(u_{3})}_{l_{5}}$}}
    child { node[draw] {$q^{(u_{3})}_{l_{6}}$}}}
    child {node[draw] {$q^{(u_{6})}_{u_{4}}$}
    child {node[draw] {$q^{(u_{4})}_{l_{7}}$}}
    child {  node[draw] {$q^{(u_{4})}_{l_{8}}$}}}};
    %child {[fill] circle (2pt)
    %child {[fill] circle (2pt)
    %child {node {great-grandchild}}}
    %child {[fill] circle (2pt)}
    %edge from parent node [right] {x}};
 
\end{tikzpicture}
\end{center}
\caption{Conditional Probabilities for $\{u_{1},\dots,u_{4}\}$ and costs for $\{u_{5},u_{6}\}$ ($e=e_{h,m})$}\label{fig:second-layer-run}
\end{figure}

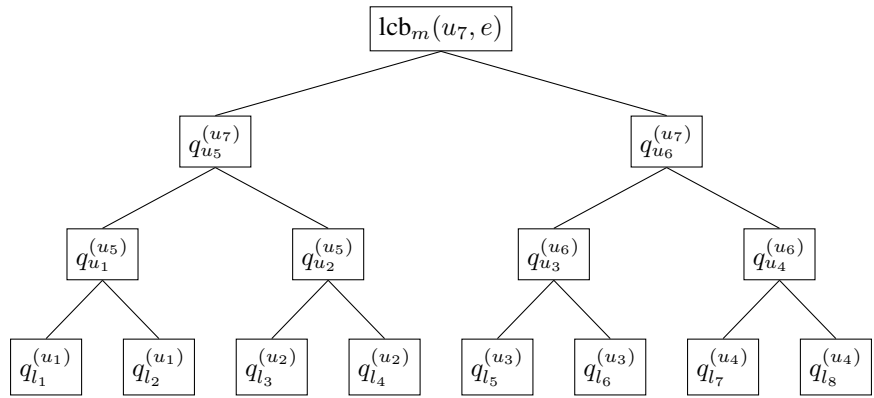
\begin{figure}
\begin{center}
\begin{tikzpicture}
[
    level 1/.style = {black, sibling distance = 6cm},
    level 2/.style = {black, sibling distance = 3 cm},
    level 3/.style = {black, sibling distance = 1.5 cm},scale=1, every node/.style={scale=1}
]
\node [draw]{ $\lcb_{m}(u_{7},e)$}
    child {node[draw] {$q_{u_{5}}^{(u_{7})}$}
    child {node[draw] {$q^{(u_{5})}_{u_{1}}$}
    child {node[draw] {$q^{(u_{1})}_{l_{1}}$}}
    child { node[draw] {$q^{(u_{1})}_{l_{2}}$}}}
    child {node[draw] {$q^{(u_{5})}_{u_{2}}$}
    child {node[draw] {$q^{(u_{2})}_{l_{3}}$}}
    child {  node[draw] {$q^{(u_{2})}_{l_{4}}$}}}}
     child {node[draw] {$q_{u_{6}}^{(u_{7})}$}
    child {node[draw] {$q^{(u_{6})}_{u_{3}}$}
    child {node[draw] {$q^{(u_{3})}_{l_{5}}$}}
    child { node[draw] {$q^{(u_{3})}_{l_{6}}$}}}
    child {node[draw] {$q^{(u_{6})}_{u_{4}}$}
    child {node[draw] {$q^{(u_{4})}_{l_{7}}$}}
    child {  node[draw] {$q^{(u_{4})}_{l_{8}}$}}}};
    %child {[fill] circle (2pt)
    %child {[fill] circle (2pt)
    %child {node {great-grandchild}}}
    %child {[fill] circle (2pt)}
    %edge from parent node [right] {x}};
 
\end{tikzpicture}
\end{center}\caption {Conditional Probabilities for $\{u_{5},u_{6}\}$ and cost for $\{u_{7}\}$ ($e=e_{h,m})$}\label{fig:third-layer-run}
\end{figure}

Then the sub-routine in Lines $8-9$ is run for $u_{j=5,6}$ in $\mathcal{OD}(V\backslash\cL)$ and one obtains the conditional probabilities $q_{u_{j}}^{(par(u_{j}))}$ for $j=\{1,2,3,4\}$  using \Cref{eq:MD_update} and costs for internal vertices $\{u_{5},u_{6}\}$ denoted by $\lcb_{m}(u_{j},e_{h,m})$ using \Cref{eq: parental cost}. In \Cref{fig:second-layer-run} we depict this calculated costs and probabilities.
Finally the sub-routine in Lines $8-9$ is run for $u_{j=7}$ in $\mathcal{OD}(V\backslash\cL)$ and one obtains the conditional probabilities $q_{u_{j}}^{(par(u_{j}))}$ for $j=\{5,6\}$  using \Cref{eq:MD_update} and costs for internal vertex $\{u_{7}\}$ denoted by $\lcb_{m}(u_{7},e_{h,m})$ using \Cref{eq: parental cost} (also signifies the expected cost if the action is sampled from the calculated probabilities after conversion using \Cref{eq: Q-Z relation}). In \Cref{fig:third-layer-run} we depict this calculated costs and probabilities.

\section{Proof of \Cref{th: Theorem-1}}
\label{sec: regret proof}
The first step in the proof of \Cref{th: Theorem-1} is to rewrite the cumulative episodic regret (\Cref{eq: Cumulative Regret}) as a sum of expected episodic regret conditioned w.r.t. data observed till the previous episode. In particular, the main idea is to upper bound $R_{N_{ep}}^{\alpha,\beta}=\sul_{m=1}^{N_{ep}}r_{m}^{\alpha,\beta}$ by $\sul_{m=1}^{N_{ep}}\E_{m}[r_{m}^{\alpha,\beta}|\mathcal{F}_{m-1}]$. To do this, we make use of \cite[Lemma 13]{kirschner2018information} as we explain below. Here, the expectation $\E_{m}$ is w.r.t. the actions in episode $m$, $D_{m}=(x_{h,m})_{h=1}^{H}$ outputted by \Cref{alg: mgpbo}, and $\mathcal{F}_{m}$ denotes the data collected by \Cref{alg: mgpbo} during the first $m$ episodes, i.e., 
\begin{equation}\label{eq: data collected}
    \mathcal{F}_{m}= \lbrace (x_{h,i},e_{h,i},y_{h,i})_{h=1}^{H}  \rbrace_{i=1}^{m}.
    %\{[(x_{1,1},e_{1,1},y_{1,1}),\cdot\cdot,(x_{H,1},e_{H,1},y_{H,1})],\cdot\cdot,[(x_{1,m},e_{1,m},y_{1,m}),\cdot\cdot,(x_{H,m},e_{H,m},y_{H,m})]\}    
\end{equation}
We note that in the episodic regret \[r_{m}^{\alpha,\beta}=\sul_{h=1}^{H}f(x_{h,m},e_{h,m})+\sul_{h=1}^{H}d(x_{h,m},x_{h-1,m})-\alpha \cdot \mathrm{cost}_{m}(D_m^{*})-\beta,\] the term $\alpha \cdot \mathrm{cost}_{m}(D_m^{*})-\beta$ is constant w.r.t. the expectation $\E_{m}$ as it is independent of the actions $(x_{h,m})_{h=1}^{H}$. Also, the episodic cost $\mathrm{cost}_{m}(D_{m})=\sul_{h=1}^{H}f(x_{h,m},e_{h,m})+\sul_{h=1}^{H}d(x_{h,m},x_{h-1,m})$ is trivially upper bounded by $H(B+\psi)$. This is because $f(x,e)\leq B$ (follows from our assumptions $\|f\|_{k}\leq B$ and $k(\cdot,\cdot)\leq 1$) and $d(x,x')\leq\psi$ as detailed in \Cref{sec: Problem Formulation}.

Then, according to \cite[Lemma 13]{kirschner2018information}, with probability at least $1-\delta$ the cumulative regret  from \Cref{eq: Cumulative Regret} is bounded as:
%\begin{align}
%    R_{N_{ep}}^{\alpha,\beta}&=\sul_{m=1}^{N_{ep}}r_{m}^{\alpha,\beta}\\
%    &=\sul_{m=1}^{N_{ep}} \Big(\mathrm{cost}_{m}(D_{m})-\alpha \cdot \mathrm{cost}_{m}(D_m^{*})-\beta\Big).
%\end{align}

%Now in \cite[Lemma 13]{kirschner2018information} with probability at least $1-\delta$ the cumulative sum of the stochastic process $\Delta_{t}$ (signifying regret) upto time $T$ is bounded by sum of the conditional expectations $\E[\Delta_{t}|\mathcal{F}_{t-1}]$ and a term logarithmically dependent on $T$ and $1/\delta$ where $\mathcal{F}_{t-1}$ denotes the data collected upto time $t-1$. And they use the fact that $0\leq \Delta_{t}\leq S$ for some $S$. In our case, as we are in the episodic setting we use this result on the episodic cost ($\mathrm{cost}_{m}(D_{m})$)  which is bounded by $0\leq\mathrm{cost}_{m}(D_{m})\leq H(B+\psi)$ and bound it by $\E_{m}[\mathrm{cost}_{m}(D_{m})|\mathcal{F}_{m-1}]$ where $\mathcal{F}_{m-1}$ is as defined in \Cref{eq: data collected}.
%By using  \cite[Lemma 3]{kirschner2018information} on the stochastic process $0\leq\mathrm{cost}_{m}(D_{m})\leq H(B+\mathrm{diam}(\mathcal{X}))$,
%Hence, we get with probability at least $1-\delta$ at any episode $N_{ep}\geq 1$, 

\begin{align}
R_{N_{ep}}^{\alpha,\beta}&=\sul_{m=1}^{N_{ep}}r_{m}^{\alpha,\beta}\\
\label{eq: regret conversion}
&\leq \sul_{m=1}^{N_{ep}}\E_{m}[r_{m}^{\alpha,\beta}|\mathcal{F}_{m-1}]+4H(B+\psi)\log\Big(\frac{4\pi^{2}N_{ep}^{2}}{3\delta}\big(\log(N_{ep})+1\big) \Big).
\end{align}
Above, we have applied \cite[Lemma 13]{kirschner2018information} to the cumulative sum of the stochastic process $r_{m}^{\alpha,\beta}$ and used the fact that each episodic cost $\mathrm{cost}_{m}(D_{m})$ is bounded by $0\leq\mathrm{cost}_{m}(D_{m})\leq H(B+\psi)$.

In what follows, we focus on bounding 
\begin{equation}\label{eq: bound goal}
\sul_{m=1}^{N_{ep}}\E_{m}[r_{m}^{\alpha,\beta}|\mathcal{F}_{m-1}]=\sul_{m=1}^{N_{ep}}\E_{m}[\mathrm{cost}_{m}(D_{m})-\alpha \cdot \mathrm{cost}_{m}(D_m^{*})-\beta|\mathcal{F}_{m-1}].    
\end{equation}

Recall that, $\E_{m}[\mathrm{cost}_{m}(D_{m})|\mathcal{F}_{m-1}]=\E_{m}[S_{m}(D_{m})|\mathcal{F}_{m-1}]+\E_{m}[M_{m}(D_{m})|\mathcal{F}_{m-1}]$. 
%The episodic movement cost $M_{m}(D_{m})$ is calculated w.r.t. the distance metric $d(\cdot,\cdot)$.
In the following sections we bound this sum of expected episodic costs using results from \cite{coester2019pure} and \cite{chowdhury2019online}. We begin by explicitly writing the episodic service costs in terms of the actions' distributions. 

\subsection{Costs in Terms of Conditional Distribution}
\label{sec: conditional cost}
Recall in \Cref{alg: mgpbo} the sequence of decisions $D_{m}=\{x_{1,m},\dots,x_{H,m}\}$ is sampled from conditional optimal coupling distribution denoted as $\zeta_{h-1,h,m}(U_{h,m}=x|U_{h-1,m}=x_{h-1,m})$ as stated in \Cref{alg: mgpbo}, Line 13. Here $(U_{h-1,m},U_{h,m})$ is a joint random variable whose marginal distributions are $l(z_{h-1,m})$ and $l(z_{h,m})$, respectively, and are computed as stated in \Cref{alg: mgpbo}, Line 11. 

Hence, the expected service cost of the algorithm w.r.t. $\lcb_{m}(\cdot,e_{h,m})$ given $x_{h-1,m}$ becomes $$\sul_{h=1}^{H}\E[\lcb_{m}(x_{h,m},e_{h,m})|x_{h-1,m},\mathcal{F}_{m-1}]=\sul_{h=1}^{H}\langle \lcb_{m}(\cdot,e_{h,m}),\zeta_{h-1,h,m}(\cdot|x_{h-1,m})\rangle.$$ Here the expectation $\E$ is w.r.t. the random variable $x_{h,m}$ whose probability distribution is $\zeta_{h-1,h,m}(U_{h,m}=x|U_{h-1,m}=x_{h-1,m})$ when conditioned on $x_{h-1,m}$. We note that the first state of each episode is sampled from $l(z_{1,m})$ (as there is no randomness in the initial state $x_{0,m}$ \footnote{The optimal coupling conditional distribution minimizing \Cref{eq: wasserstein definition} for $h=1$ will be trivially satisfied by $l(z_{1,m})$ as the random variable $U_{0,m}$ is fixed at $x_{0,m}$}). Also, within any given episode, $\lcb_{m}(\cdot,\cdot)$ in $\algnm$ is fixed and does not get updated. Hence, by taking the total expectation over the whole episode and by using the law of total expectation \footnote{$\E_{x_{1,m},x_{2,m}}[\lcb_{m}(x_{2,m},e_{2,m})]=\sul_{y\in \X}\sul_{x\in \X}\lcb_{m}(x,e_{h,m})\zeta_{1,2,m}(x|y)l(z_{1,m})(y)]=\sul_{x\in \X}\lcb_{m}(x,e_{h,m})\sul_{y\in \X}\zeta_{1,2,m}(x,y)=\sul_{x\in \X}\lcb_{m}(x,e_{h,m})l(z_{2,m})(x)$ (can be similarly proved for any $h$ using an inductive argument)}, we arrive at: 
\begin{equation}
\label{eq: conditional to marginal expectation}
    \E_{m}\Big[\sul_{h=1}^{H}\langle \lcb_{m}(\cdot,e_{h,m}),\zeta_{h-1,h,m}(\cdot|x_{h-1,m})\rangle|\mathcal{F}_{m-1}\Big]=\sul_{h=1}^{H}\langle \lcb_{m}(\cdot,e_{h,m}),l(z_{h,m})\rangle.
\end{equation}
On the other hand, the actual service cost for episode $m$ is $$S_{m}(D_{m})=\sul_{h=1}^{H}f(x_{h,m},e_{h,m})\quad \textrm{where} \hspace{0.5 em}x_{h,m}\sim \zeta_{h-1,h,m}(\cdot|x_{h-1,m}).$$  Hence, we can write $$\E[f(x_{h,m},e_{h,m})|x_{h-1,m},\mathcal{F}_{m-1}]=\langle f(\cdot,e_{h,m}),\zeta_{h-1,h,m}(\cdot|x_{h-1,m})\rangle,$$
and
$$\sul_{h=1}^{H}\E[f(x_{h,m},e_{h,m})|x_{h-1,m},\mathcal{F}_{m-1}]=\sul_{h=1}^{H}\langle f(\cdot,e_{h,m}),\zeta_{h-1,h,m}(\cdot |x_{h-1,m})\rangle.$$ 

Moreover, conditioning on the event of \Cref{lemma:confidence_lemma},  the cumulative cost above can be bounded in terms of its lower confidence bound as
\begin{equation}\label{eq: f-confidence bound}
\begin{split}
\sul_{h=1}^{H}\langle f(\cdot,e_{h,m}),\zeta_{h-1,h,m}(\cdot |x_{h-1,m})\rangle\leq & \sul_{h=1}^{H}\langle \lcb_{m}(\cdot,e_{h,m}),\zeta_{h-1,h,m}(\cdot |x_{h-1,m})\rangle\\ &+\sul_{h=1}^{H}\langle 2\beta_{m}\sigma_{m}(\cdot,e_{h,m}),\zeta_{h-1,h,m}(\cdot |x_{h-1,m})\rangle.
\end{split}
\end{equation}
Also, by using the similar argument that led to \Cref{eq: conditional to marginal expectation}, we have,
\begin{equation}\label{eq: conditional to marginal expectation-2}
\E_{m}[S_{m}(D_{m})|\mathcal{F}_{m-1}]=\E_{m}\Big[\sul_{h=1}^{H}\langle f(\cdot,e_{h,m}),\zeta_{h-1,h,m}(\cdot|x_{h-1,m})\rangle|\mathcal{F}_{m-1}\Big]=\sul_{h=1}^{H}\langle f(\cdot,e_{h,m}),l(z_{h,m})\rangle. 
\end{equation}
%Then taking an expectation over \Cref{eq: f-confidence bound} and using \Cref{eq: conditional to marginal expectation} and \Cref{eq: conditional to marginal expectation-2}, 
Then, combining \Cref{eq: conditional to marginal expectation}, \Cref{eq: f-confidence bound} and \Cref{eq: conditional to marginal expectation-2} we obtain
\begin{align}
%\begin{split}
%\E_{m}\Big[\sul_{h=1}^{H}\langle f(\cdot,e_{h,m}),\zeta_{h-1,h,m}(\cdot|x_{h-1,m})\rangle|\mathcal{F}_{m-1}\Big]\leq &\E_{m}\Big[\sul_{h=1}^{H}\langle \lcb_{m}(\cdot,e_{h,m}),\zeta_{h-1,h,m}(\cdot|x_{h-1,m})\rangle|\mathcal{F}_{m-1}\Big]\\&+\E_{m}[\sul_{h=1}^{H}\langle 2\beta_{m}\sigma_{m}(\cdot,e_{h,m}),\zeta_{h-1,h,m}(\cdot |x_{h-1,m})\rangle|\mathcal{F}_{m-1}].
%\end{split}\\
\begin{split}
\sul_{h=1}^{H}\langle f(\cdot,e_{h,m}),l(z_{h,m})\rangle\leq & \sul_{h=1}^{H}\langle \lcb_{m}(\cdot,e_{h,m}),l(z_{h,m})\rangle\\&+\E_{m}[\sul_{h=1}^{H}\langle 2\beta_{m}\sigma_{m}(\cdot,e_{h,m}),\zeta_{h-1,h,m}(\cdot |x_{h-1,m})\rangle|\mathcal{F}_{m-1}].
\end{split}
\end{align}
By summing over all episodes (note that \Cref{lemma:confidence_lemma} holds uniformly over all episodes), we arrive at
\begin{align}\label{eq: decompose f}
    \begin{split}
    \sul_{m=1}^{N_{ep}}\sul_{h=1}^{H}\langle f(\cdot,e_{h,m}),l(z_{h,m})\rangle &\leq \sul_{m=1}^{N_{ep}}\sul_{h=1}^{H}\langle \lcb_{m}(\cdot,e_{h,m}),l(z_{h,m})\rangle \\ &\quad +\sul_{m=1}^{N_{ep}}\E_{m}[\sul_{h=1}^{H}\langle 2\beta_{m}\sigma_{m}(\cdot,e_{h,m}),\zeta_{h-1,h,m}(\cdot |x_{h-1,m})\rangle|\mathcal{F}_{m-1}].
    \end{split}
\end{align}

Now we define the offline optimal sequence for the cost sequence provided to \Cref{alg: mgpbo} in episode $m$ $\{\lcb_{m}(\cdot,e_{1,m}),\dots,\lcb_{m}(\cdot,e_{H,m})\}$ as $\{bx^{*}_{1,m},\dots,bx^{*}_{H,m}\}$ which will be useful in the analysis to bound $\sul_{m=1}^{N_{ep}}\sul_{h=1}^{H}\langle \lcb_{m}(\cdot,e_{h,m}),l(z_{h,m})\rangle $. Note that this offline optimal sequence is different from $D_{m}^{*}$ as it is calculated for the cost sequence  $\{\lcb_{m}(\cdot,e_{1,m}),\dots,\lcb_{m}(\cdot,e_{H,m})\}$ and not  $\{f(\cdot,e_{1,m}),\dots,f(\cdot,e_{H,m})\}$.

In \Cref{alg: mgpbo}, we use the mirror descent approach as proposed in \cite{coester2019pure} with input as the cost sequence $\{\lcb_{m}(\cdot,e_{1,m}),\dots,\lcb_{m}(\cdot,e_{H,m})\}$ to output the actions $D_{m}$. Since \Cref{alg: mgpbo} uses $\lcb_{m}(\cdot,e_{h,m})$ rather than $f(\cdot,e_{h,m})$ the guarantees provided in \cite{coester2019pure} for action sequence $D_{m}$ will hold only w.r.t. $\{bx^{*}_{1,m},\dots,bx^{*}_{H,m}\}$ (the offline optimal sequence w.r.t. $\{\lcb_{m}(\cdot,e_{1,m}),\dots,\lcb_{m}(\cdot,e_{H,m})\}$). Hence we can invoke  \citet[Corollary 4]{coester2019pure}, for the probability vector sequence $\{l(z_{1,m}),\dots,l(z_{H,m})\}$ to guarantee that the sequence is  1-competitive in service costs (w.r.t. $\{\lcb_{m}(\cdot,e_{1,m}),\dots,\lcb_{m}(\cdot,e_{H,m})\}$) and $\mathcal{O}((\log n)^{2})-$competitive for movement costs w.r.t. the offline optimal sequence  $\{bx^{*}_{1,m},\dots,bx^{*}_{H,m}\}$.  Using the definition of refined competitive ratio guarantees \Cref{eq: service guarantee} and \Cref{eq: movement guarantee} we have,
\begin{align}
    \label{eq: from coester-1}
    \begin{split}
        \sul_{h=1}^{H}\langle \lcb_{m}(\cdot,e_{h,m}),l(z_{h,m})\rangle &\leq \mathcal{O}(1)+\sul_{h=1}^{H} \Big( \lcb_{m}(bx^{*}_{h,m},e_{h,m}) +d(bx^{*}_{h,m},bx^{*}_{h-1,m})\Big) ,
    \end{split}\\
    \begin{split}
        \label{eq: from coester-2}
    \E_{m}[M_{m}(D_{m})|\mathcal{F}_{m-1}] &\leq   \mathcal{O}(1)\\&+\mathcal{O}((\log n)^{2})\sul_{h=1}^{H} \Big(\lcb_{m}(bx^{*}_{h,m},e_{h,m}) +d(bx^{*}_{h,m},bx^{*}_{h-1,m})\Big).
    \end{split}
\end{align}
Focusing on $\sul_{m=1}^{N_{ep}}\sul_{h=1}^{H}\langle \lcb_{m}(\cdot,e_{h,m}),l(z_{h,m})\rangle $ in \Cref{eq: decompose f}  and by using  \Cref{eq: from coester-1}, we have
\begin{equation}\label{eq: bound lcb optimal}
\begin{split}
    \sul_{m=1}^{N_{ep}}\sul_{h=1}^{H}\langle \lcb_{m}(\cdot,e_{h,m}),l(z_{h,m})\rangle \leq & \sul_{m=1}^{N_{ep}}\Big(\mathcal{O}(1)+ \sul_{h=1}^{H} \big( \lcb_{m}(bx^{*}_{h,m},e_{h,m}) +d(bx^{*}_{h,m},bx^{*}_{h-1,m})\big)\Big).  
\end{split}
\end{equation}
%Let $(zf_{h,m}^{*})_{h=1}^{H}$ be the offline optimal sequence starting at $x_{0,m}$ for the cost sequence $[f(\cdot,e_{h,m})]_{h=1}^{H}$ for a particular episode $m$ w.r.t $d_{\T}(\cdot,\cdot)$ (also denoted as $D_{m,\T}^{*}$). 

Since $D_{m}^{*}=\{x_{1,m}^{*},\dots,x_{H,m}^{*}\}$ is not optimal w.r.t. $\lcb_{m}(\cdot,e_{h,m})$, it will incur more cost than $\{bx^{*}_{1,m},\dots,bx^{*}_{H,m}\}$ w.r.t.~ $\lcb_{m}(\cdot,e_{h,m})$, i.e,
\begin{equation}
\label{eq: lcb optimal vs f optimal}
\begin{split}
    \sul_{h=1}^{H}\Big( \lcb_{m}(bx^{*}_{h,m},e_{h,m}) +d(bx^{*}_{h,m},bx^{*}_{h-1,m})\Big)\leq &  \sul_{h=1}^{H}  \Big(\lcb_{m}(x^{*}_{h,m},e_{h,m})  +d(x^{*}_{h,m},x^{*}_{h-1,m}) \Big).
\end{split}
\end{equation}
Hence, by plugging \Cref{eq: lcb optimal vs f optimal} in \Cref{eq: bound lcb optimal} we have ,
\begin{align}
    \begin{split}
    \sul_{m=1}^{N_{ep}}\sul_{h=1}^{H}\langle \lcb_{m}(\cdot,e_{h,m}),l(z_{h,m})\rangle \leq &  \sul_{m=1}^{N_{ep}} \Big(\mathcal{O}(1)+\sul_{h=1}^{H} \big(\lcb_{m}(x^{*}_{h,m},e_{h,m})  +d(x^{*}_{h,m},x^{*}_{h-1,m})\big) \Big)
    \end{split}\\
    \begin{split}
    \label{eq: service lcb bound}
     \leq &  \sul_{m=1}^{N_{ep}}\Big( \mathcal{O}(1)+\sul_{h=1}^{H} \big(f(x^{*}_{h,m},e_{h,m})  +d(x^{*}_{h,m},x^{*}_{h-1,m})\big)\Big),
     \end{split}
\end{align}
where the last equation \Cref{eq: service lcb bound} holds because we have conditioned on the event that confidence bounds hold true simultaneously for all episodes. 

We can now use the above results to bound the cumulative service cost  as
\begin{align}
   \sul_{m=1}^{N_{ep}}\E_{m}[S_{m}|\mathcal{F}_{m-1}]& \stackrel{\text{(i)}}{=} 
    \sul_{m=1}^{N_{ep}}\sul_{h=1}^{H}\langle f(\cdot,e_{h,m}),l(z_{h,m})\rangle\\  \begin{split}
    &\stackrel{\text{(ii)}}{\leq} \sul_{m=1}^{N_{ep}}\sul_{h=1}^{H}\langle \lcb_{m}(\cdot,e_{h,m}),l(z_{h,m})\rangle \\  &+\sul_{m=1}^{N_{ep}}\E_{m}[\sul_{h=1}^{H}\langle 2\beta_{m}\sigma_{m}(\cdot,e_{h,m}),\zeta_{h-1,h,m}(\cdot |x_{h-1,m})\rangle|\mathcal{F}_{m-1}].
    \end{split}\\
    \label{eq: service cost bound}
    \begin{split}
    &\stackrel{\text{(iii)}}{\leq}   \sul_{m=1}^{N_{ep}}\Big(\mathcal{O}(1)+ \sul_{h=1}^{H} \big(f(x^{*}_{h,m},e_{h,m})  +d(x^{*}_{h,m},x^{*}_{h-1,m})\big)\Big)\\
    &+\sul_{m=1}^{N_{ep}}\E_{m}[\sul_{h=1}^{H}\langle 2\beta_{m}\sigma_{m}(\cdot,e_{h,m}),\zeta_{h-1,h,m}(\cdot |x_{h-1,m})\rangle|\mathcal{F}_{m-1}].
     \end{split}
\end{align}
Here (i) follows from \Cref{eq: conditional to marginal expectation-2}, (ii) from \Cref{eq: decompose f} and (iii) from \Cref{eq: service lcb bound}.
Finally, we focus on bounding movement cost by using \Cref{eq: from coester-2} and \Cref{eq: service lcb bound}:
\begin{align}
\sul_{m=1}^{N_{ep}}\E_{m}\Big[M_{m}(D_{m})|\mathcal{F}_{m-1}\Big]=&\sul_{m=1}^{N_{ep}} \E_{m}\Big[\sul_{h=1}^{H}d(x_{h,m},x_{h-1,m})|\mathcal{F}_{m-1}\Big]\\
\leq & \sul_{m=1}^{N_{ep}}\Big(\mathcal{O}(1)+\mathcal{O}((\log n)^{2})\sul_{h=1}^{H} \big(\lcb_{m}(bx^{*}_{h,m},e_{h,m}) +d(bx^{*}_{h,m},bx^{*}_{h-1,m})\big) \Big)\\
    \begin{split}\label{eq: movement cost bound}
         \leq & \sul_{m=1}^{N_{ep}}\Big(\mathcal{O}(1)+\mathcal{O}((\log n)^{2})\sul_{h=1}^{H}\big(f(x^{*}_{h,m},e_{h,m}) +d(x^{*}_{h,m},x^{*}_{h-1,m})\big)\Big). 
    \end{split}
\end{align}
\subsection{Bounding Learning Error}
\label{sec: le bound}
As a last step, we focus on bounding the second term in \Cref{eq: decompose f} that we refer to as the \emph{learning error}, i.e.,   $\sul_{m=1}^{N_{ep}}\E_{m}\Big[\sul_{h=1}^{H}\langle 2\beta_{m}\sigma_{m}(\cdot,e_{h,m}),\zeta_{h-1,h,m}(\cdot|x_{h-1,m})\rangle|\mathcal{F}_{m-1}\Big]$. After that, we can finally bound the cumulative regret using the previously obtained bounds on service and movement costs (\Cref{eq: service cost bound}and \Cref{eq: movement cost bound}, respectively). %To bound it, we also use  \cite[Lemma 3]{kirschner2018information}.\looseness=-1
%and bound it using techniques from \cite{kirschner2018information}. 

Consider the stochastic process,
$$\Delta_{m}=\sul_{h=1}^{H}\sigma_{m}^{2}(x_{h,m},e_{h,m}).$$Here $\sigma_{m}(\cdot,\cdot)$ is constructed from the data up to $m-1$ episodes defined earlier as $\mathcal{F}_{m-1}$. Now, using a similar argument that led to \Cref{eq: conditional to marginal expectation} (fixed $\sigma_{m}(\cdot,\cdot)$ for any given episode and law of total expectation), the conditional mean of $\Delta_{m}$ given is $\mathcal{F}_{m-1}$
$$\E_{m}[\Delta_{m}|\mathcal{F}_{m-1}]=\sul_{h=1}^{H}\langle \sigma_{m}^{2}(\cdot,e_{h,m}),l(z_{h,m})\rangle.$$ 
%This holds since we do not condition w.r.t. $x_{h-1,m}$ at each point, but conditioned only w.r.t. samples up to the previous episode. Hence, it becomes the expectation over the whole episode event though we sample from $\zeta_{h-1,h,m}$.

Now in order to bound this sum of conditional means of posterior variance  $\sul_{m=1}^{N_{ep}}\E_{m}[\Delta_{m}|\mathcal{F}_{m-1}]$, by observed posterior variance $\sul_{m=1}^{N_{ep}}\Delta_{m}$ we use \cite[Lemma 3]{kirschner2018information}. Note that $\sigma^{2}_{m}(x,e)\leq 1$ by our assumption $k(\cdot,\cdot)\leq 1$ in \Cref{sec: intro} and the stochastic process $\Delta_{m}$ can be bounded as follows $$ \Delta_{m}=\sul_{h=1}^{H}\sigma_{m}^{2}(x_{h,m},e_{h,m})\leq \sul_{h=1}^{H}(1)\leq H.$$ Hence applying \cite[Lemma 3]{kirschner2018information} with probability at least $1-\delta$, $$\sul_{m=1}^{N_{ep}}\E_{m}[\Delta_{m}|\mathcal{F}_{m-1}]\leq 2\Big(\sul_{m=1}^{N_{ep}}(\Delta_{m})\Big) + 4H\log(1/\delta)+8H\log(4 H)+1,$$
%where $\Delta_{N_{ep}}\leq b_{N_{ep}}=H $
%($\because \Delta_{m}=\sul_{h=1}^{H}\sigma_{m}^{2}(x_{h,m},e_{h,m})\leq \sul_{h=1}^{H}(1)\leq H$).
This implies,
\begin{equation}\label{eq: kirs lemma-3}
\sul_{m=1}^{N_{ep}}\sul_{h=1}^{H}\langle \sigma_{m}^{2}(\cdot,e_{h,m}),l(z_{h,m})\rangle\leq 2\sul_{m=1}^{N_{ep}}\sul_{h=1}^{H}\sigma_{m}^{2}(x_{h,m},e_{h,m})+4H\log(1/\delta)+8H\log(4H)+1    .
\end{equation}

Note that \Cref{eq: kirs lemma-3} cannot be directly bounded using bounds for sum of observed posterior variance as done in \cite{Srinivas2010} and \cite{chowdhury2017kernelized}. This is because $\sigma_{m}(\cdot,\cdot)$ is not updated continuously and is constant within any given episode $m$. Hence we first need to bound $\sul_{m=1}^{N_{ep}}\sul_{h=1}^{H}\sigma_{m}^{2}(x_{h,m},e_{h,m})$ by $\sul_{m=1}^{N_{ep}}\sul_{h=1}^{H}\sigma_{h,m}^{2}(x_{h,m},e_{h,m})$ where $\sigma_{h,m}(\cdot,\cdot)$ is constructed based on all $\{(x_{h,m},e_{h,m},y_{h,m})\}$ observed up to the round $h$ in the $m$-th episode. Towards this end, we recall the following Lemma from \citet{chowdhury2019online}.

\begin{lemma}\label{lemma: Posterior Variance Bound}
Let $k : \X \times \X \to \mathbb{R}$ be a symmetric positive
semi-definite kernel such that it has bounded variance, i.e. $k(x, x) \leq 1$ for all $x \in \X$ and $f \sim GP_{\X} (0, k)$ be
a sample from the associated Gaussian process over $\X$, then for all $s \geq 1$ and $x \in X$,

\begin{equation}\label{eq: sequential bound sigma}
\sigma_{s-1}^{2}(x) \leq (1+\lambda^{-1})\sigma_{s}^{2}(x),     
\end{equation}

and
\begin{equation}\label{eq: cumulative bound sigma}
\sum_{s=1}^{t}\sigma^{2}_{s-1}\leq(1+2\lambda)\sum_{s=1}^{t}\frac{1}{2}\ln(1+\lambda^{-1}\sigma_{s-1}(x_{s}))\leq 2\lambda \gamma_{t}(k,\X).
\end{equation}
\end{lemma}

Following the proof technique of \citet[Lemma-11]{chowdhury2019online} and using \Cref{eq: sequential bound sigma} and \Cref{eq: cumulative bound sigma} to bound $\sul_{m=1}^{N_{ep}}\sul_{h=1}^{H}\sigma_{m}^{2}(x_{h,m},e_{h,m})$, we have for $\lambda=H$,
\begin{align}
    \sul_{m=1}^{N_{ep}}\sul_{h=1}^{H}\sigma_{m}^{2}(x_{h,m},e_{h,m})&\leq \sul_{m=1}^{N_{ep}}\Big(\sul_{h=1}^{H}(1+1/H)^{h-1}\sigma_{h,m}^{2}(x_{h,m},e_{h,m})\Big)\\
    &\leq (1+1/H)^{H-1}\sul_{m=1}^{N_{ep}}\sul_{h=1}^{H} \sigma_{h,m}^{2}(x_{h,m},e_{h,m})\\
    &\leq (1+1/H)^{H-1}(1+2H)\sul_{m=1}^{N_{ep}}\sul_{h=1}^{H} \frac{1}{2}\ln\Big(1+\frac{1}{H}\sigma_{h,m}^{2}(x_{h,m},e_{h,m})\Big)\\
    \label{eq: chowlemma}
    &\leq 2eH\gamma_{N_{ep}H}.
\end{align}
Here the last inequality is due to $(1 + H^{-1})^{H}\leq  e$
and $(1 + H^{-1})^{-1}(2H + 1) \leq 2H$.
%Hence the $\exp(\gamma_{H})$ term in regret will be replaced by $H$ leading to a linear dependence of $H$ in regret similar. Such a linear dependence in the episode length $H$ is also found in \citet{chowdhury2019online,curi2021combining} where a similar episodic setting is considered. 

Now, we are in position to focus on the learning error. It turns out that the learning error can be simplified by using similar arguments as in \Cref{eq: conditional to marginal expectation}, i.e., 
\begin{align}
    \sul_{m=1}^{N_{ep}}\E_{m}\Big[\sul_{h=1}^{H}\langle 2\beta_{m}\sigma_{m}(\cdot,e_{h,m}),\zeta_{h-1,h,m}(\cdot|x_{h-1,m})\rangle\Big|\mathcal{F}_{m-1}\Big]&=\sm 2\beta_{m}\langle \sigma_{m}(\cdot,e_{h,m}),l(z_{h,m})\rangle.
\end{align}
As $\beta_{m}$ is a non-decreasing sequence as stated in \Cref{lemma:confidence_lemma}, we have
\begin{align}
    \sm 2\beta_{m}\langle \sigma_{m}(\cdot,e_{h,m}),l(z_{h,m})\rangle
    &\leq 2\beta_{N_{ep}}\sm \langle \sigma_{m}(\cdot,e_{h,m}),l(z_{h,m})\rangle\\
    \label{eq: cauchy schwartz usage}
    &\leq  2\beta_{N_{ep}}\sqrt{N_{ep}H\sm (\langle \sigma_{m}(\cdot,e_{h,m}),l(z_{h,m})\rangle)^{2}}\\
    \label{eq: jensen usage}
    &\leq  2\beta_{N_{ep}}\sqrt{N_{ep}H\sm \langle \sigma_{m}^{2}(\cdot,e_{h,m}),l(z_{h,m})\rangle}.
\end{align}
We obtain \Cref{eq: cauchy schwartz usage} using Cauchy-Schwartz inequality and \Cref{eq: jensen usage} using Jensen's inequality since $l(z_{h,m})$ is a probability distribution.

Finally, the learning error will be bounded by 
\begin{align}
\begin{split}\label{eq: Gp machinery}
\sul_{m=1}^{N_{ep}}\E_{m}\sul_{h=1}^{H}\langle 2\beta_{m}\sigma_{m}(\cdot,e_{h,m}),\zeta_{h-1,h,m}(\cdot|x_{h-1,m})\rangle|\mathcal{F}_{m-1}]\\
\quad \stackrel{\text{(i)}}{\leq}  2\beta_{N_{ep}}\sqrt{N_{ep}H\sm \langle \sigma_{m}^{2}(\cdot,e_{h,m}),l(z_{h,m})\rangle}
\end{split}\\
\label{eq: use kirschner results}
&\stackrel{\text{(ii)}}{\leq} 2\beta_{N_{ep}}\sqrt{2N_{ep}H\sul_{m=1}^{N_{ep}}\sul_{h=1}^{H}\sigma_{m}^{2}(x_{h,m},e_{h,m})+4H\log(1/\delta)+8H\log(4H)+1 } \\
\label{eq: learning error bound}
&\stackrel{\text{(iii)}}{\leq}\mathcal{O}\Big(\beta_{N_{ep}}\sqrt{N_{ep}H(2eH\gamma_{N_{ep}H}+4H\log(1/\delta)+8H\log(4H)+1)}\Big).   
\end{align}
%Here we have $\exp{(\gamma_{H})}$ exponential in H for Matern Kernel. But using ideas from \cite{desautels2014parallelizing} (Uncertainty sampling for first $T_{init}$ steps), $\exp{(\gamma_{H})}$ can be bounded by a constant. (Lemma-4 and Theorem-5 from \cite{desautels2014parallelizing}). 
Here (i) follows from \Cref{eq: jensen usage}, (ii)  from \Cref{eq: kirs lemma-3} and (iii) from \Cref{eq: chowlemma}. 

\subsection{Bounding the regret}
\label{sec: final regret bound}

To bound the cumulative regret, recall our initial goal in \Cref{eq: bound goal} to bound  
$$\sul_{m=1}^{N_{ep}}\E_{m}[\mathrm{cost}_{m}(D_{m})|\mathcal{F}_{m-1}]= \sul_{m=1}^{N_{ep}}\E_{m}[S_{m}(D_{m})+M_{m}(D_{m})|\mathcal{F}_{m-1}].$$
By using \Cref{eq: learning error bound} in \Cref{eq: service cost bound}, we bound the service costs $\sul_{m=1}^{N_{ep}}\E_{m}[S_{m}(D_{m})|\mathcal{F}_{m-1}]$ as

\begin{equation}\label{eq: service cost bound with error}
    \begin{split}
    \sul_{m=1}^{N_{ep}}\E_{m}[S_{m}(D_{m})|\mathcal{F}_{m-1}]&\leq   \sul_{m=1}^{N_{ep}}\Big( \mathcal{O}(1)+\sul_{h=1}^{H} \big(f(x^{*}_{h,m},e_{h,m})  +d(x^{*}_{h,m},x^{*}_{h-1,m})\big)\Big)\\
    &+\mathcal{O}\Big(\beta_{N_{ep}}\sqrt{N_{ep}H(2eH\gamma_{N_{ep}H}+4H\log(1/\delta)+8H\log(4H)+1)}\Big).
     \end{split}
\end{equation}
Also, \Cref{eq: movement cost bound} bounds movement costs as follows:
\begin{equation*}
    \begin{split}
        \sul_{m=1}^{N_{ep}}\E_{m}[M_{m}(D_{m})|\mathcal{F}_{m-1}] \leq & \sul_{m=1}^{N_{ep}}\Big(\mathcal{O}(1)+\mathcal{O}((\log n)^{2})\sul_{h=1}^{H}\big(f(x^{*}_{h,m},e_{h,m}) +d(x^{*}_{h,m},x^{*}_{h-1,m})\big)\Big). 
    \end{split}
\end{equation*}

Combining this we obtain,
\begin{align}\label{eq: sub theorem final}
       \sul_{m=1}^{N_{ep}}\E_{m}[\mathrm{cost}_{m}(D_{m})&|\mathcal{F}_{m-1}]= \sul_{m=1}^{N_{ep}}\E_{m}[S_{m}(D_{m})+M_{m}(D_{m})|\mathcal{F}_{m-1}]\\
\begin{split}
       \leq & \sul_{m=1}^{N_{ep}}\Big(\mathcal{O}(1)+\mathcal{O}((\log n)^{2})\sul_{h=1}^{H}\big(f(x^{*}_{h,m},e_{h,m}) +d(x^{*}_{h,m},x^{*}_{h-1,m})\big)\Big) \\ & +\mathcal{O}\Big(\beta_{N_{ep}}\big(N_{ep}H(2eH\gamma_{N_{ep}H}+4H\log(1/\delta)+8H\log(4H))\big)^{\frac{1}{2}}\Big) 
\end{split}\\
\label{eq: combined expected bound}
\begin{split}
       \leq &   \sul_{m=1}^{N_{ep}}\Big(\mathcal{O}(1)+\mathcal{O}((\log n)^{2}) \cdot \mathrm{cost}_{m}(D_m^{*})\Big)\\ & + \mathcal{O}\Big(\beta_{N_{ep}}\big(N_{ep}H(2eH\gamma_{N_{ep}H}+4H\log(1/\delta)+8H\log(4H))\big)^{\frac{1}{2}}\Big). 
\end{split}
\end{align}
Thus, by setting set $\alpha=\mathcal{O}((\log n)^{2})$ and $\beta=\mathcal{O}(1)$ the expected regret $\sul_{m=1}^{N_{ep}}\E_{m}[R_{m}^{\alpha,\beta}|\mathcal{F}_{m-1}]$ from \Cref{eq: regret conversion} can be bounded using \Cref{eq: combined expected bound} as 
\begin{align}
    \sul_{m=1}^{N_{ep}}\E_{m}[r_{m}^{\alpha,\beta}|\mathcal{F}_{m-1}]&=\sul_{m=1}^{N_{ep}}\E_{m}[\mathrm{cost}_{m}(D_{m})-\mathcal{O}((\log n)^{2}) \cdot \mathrm{cost}_{m}(D_m^{*})-\mathcal{O}(1)|\mathcal{F}_{m-1}]\\
    \label{eq: Theorem-1 final}
    &\leq \mathcal{O}\Big(\beta_{N_{ep}}\big(N_{ep}H(2eH\gamma_{N_{ep}H}+4H\log(1/\delta)+8H\log(4H))\big)^{\frac{1}{2}}\Big).
\end{align}

Now, for the final step of the regret guarantees is to ensure that \Cref{eq: Theorem-1 final} holds with probability at least $1-\delta$. For this,  we take the union bound over the events such that \Cref{lemma:confidence_lemma}, \Cref{eq: kirs lemma-3} and \Cref{eq: regret conversion} hold. This effectively replaces $\delta$ by $\delta/3$ in each of the statements. Hence we get the required regret guarantee with probability at least $1-\delta$ as stated in \Cref{th: Theorem-1}:
\begin{equation}
R_{N_{ep}}^{\alpha,\beta}\leq  \mathcal{O}\Big(\beta_{N_{ep}}  \big(H^{2}N_{ep} \gamma_{HN_{ep}} 
  + H\log(\tfrac{H}{\delta}) \big)^\frac{1}{2}+ H(B+\psi)\log\big(\tfrac{N_{ep}\log(N_{ep})}{\delta}\big)\Big).
\end{equation}

\section{Experimental Details}
\label{sec:exptsupp}
In this section, we explain how we arrived at the energy generation formula and objective function based on the discrete-time power generation formula (Eqs.~(10), (11)) from \cite{baheri2017}:
\begin{equation}\label{eq:power_generation_formula}
    P(x,t)=c_{1}(\min\lbrace{V_{w}(x,t),V_{r}\rbrace})^{3}-c_{2}V_{w}^{2}(x,t)-c_{3}V_{r}^{2}\frac{|x-x'|}{\Delta t'}.
\end{equation} 
The first two terms in \Cref{eq:power_generation_formula} determine the instantaneous power generated at a particular altitude. As we update altitudes only with a time gap of $\Delta t=60$ min, the energy generated at this altitude for $\Delta t$ time would be 
$ E_{S}(x,t)=\big(c_{1}(\min\lbrace V_{w}(x,t),V_{r}\rbrace)^{3}-c_{2}V_{w}^{2}(x,t)\big)\Delta t.
$ 
%Based on this, we define our objective function $f(x,t)$ to be, \begin{equation}\label{eq: power objective function-supp} f(x,t)=(c_{1}(\min{V_{w}(x,t),V_{r}})^{3}-c_{2}V_{w}^{2}(x,t))*(\Delta t)\end{equation}
Here $x$ denotes the altitude and corresponds to our action space and $t$ denotes to the time of day and corresponds to our contexts. Next, $x'$ denotes the altitude at time $t-1$, and $\Delta t'$ is the time taken to change altitude. After time interval $\Delta t'$, the altitude is fixed for the rest of $\Delta t$ and the third in \Cref{eq:power_generation_formula} becomes zero. Also, $\Delta t'$ may or may not be equal to $\Delta t$ (usually, it is lower). But the energy lost in changing altitude will be $E_{M}(x,x')=c_{3}V_{r}^{2}\frac{|x-x'|}{\Delta t'}*\Delta t'=c_{3}V_{r}^{2}|x-x'|$. As we update $x$ only in intervals of $\Delta t$, the energy lost for movement for every $\Delta t$ time would be $E_{M}(x,x')$.

\subsection{Plots from Other Locations}
\begin{figure}[h!]
    \centering
    \begin{subfigure}[b]{.3\textwidth}
        \centering
        \includegraphics[width=\linewidth]{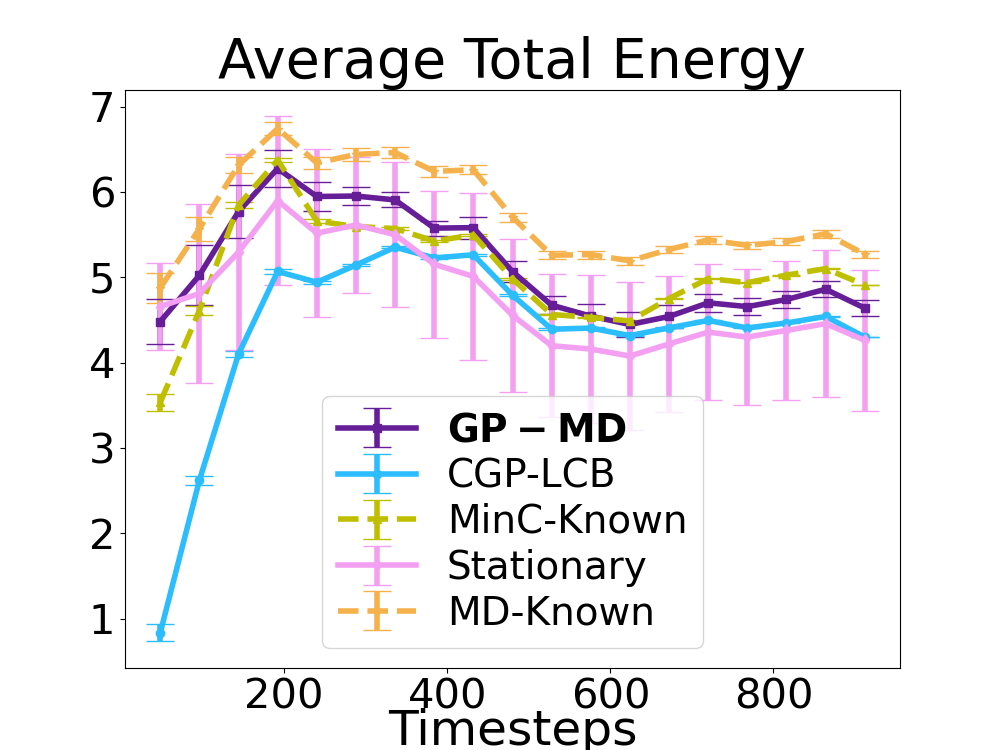}
        \caption{Total Energy Generated for $\rho=4$}
        \label{fig:fig31}
    \end{subfigure}
    \hspace{1.5em}
    \begin{subfigure}[b]{.3\textwidth}
        \centering
        \includegraphics[width=\linewidth]{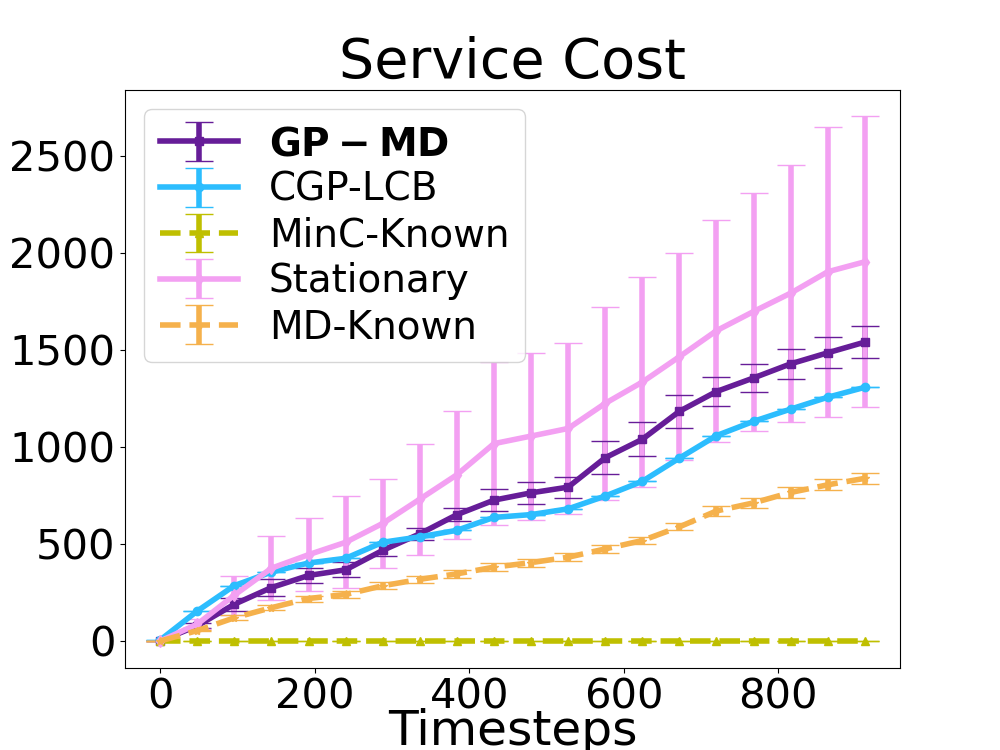}
        %\caption{Figure2}
        \caption{Service Cost for $\rho=4$}
        \label{fig:fig32}
    \end{subfigure}
    \hspace{1.5em}
    \begin{subfigure}[b]{.3\textwidth}
        \centering
        \includegraphics[width=\linewidth]{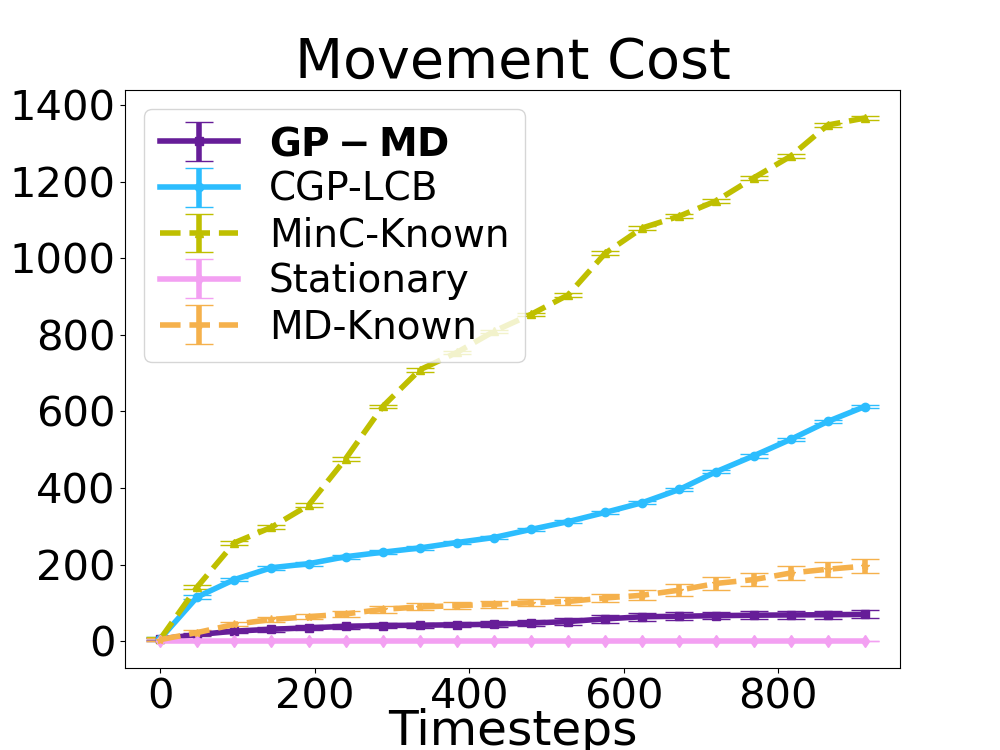}
        %\caption{Figure3}
        \caption{Movement Cost for $\rho=4$}
        \label{fig:fig33}
    \end{subfigure}

    \caption{\footnotesize  AWE altitude optimization task;
    \Cref{fig:fig31}: The average total generated energy over $960$ hours based on the wind data at a single location (Latitude$=53$ and Longitude$=-4$) for $\rho=4$. \Cref{fig:fig32,fig:fig33}: The service and movement costs for $\rho=4$.}
    \label{fig:figure3}
\end{figure}

\begin{figure}[h!]
    \centering
    \begin{subfigure}[b]{.3\textwidth}
        \centering
        \includegraphics[width=\linewidth]{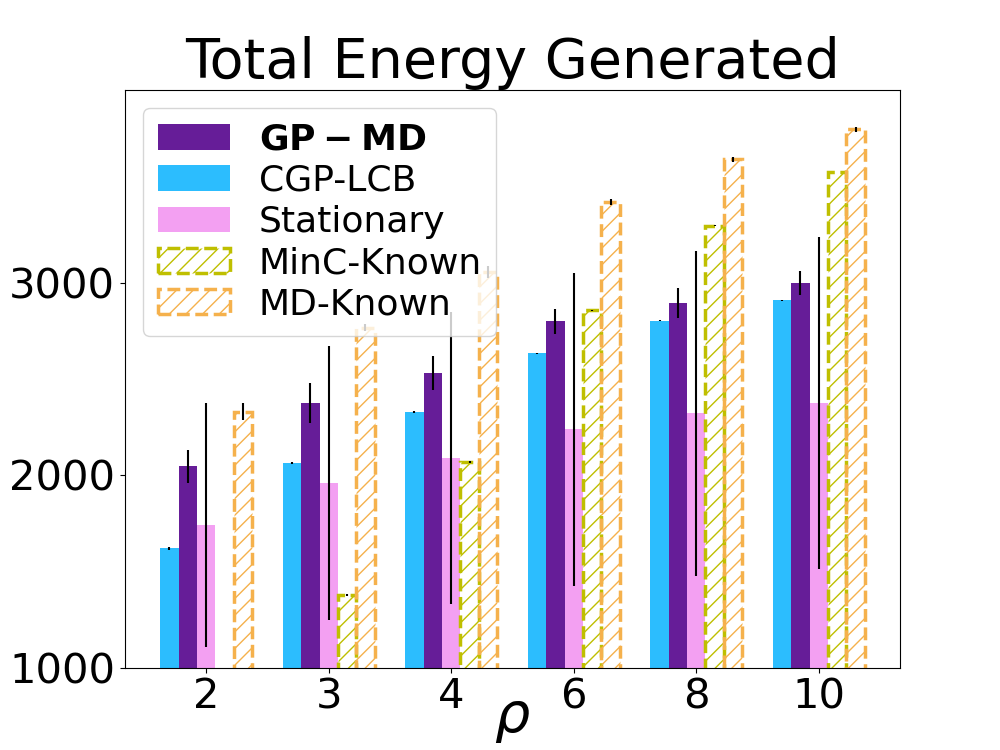}
        \caption{Total Energy Generated for varying $\rho$ }
        \label{fig:fig41}
    \end{subfigure}
    \hspace{1.5em}
    \begin{subfigure}[b]{.3\textwidth}
        \centering
        \includegraphics[width=\linewidth]{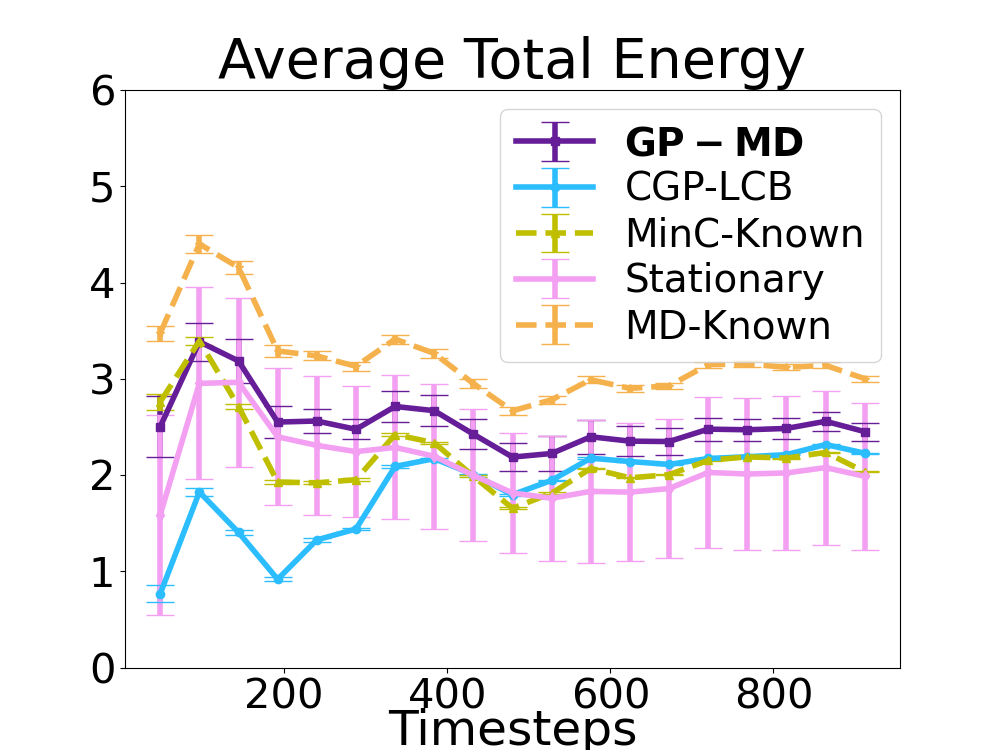}
        %\caption{Figure2}
        \caption{ Total Energy Generated for $\rho=4$}
        \label{fig:fig42}
    \end{subfigure}
    \hspace{1.5em}
    \begin{subfigure}[b]{.3\textwidth}
        \centering
        \includegraphics[width=\linewidth]{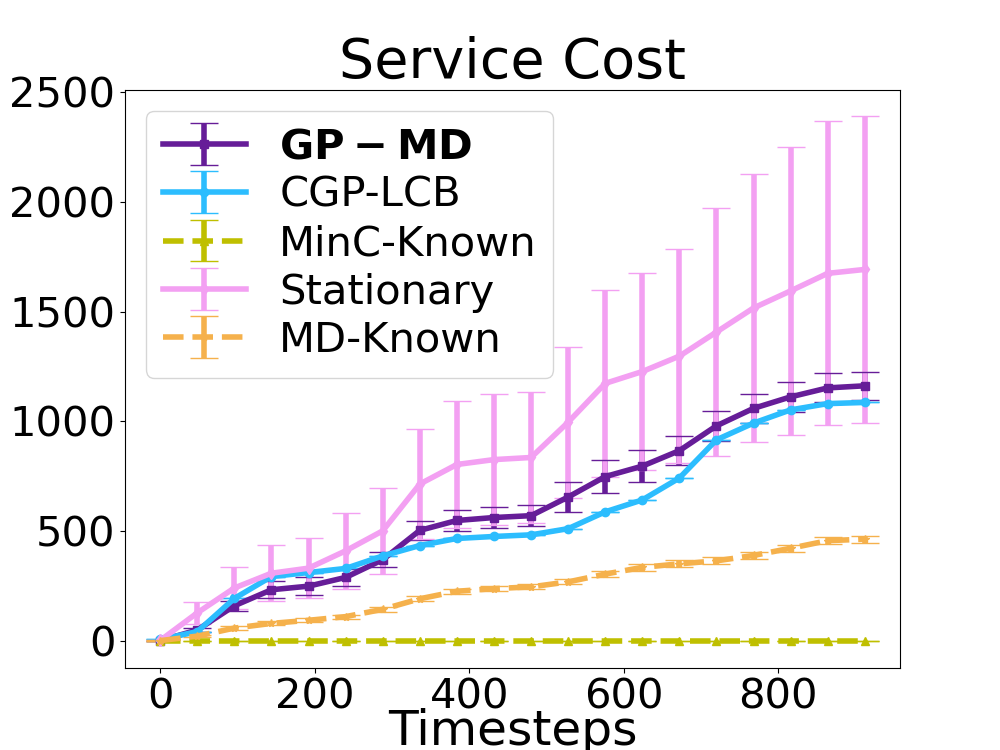}
        %\caption{Figure3}
        \caption{Service Cost for $\rho=4$}
        \label{fig:fig43}
    \end{subfigure}
%        \caption{Here $\alpha$ is defined as $\frac{\rho}{1+\rho}$ and cost values are divided by $1+\rho$ so as to make the costs for different $\rho$ comparable.}
    \caption{\footnotesize AWE altitude optimization task; \Cref{fig:fig41}: Total energy generated for $960$ hours based on the wind data at a single location (Latitude$=47$ and Longitude$=6$). %for various values of tradeoff parameter $\rho$. %All values are divided by a factor of $(1+\rho)$*(minimum movement cost) for easy visualization of performance comparisons. 
    \algnm~outperforms previously used \textsc{CGP-LCB} (that optimizes for service costs only) for a range of $\rho$ values that favor the service against the movement cost. \Cref{fig:fig42}: The average total generated energy over $960$ hours. \Cref{fig:fig43}: The service cost for $\rho=4$.  }
    \label{fig:figure4}
\end{figure}

\begin{figure}[h!]
    \label{fig:fig5}
    \centering
    \begin{subfigure}[b]{.3\textwidth}
        \centering
        \includegraphics[width=\linewidth]{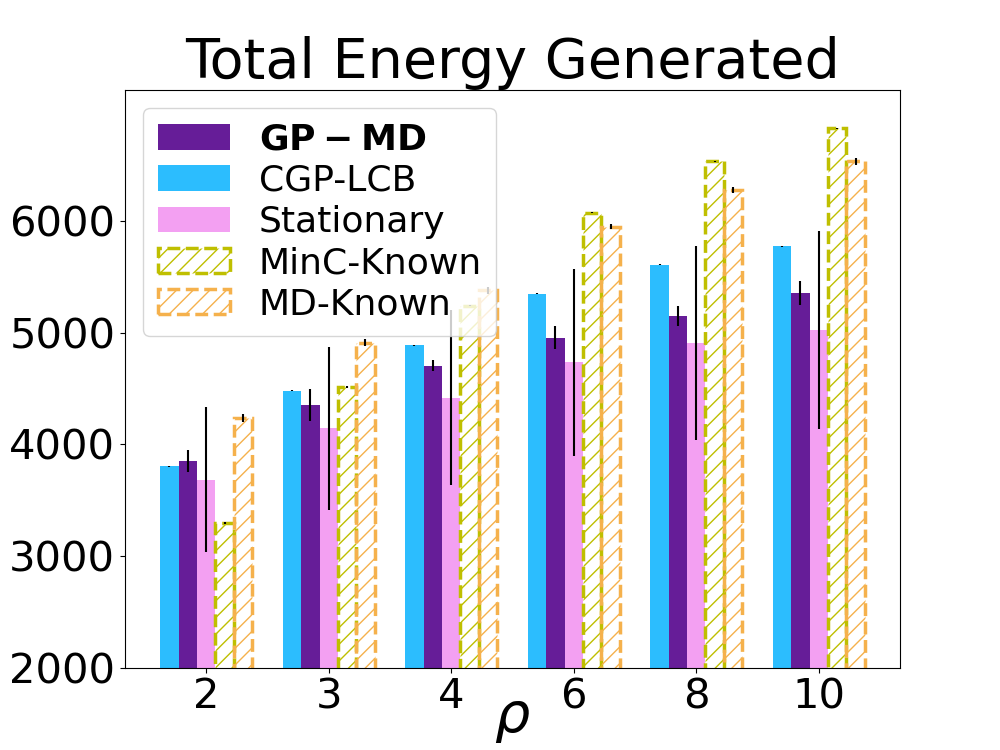}
        \caption{Total Energy Generated for varying $\rho$ }
        \label{fig:fig51}
    \end{subfigure}
    \hspace{1.5em}
    \begin{subfigure}[b]{.3\textwidth}
        \centering
        \includegraphics[width=\linewidth]{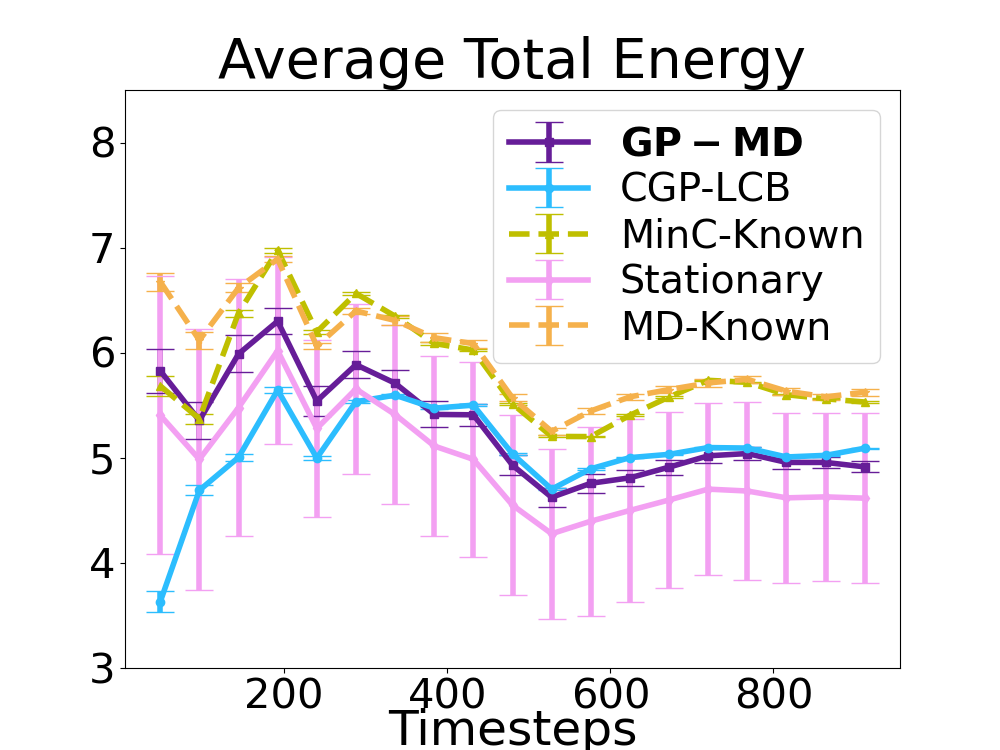}
        %\caption{Figure2}
        \caption{ Total Energy Generated for $\rho=4$}
        \label{fig:fig52}
    \end{subfigure}
    \hspace{1.5em}
    \begin{subfigure}[b]{.3\textwidth}
        \centering
        \includegraphics[width=\linewidth]{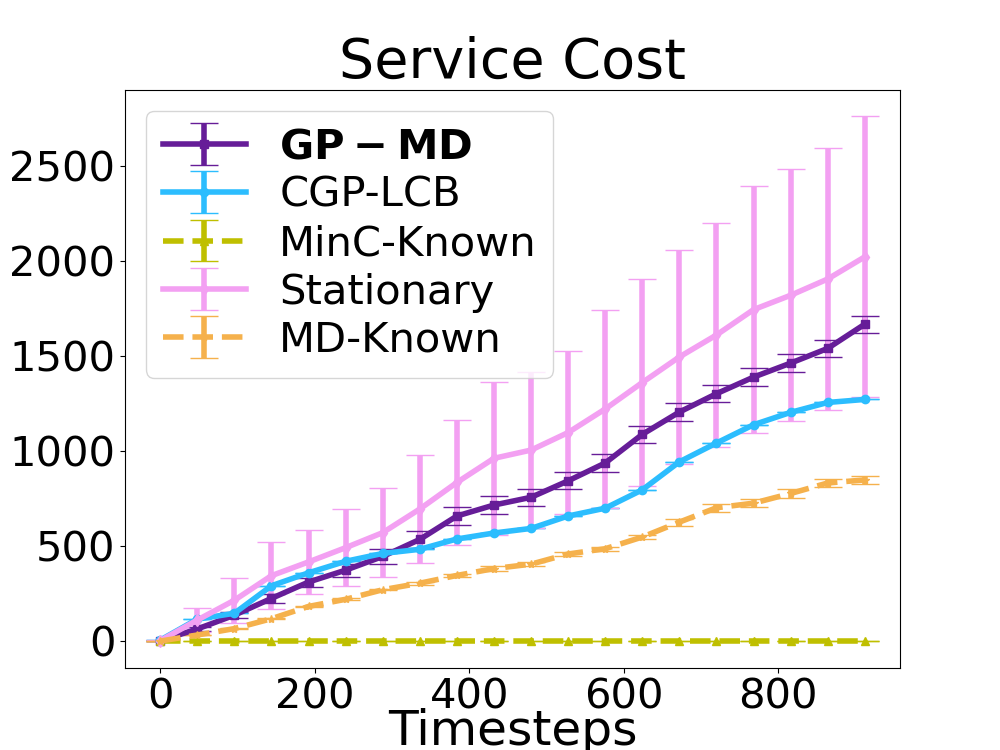}
        %\caption{Figure3}
        \caption{Service Cost for $\rho=4$}
        \label{fig:fig53}
    \end{subfigure}
%        \caption{Here $\alpha$ is defined as $\frac{\rho}{1+\rho}$ and cost values are divided by $1+\rho$ so as to make the costs for different $\rho$ comparable.}
     \caption{\footnotesize AWE altitude optimization task; \Cref{fig:fig51}: Total energy generated for $960$ hours based on the wind data (different months from \Cref{fig:fig11}) at a single location (Latitude$=53$ and Longitude$=-10$). %for various values of tradeoff parameter $\rho$. %All values are divided by a factor of $(1+\rho)$*(minimum movement cost) for easy visualization of performance comparisons. 
    \algnm~ outperforms previously used \textsc{CGP-LCB} (that optimizes for service costs only) only for relatively low $\rho$ values that favor the service against the movement cost. \Cref{fig:fig52}: The average total generated energy over $960$ hours. \Cref{fig:fig53}: The service cost for $\rho=4$.  }
    %    \caption{Comparison of our algorithm with various baselines for altitude optimization at for Loc. 1 ($\text{Lat.}=53, \text{Long.}=-10$) using wind data from October-November 2016:\cref{fig:fig51} corresponds to the total energy generated for a time horizon of 960 hours for various values of tradeoff parameter $\rho$. \cref{fig:fig52} zooms into a particular $\rho=4$ and compares the energy generated. \cref{fig:fig53} shows the service (artificially defined for \cref{alg: mgpbo}) for the same $\rho=4$.  }:
    \label{fig:figure5}
\end{figure}

We present the remaining plots corresponding to the AWE altitude optimization task considered in \Cref{sec:AO_in_AWE}:
\begin{itemize}
    \item  \Cref{fig:figure3}: First, we show additional plots that correspond to Loc. 2 ($\text{Lat.}=53, \text{Long.}=-4$) whose total energy generated for varying $\rho$ was shown in \Cref{fig:fig26}. 
    \item \Cref{fig:figure4}: Next, we consider another Loc. 3 ($\text{Lat.}=47, \text{Long.}=6$; we used $\mathrm{outputscale}=12.94$). 
    \item \Cref{fig:figure5}: Lastly, we show the results that correspond to Loc. 1 ($\text{Lat.}=53, \text{Long.}=-10$), but we use data from a different range (October-November 2016; again, we used $\mathrm{outputscale}=12.94$). 
\end{itemize}

\end{document}